%% file: neurips_2025.tex
\documentclass{article}
\PassOptionsToPackage{numbers,sort&compress}{natbib}
\usepackage[preprint]{neurips_2025}

\usepackage[utf8]{inputenc} 
\usepackage[T1]{fontenc}    
\usepackage{hyperref}       
\hypersetup{
    colorlinks=true,
    linkcolor=blue, %
    citecolor=blue,
    urlcolor=blue,
    hyperfootnotes=blue,
    bookmarks=true
}
\usepackage{url}            
\usepackage{booktabs}       
\usepackage{amsfonts}       
\usepackage{nicefrac}       
\usepackage{microtype}      
\usepackage{xcolor}         
\usepackage{graphicx}
\usepackage{amsmath,amssymb}
\usepackage{enumitem}
\usepackage{multirow}
\usepackage{xspace}
\usepackage[small]{caption}
\usepackage{tcolorbox}
\usepackage{listings}
\usepackage{bm}
\usepackage{subcaption}
\usepackage{float}
\usepackage{adjustbox}
\usepackage{dblfloatfix}
\usepackage{rotating}
\usepackage{wrapfig}

\usepackage{multirow, float, array}
\usepackage{pifont}
\usepackage{seqsplit}          

\usepackage{tikz}
\usetikzlibrary{arrows.meta}

\definecolor{oiorange}{HTML}{E69F00}
\definecolor{oiskyblue}{HTML}{56B4E9}
\definecolor{oigreen}{HTML}{009E73}
\definecolor{oiyellow}{HTML}{F0E442}
\definecolor{oiblue}{HTML}{0072B2}
\definecolor{oivermillion}{HTML}{D55E00}
\definecolor{oipurple}{HTML}{CC79A7}

\definecolor{mgblue}{HTML}{0072B2}
\definecolor{mgbluedark}{HTML}{004E7C}
\definecolor{mgbluelight}{HTML}{E3F0F9}
\definecolor{mgpurple}{HTML}{7B68AE}
\definecolor{mgpurpledark}{HTML}{5B4A8A}
\definecolor{mgpurplelight}{HTML}{EDE8F5}

\usepackage{makecell}
\newcommand{\cmark}{\ding{51}} 

\lstdefinestyle{mindgameslog}{
  basicstyle=\ttfamily\footnotesize,
  breaklines=true,
  breakatwhitespace=false,
  columns=fullflexible,
  keepspaces=true,
  showstringspaces=false,
  frame=none
}

\newtcolorbox{logbox}{
  colback=gray!3,
  colframe=gray!40,
  boxrule=0.6pt,
  arc=3pt,
  left=6pt,
  right=6pt,
  top=6pt,
  bottom=6pt
}

\newcommand{\mindgames}{\textsc{MindGames}\xspace}

\title{
\mindgames: A Live Arena for Evaluating Social and Strategic Reasoning in Multi-Agent LLMs
}

\author{\mindgames Organizer \& Participation Teams, NeurIPS 2025 Competition}

\begin{document}

\maketitle

\input{sections/0_new_abstract}

\input{sections/1_new_intro}

\input{sections/2_related_work}

\input{sections/3_competition_setup}

\input{sections/4_submissions_summary}

\input{sections/5_evaluation_analysis}

\input{sections/6_discussion}
\input{sections/9_conclusion}

\label{sec:biography}
\bibliographystyle{unsrtnat}
\bibliography{ref}
\newpage
\section*{Project Contributors}
\label{sec:contributors}

\paragraph{Core Contributors.} Kevin Wang, Anna Th\"oni, Benjamin Kempinski, Bobby Cheng, Jianzhu Yao

\paragraph{Core Advisors.} Atlas Wang, Tal Kachman

\paragraph{Contributors.} Benjamin Finch, Leon Guertler, Viraj Nadkarni, Yihan Jiang

\paragraph{Advisors.} Mathieu Lauri\`ere, Leshem Choshen, Yoram Bachrach, Pramod Viswanath, Maria Polukarov, Cheston Tan

\paragraph{Sponsors.} Modal Labs, Sentient Foundation, Mithril

\paragraph{Competition Organization.}
Kevin Wang and Jianzhu Yao organized the overall competition structure and co-developed the starter kit. Bobby Cheng and Leon Guertler provided initial recommendations on game selection and infrastructure, with Bobby Cheng subsequently maintaining the competition infrastructure and leaderboard. Anna Th\"oni and Benjamin Kempinski developed selected game environments and their corresponding tutorial code. Kevin Wang and Benjamin Finch developed and maintained the competition website. Anna Th\"oni, Benjamin Kempinski, Jianzhu Yao, Kevin Wang, and Bobby Cheng provided support to participating teams throughout the competition.

\paragraph{Advisory.}
Atlas Wang and Tal Kachman led the advisory effort, providing guidance on the competition format, paper development, and overall planning. Atlas Wang also contributed to the overall organization of the manuscript. Mathieu Lauri`ere helped write and validate the mathematical formulations. Leshem Choshen, Yoram Bachrach, Pramod Viswanath, Maria Polukarov, and Cheston Tan provided broader feedback and guidance.

\paragraph{Analysis and Paper Writing.}
Bobby Cheng, Benjamin Kempinski, Anna Th\"oni, Jianzhu Yao, and Kevin Wang all contributed substantially to the analysis and writing of the report. Jianzhu Yao and Bobby Cheng helped shape the initial framing of the paper. Kevin Wang led the analysis of team methods. Benjamin Kempinski led the error analysis and related work section. Anna Th\"oni led the behavioral diversity analysis. Bobby Cheng led the dataset release and role advantage analysis.

\paragraph{Participation Teams.}
We recognize the participating teams who submitted technical reports describing their approaches. Authors within each team are listed in the order they appear in each team's report.
\begin{itemize}[leftmargin=1.5em]
    \item \textbf{In2AI}: Aliaksei Korshuk, Alexander Buyantuev, Ilya Makarov \textit{(Innopolis University)}.
    \item \textbf{STARS}: Siyuan Wu \textit{(PayPal)}.
    \item \textbf{RLGaming}: Yu-Chi Cheng, Yan-Ru Ju, Ti-Rong Wu, I-Hsuan Chu, Yu-Yu Yang, I-Chen Wu \textit{(National Taiwan University; National Tsing Hua University; National Yang Ming Chiao Tung University; Academia Sinica)}.
    \item \textbf{Odyssean}: Yitian Huang, Qinlu Cao, Yiheng Sun, Yuhong Dai, Hongkun Yao, Jingxuan Fu, Jiwei Zhang, Hao Liao \textit{(Shenzhen University)}.
    \item \textbf{Tungsten}: Mossimo Ebeling.
    \item \textbf{Phoebus}: Mihir S Arya, Avinash Anish, Aditya Ranjan \textit{(RV College of Engineering, Bangalore)}.
    \item \textbf{CerebrAI}: Kirtana Sunil Phatnani, Paval KS, Vrushali Mehta, Aravind S, Nikhil Arora, Tanya Upadhyay, Amol Bandagale \textit{(Fractal Analytics India)}.
    \item \textbf{Awu}: Yuan Lu, ChunEn Hsiao, YuTing Lin, Arvin Chung.
    \item \textbf{DeceptionNet}: Jerry John Thomas \textit{(Indian Institute of Technology Palakkad)}.
    \item \textbf{Team GB}: Govind Arun \textit{(University of Maryland, College Park)}, Sadhvik Bathini \textit{(Indian Institute of Technology Kharagpur)}.
\end{itemize}

\appendix
\input{sections/A_appendix_methods}

\input{sections/B_appendix_observations}
\input{sections/C_appendix_rankings}

\input{sections/D_appendix_naming}
\input{sections/E_appendix_trueskill_timeseries}
\input{sections/F_appendix_evaluation_protocol}
\input{sections/G_appendix_tournament_protocol}

\end{document}

%% file: sections/0_new_abstract.tex
\vskip 0.075in
\begin{tcolorbox}[
    colback=mgpurplelight,
    colframe=mgpurplelight,
    arc=4pt,
    boxrule=0pt,
    left=10pt, right=10pt,
    top=8pt, bottom=8pt,
]
\small
Large language models (LLMs) are increasingly deployed as interactive agents, yet their capacity for social and strategic reasoning over extended interaction remains poorly understood. Existing evaluations rely on static vignettes or single-game benchmarks that cannot capture the sustained, multi-faceted reasoning that real-world multi-agent settings demand. We introduce \mindgames{}, a multi-game arena and evaluation platform for LLM agents that operationalizes complementary reasoning demands relevant to ``theory of mind": belief attribution under hidden information, opponent modeling through repeated strategic interaction, cooperative inference under knowledge asymmetries, and sustained deception in social deduction. Built on TextArena, \mindgames{} provides a unified interaction interface, TrueSkill-based rating, and full trajectory logging across four game environments. 

We instantiate \mindgames{} through the NeurIPS 2025 competition cycle, which assessed 944 submitted agents from 76 teams across four games: Colonel Blotto, Iterated Prisoner's Dilemma, Codenames, and Secret Mafia. Our analysis surfaces both agent-level and evaluation-level limitations: brittle rule adherence remains a major bottleneck, top-performing systems repeatedly rely on explicit structural scaffolding, and leaderboard validity differs sharply across environments. In particular, failure-heavy environments can reward robustness to opponent errors as much as strategic ability, with Secret Mafia exhibiting a pronounced error-survival confound in this cycle. We release a dataset of \textbf{29{,}571 multi-agent games} \textbf{(comprising 94{,}132 player trajectories and 243M tokens)} with turn-level observations, actions, and rewards, together with the \textbf{MindGames Reference Set (MG-Ref)}, a frozen pool of top-ranked, low-error Stage~II submissions paired with a deterministic offline tournament protocol that lets new agents be scored under the same error-attribution lens used in this analysis. By releasing the benchmark interface, dataset, MG-Ref, and starter kit, \mindgames{} provides a reproducible resource for studying progress in multi-agent social and strategic reasoning.

\smallskip
{\footnotesize
\textbf{\color{mgpurpledark}Keywords:}~LLM Agents, Multi-Agent, Theory of Mind, Social Reasoning, Interactive Games\\[2pt]
\textbf{\color{mgpurpledark}Website:}~\url{https://www.mindgamesarena.com/}\\[2pt]
\textbf{\color{mgpurpledark}Dataset:}~\url{https://huggingface.co/datasets/mindgameschallenge/MGC2025}\\[2pt]
\textbf{\color{mgpurpledark}Code \& MG-Ref:}~\url{https://github.com/mind-games-challenge/mindgames-starter-kit}
}
\end{tcolorbox}
\vskip 1ex

%% file: sections/1_new_intro.tex
\section{Introduction and Motivation}

Theory of mind, the capacity to attribute beliefs, intentions, and goals to other agents and to act on those attributions, is a prerequisite for effective multi-agent and human-agent interaction \citep{premack1978does}. In settings from negotiation and resource allocation to cooperative planning and adversarial deception, success depends not on isolated task competence but on modeling what others know, predicting what they will do, and adapting accordingly. As large language models (LLMs) are increasingly deployed as interactive agents in such settings, evaluating the extent and limits of their social and strategic reasoning capabilities becomes a central research question.

\begin{figure}[t]
    \centering
    \includegraphics[width=\linewidth, trim=0 140 0 0, clip]{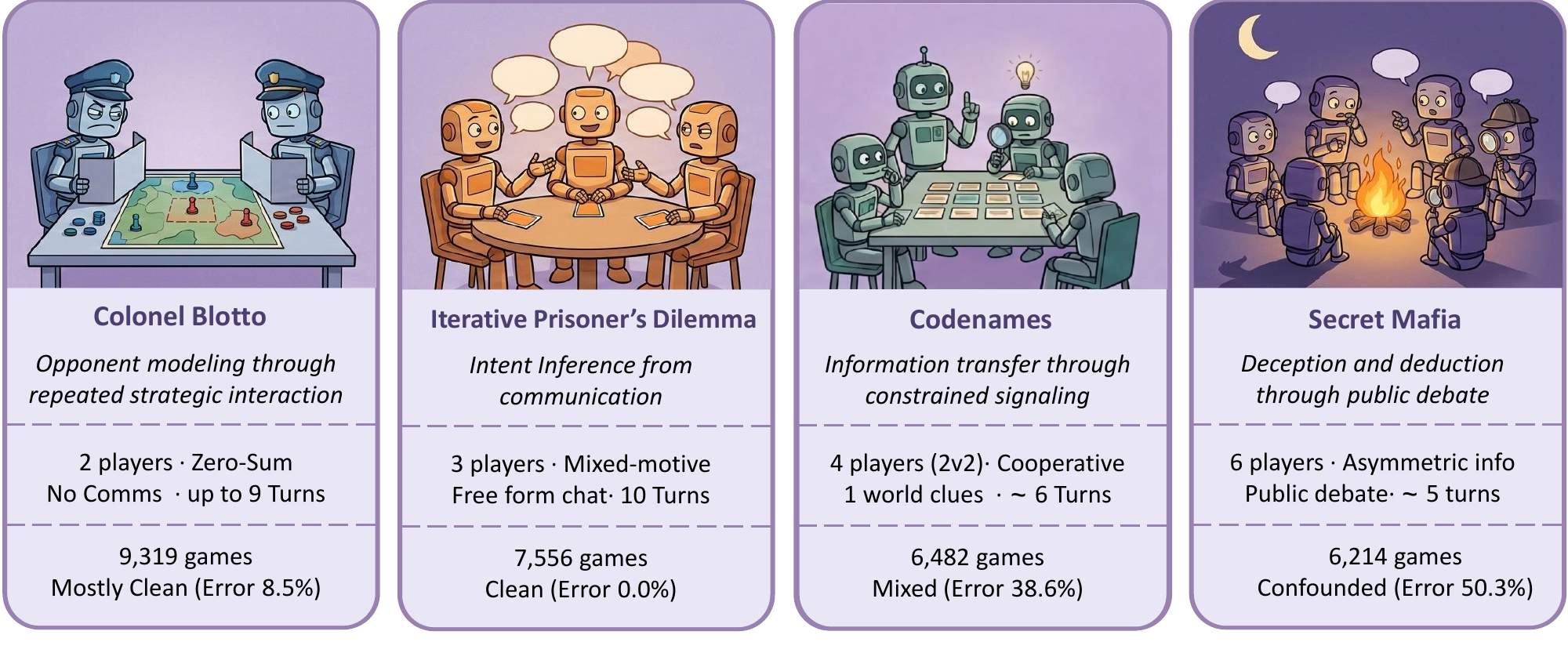}
    \caption{The \mindgames{} game suite and evaluation validity gradient from Stage~II. Each card shows the reasoning demand, game structure, scale, and game-level error rate. IPD and Colonel Blotto yield clean leaderboard signals; Codenames rankings mix strategic skill with constraint-following ability; For Secret Mafia, the overall results were skewed by a small subset of models that generated an unusually high number of errors.}
    \label{fig:game-summary}
\end{figure}

Existing evidence paints a mixed picture. LLMs pass scripted false-belief tasks under standard conditions yet fail under minor perturbations \citep{ullman2023large, tombenchmarks2024broken}, and interactive benchmarks expose weaknesses in belief tracking, deception detection, and commitment consistency across turns \citep{kim2023fantom, huang2025gamabench}. In multi-agent environments, these deficiencies produce brittle coordination, susceptibility to manipulation, and incoherent long-horizon strategies \citep{la2025large, li-etal-2023-theory}. Static vignette-based evaluations cannot capture these dynamics; measuring progress towards theory of mind in LLMs requires settings where agents must reason about and respond to other agents in real time, over sustained interaction.

Multi-agent \emph{games} provide a principled evaluation substrate for these capabilities. Games offer controlled rules, well-defined outcomes, and measurable success criteria, while naturally eliciting belief inference, strategic adaptation, and long-horizon planning through repeated interaction \citep{xu2023language, akata2025playing, hu2025lmgame, sun2025game}. Different game environments probe different facets of interactive social reasoning: hidden-information games test belief attribution under uncertainty; cooperative games test collaborative inference across knowledge asymmetries; and social deduction games test deception and deduction across evolving belief states. In games with multiple rounds per pairing, agents must additionally model and adapt to specific opponents, exercising in-context opponent modeling that single-shot evaluations miss entirely \citep{albrecht2018autonomous, oguntola2025theory}. Yet existing game-based benchmarks remain fragmented across environments, evaluation criteria, and game types, and most do not release large-scale trajectory corpora for downstream analysis \citep{zhu2025multiagentbench, song2025beyond}.

\mindgames{} addresses these gaps by providing a standardized multi-game evaluation platform for interactive, multi-agent reasoning. Built on TextArena \citep{guertler2025textarena}, \mindgames{} unifies environment design, agent interfaces, matchmaking, and full trajectory logging into a single evaluation protocol. Users implement a text-only agent API, submit to an online matchmaking server, and receive TrueSkill-based ratings \citep{herbrich2006trueskill} alongside complete game trajectories for post-hoc analysis. The benchmark spans four games selected along complementary axes of information structure and incentive alignment (Section~\ref{sec:setup}): opponent modeling under hidden information (Colonel Blotto), trust and betrayal through communication (Iterated Prisoner's Dilemma), collaborative inference under constrained signaling (Codenames), and sustained deception under partial observability (Secret Mafia).

The NeurIPS 2025 \mindgames{} competition cycle provides the first large-scale instantiation. Organized into a \textit{Social Deduction Track} (Secret Mafia) and a \textit{Generalization Track} (Colonel Blotto, IPD, Codenames), each with an \textit{Unlimited} and an \textit{Efficient} (${\leq}$9B open-weight) division, the challenge attracted 944 agent submissions from 76 teams. Approaches ranged from fine-tuned open-weight models to agentic systems with structured prompting, memory, and code execution. Beyond producing a live competition, this first cycle also reveals an important measurement lesson: leaderboard reliability is strongly environment-dependent. IPD and Colonel Blotto provide relatively interpretable signals, Codenames mixes strategic performance with error avoidance, and Secret Mafia exposes a pronounced error-survival confound in which rankings can reflect robustness to opponent failures as much as strategic capability. We release the full trajectory dataset together with the benchmark interface and the starter kit. Our contributions can be outlined as follows:
\begin{itemize}[leftmargin=1.5em]
    \item \textbf{A multi-game benchmark for interactive social and strategic reasoning in LLM agents.}
    \mindgames{} provides a unified evaluation platform spanning four game environments that probe complementary reasoning demands related to belief attribution, opponent modeling, cooperative inference, and strategic deception. Built on TextArena \citep{guertler2025textarena}, it supports online matchmaking, TrueSkill-based rating, and full trajectory logging for reproducible evaluation.

    \item \textbf{A rich dataset of multi-agent interaction trajectories.}
    The NeurIPS~2025 evaluation cycle yields 29{,}571 games played (243M tokens), comprising 94{,}132 player trajectories across four environments with turn-level observations, actions, rewards, and metadata.

    \item \textbf{The MindGames Reference Set (MG-Ref) and an offline tournament protocol.}
    We curate a frozen pool of top-ranked Stage~II Efficient submissions with low caused-error rates, together with a deterministic, role-balanced tournament schedule that meets the Stage~II minimum-play thresholds and reports the same error-attribution columns as the main analysis. Participants can rate a new agent against MG-Ref offline and obtain \mindgames{}-compatible TrueSkill and error-attribution metrics.

    \item \textbf{A measurement audit of evaluation validity across environments.}
    Using the competition logs, we analyze how invalid actions, forfeit mechanisms, and metric choice shape leaderboard meaning. We identify a clear error-survival confound in Secret Mafia, show that evaluation reliability differs substantially across games, and characterize where rankings more plausibly reflect strategic play versus robustness to failures.

    \item \textbf{A synthesis of design patterns and failure modes from top-performing systems.}
    Among 944 submitted agents from 76 teams, we distill recurring design choices from successful submissions and document systematic weaknesses in rule adherence, state tracking, and credit assignment, providing practical guidance for future multi-agent LLM systems.
\end{itemize}

The paper is structured as follows: We review the existing literature in Section~\ref{sec:related}. The benchmark design is presented in Section~\ref{sec:setup}. We then analyze submitted methods and gameplay trajectories, distilling observed design patterns from top-performing systems and documenting recurring failure modes in rule adherence, state tracking, and credit assignment (Section~\ref{sec:subsummary}). Section~\ref{sec:evalanalysis} studies evaluation quality through confound analysis, metric disagreement, and behavioral diversity. We discuss implications for benchmark design and interpretation in Section~\ref{sec:discuss}, and conclude in Section~\ref{sec:conclusion}.

%% file: sections/2_related_work.tex
\section{Related Work}
\label{sec:related}

\begin{wraptable}{r}{0.58\textwidth}
\vspace{-1.5em}
\centering
\footnotesize
\setlength{\tabcolsep}{4pt}
\caption{Multi-agent LLM evaluation approaches. All listed benchmarks support interactive play; columns show differentiating properties.}
\label{tab:benchmark-comparison}
\begin{tabular}{@{}lcccc@{}}
\toprule
\textbf{Benchmark}
  & \rotatebox{55}{\textbf{Multi-agent}}
  & \rotatebox{55}{\textbf{Diverse games}}
  & \rotatebox{55}{\textbf{ToM games}}
  & \rotatebox{55}{\textbf{Traj.\ data}} \\
\midrule
AvalonBench \cite{light2023avalonbench}   & \cmark &        & \cmark &        \\
GTBench \cite{duan2024gtbench}      & \cmark & \cmark &        &        \\
GameBench \cite{hu2025lmgame}     & \cmark & \cmark &        &        \\
GAMABench \cite{huang2025gamabench}     & \cmark & \cmark &        &        \\
SPINBench \cite{yao2025spin}    & \cmark & \cmark &        &        \\
WereBench \cite{song2025beyond}     & \cmark &        & \cmark & \cmark \\
Sotopia \cite{zhou2023sotopia}      & \cmark &        &        &        \\
MultiAgentBench \cite{zhu2025multiagentbench} & \cmark & \cmark &        &        \\
TextArena \cite{guertler2025textarena}    & \cmark & \cmark &  \cmark   &        \\
\midrule
\mindgames{} (ours) & \cmark & \cmark & \cmark & \cmark \\
\bottomrule
\end{tabular}
\vspace{-1.5em}
\end{wraptable}

Evaluating social and strategic reasoning in LLM agents requires benchmarks that support \emph{interactive multi-agent play}, cover \emph{diverse game types}, place \emph{theory-of-mind-relevant reasoning demands} near the center of the task, and produce \emph{reusable trajectory data} for post-hoc analysis and training. Existing benchmarks typically satisfy only a subset of these properties. As summarized in Table~\ref{tab:benchmark-comparison}, \mindgames{} is designed to bring them together within a single live evaluation resource.

\paragraph{Theory of mind evaluation: from vignettes to interaction.}
Theory of mind benchmarks for LLMs originated as direct adaptations of cognitive-science protocols: scripted false-belief tasks and belief-attribution vignettes \citep{chen2024tombench, wu2023hi, kim2023fantom, gandhi2023understanding, xu2024opentom}. While useful for establishing baseline capabilities, these static evaluations abstract away the dynamics of sustained interaction, including real-time belief updating, adaptation to non-stationary partners, and strategic reasoning under uncertainty \citep{byom2013theory}. Recent critiques argue that LLMs may exploit surface patterns rather than performing genuine ToM reasoning, and that benchmarks must incorporate interactive, adversarial, or counterfactual components to distinguish social reasoning from pattern matching \citep{tombenchmarks2024broken, ullman2023large}. Interactive game environments address this limitation by requiring agents to maintain and update beliefs about other agents across multiple turns, under conditions where those agents are simultaneously reasoning about them.

\paragraph{Multi-agent game benchmarks.}
Interactive single-game benchmarks probe specific facets of social and strategic reasoning through focused environments: social deduction and deception detection in Avalon \citep{light2023avalonbench} and Werewolf \citep{bailis2024werewolf, xu2023language, song2025beyond, kao2025hidden, kim2025fine}, cooperative inference under information asymmetry in Codenames \citep{stephenson2025codenames} and Hanabi \citep{bard2020hanabi}, and strategic negotiation requiring opponent modeling in Diplomacy \citep{meta2022human, guan2024richelieu}. Multi-game benchmarks broaden coverage: GTBench \citep{duan2024gtbench} and GAMABench \citep{huang2025gamabench} evaluate strategic reasoning across classical game-theoretic scenarios, GameBench \citep{costarelli2024gamebench} and SPINBench \citep{yao2025spin} test generalization across diverse game types, and MultiAgentBench \citep{zhu2025multiagentbench} evaluates collaboration and competition with milestone-based metrics. However, these benchmarks typically assess social reasoning only indirectly through win rates or task completion, without making belief attribution, opponent modeling, or deception an explicit organizing principle of the benchmark design. They also rely primarily on one-off evaluation, with no built-in mechanism for tracking progress across time or controlling for population-dependent rating effects.

\paragraph{Opponent modeling and multi-agent reasoning.}
A parallel research thread investigates whether LLMs can explicitly model other agents. \citet{li-etal-2023-theory} find evidence of emergent collaborative behaviors and higher-order ToM in cooperative text games, but also systematic failures in long-horizon planning. \citet{oguntola2025theory} and \citep{kemp2025gameofthoughts} extends opponent modeling to higher-order belief reasoning, with the former using ToM modeling error as a signal for inducing deceptive behavior, and the latter using iterative game simulations to reduce strategy exploitability. \citet{la2025large} demonstrate that LLMs struggle with multi-agent coordination requiring belief-dependent action selection. In social deduction, LLMs show diverse but generally weak performance on deception, counterfactual reasoning, and strategy alignment with human play \citep{song2025beyond, kao2025hidden, kim2025fine}. A systematic survey of game theory and LLMs \citep{sun2025game} similarly concludes that while LLMs exhibit recognizable negotiation tactics and emerging forms of social reasoning, robust multi-agent coordination remains limited. These findings motivate benchmarks that separate distinct reasoning demands such as belief attribution, opponent modeling, deception, and cooperative inference, rather than collapsing them into a single score.

\paragraph{On evaluation infrastructure.}
Live evaluation platforms come closest to testing sustained interactive reasoning. The Concordia Contest \citep{smith2025evaluating} evaluated agents with TrueSkill-based rating across mixed-motive scenarios, MafiaBench \citep{mafiabench2024} runs social deduction tournaments, and TextArena \citep{guertler2025textarena} supports over 100 games with live scoring. These efforts demonstrate the value of head-to-head play, but they typically target general strategic skill rather than a curated set of complementary social-reasoning demands, and they do not generally release large-scale trajectory datasets for downstream analysis. \mindgames{} builds on this line of work by combining live evaluation infrastructure with a deliberately curated game suite, an explicit focus on complementary forms of social and strategic reasoning, and a released trajectory corpus for post-hoc analysis and future training.

%% file: sections/3_competition_setup.tex
\section{Benchmark Design}
\label{sec:setup}

This section specifies each component in detail: the interaction platform, the game environments and the reasoning behind their selection, the evaluation structure, the rating system, and the released dataset. Operational details specific to the NeurIPS 2025 evaluation cycle (exact dates, stage logistics, qualification criteria) are documented in Appendix~\ref{appendix:protocol}.

\subsection{Interaction Platform}
\label{subsec:env}

\begin{figure}[t]
    \centering
    \includegraphics[width=\linewidth]{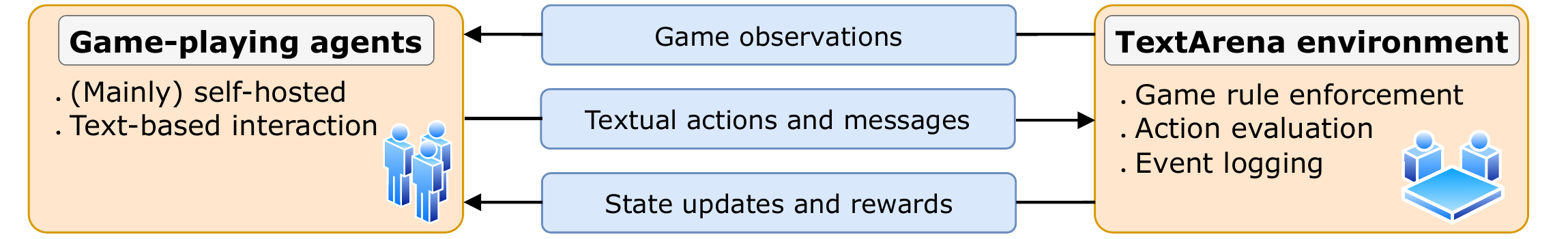}
    \caption{Interaction loop between agents and the TextArena environment. The figure describes a single turn of sequential interaction: agents first receive game observations, then submit textual actions, and finally receive updated observations, state transitions (updates to the internal unobservable state), and terminal rewards when applicable. All interactions are logged for post-hoc analysis and dataset construction.}
    \label{fig:TextArenaEnv}
\end{figure}

\mindgames{} is built on TextArena \citep{guertler2025textarena}, a text-only environment that implements a gymnasium-style interaction loop. A deterministic simulator maintains an internal state $s_t$ (including hidden information), renders a text observation for the acting agent, and receives a textual action in response. The environment then advances the internal state to $s_{t+1}$ via a state transition function, and returns terminal rewards as a dictionary (e.g., \texttt{\{"Player 0": -1, "Player 1": 1\}}) at the end of the game. Players do not receive intermediate rewards. Figure~\ref{fig:TextArenaEnv} illustrates this interaction loop, showing how agents submit actions and receive observations while the environment manages the unobservable internal state transitions.

TextArena provides two properties essential for a live benchmark. First, it supports an online matchmaking server with TrueSkill-based rating \citep{herbrich2006trueskill}, enabling scalable head-to-head evaluation across many agents. Second, it records complete game trajectories, including acting player identity, role, public and private observations, emitted actions, game phase, and resulting state and reward updates, enabling both post-hoc analysis and dataset construction. Examples of each game's starting observations are shown in Appendix~\ref{app:observation}.

\subsection{Game Environments}
\label{subsec:games}

\begin{table}[t]
    \centering
    \resizebox{\linewidth}{!}{%
    \begin{tabular}{lrrrrrrrrrrrr}
        \toprule
        & \multicolumn{4}{c}{Stage I (Online Ladder)} 
        & \multicolumn{4}{c}{Stage II (Final Evaluation)} 
        & \multicolumn{4}{c}{Overall} \\
        \cmidrule(lr){2-5} \cmidrule(lr){6-9} \cmidrule(lr){10-13}
        Environment 
        & $|\mathcal{G}|$ & $|\mathcal{T}|$ & $\mu(T)$ & Tokens (M)
        & $|\mathcal{G}|$ & $|\mathcal{T}|$ & $\mu(T)$ & Tokens (M)
        & $|\mathcal{G}|$ & $|\mathcal{T}|$ & $\mu(T)$ & Tokens (M) \\
        \midrule
        Secret Mafia     
        & 5{,}332   & 29{,}643  & 5.3 & 88.7 
        & 882       & 4{,}716   & 5.4 & 16.1 
        & 6{,}214   & 34{,}359  & 5.3 & 104.8 \\

        Three-player IPD 
        & 6{,}544   & 19{,}632  & 10.0 & 57.2 
        & 1{,}012   & 3{,}033   & 10.0 & 8.8 
        & 7{,}556   & 22{,}665  & 10.0 & 66.0 \\

        Codenames        
        & 5{,}603   & 15{,}678  & 4.7 & 33.2 
        & 879       & 3{,}237   & 6.5 & 5.6 
        & 6{,}482   & 18{,}915  & 5.0 & 38.9 \\

        Colonel Blotto   
        & 7{,}828   & 15{,}243  & 7.4 & 30.0 
        & 1{,}491   & 2{,}950   & 6.9 & 3.2 
        & 9{,}319   & 18{,}193  & 7.4 & 33.3 \\
        \midrule
        \textbf{Total}   
        & 25{,}307  & 80{,}196  & 6.7 & 209.3 
        & 4{,}264   & 13{,}936  & 7.0 & 33.8 
        & 29{,}571  & 94{,}132  & 6.8 & 243.0 \\
        \bottomrule
    \end{tabular}%
    }
    \vspace{5pt}
    \caption{Dataset statistics by environment and evaluation stage. We use $|\mathcal{G}|$ and $|\mathcal{T}|$ to denote the total number of games and player trajectories, respectively, for each environment. $\mu(T)$ denotes the average number of turns per game $G$. The Overall column aggregates Stage I and Stage II. Token counts are computed using the \texttt{Qwen/Qwen3-4B} tokenizer \citep{yang2025qwen3}.}
    \label{tab:dataset-overview}
    \vspace{-1.5em}
\end{table}

A useful multi-agent benchmark must test complementary forms of interactive reasoning rather than a single capability. We selected four game environments along two design axes: (1)~the \emph{information structure} (full vs.\ partial observability, private vs.\ public communication) and (2)~the \emph{incentive alignment} (cooperative, competitive, or mixed-motive). These four environments span two-player to six-player settings, short-horizon to long-horizon interactions, and exercise distinct reasoning demands relevant to social and strategic reasoning: opponent modeling under hidden information (Colonel Blotto), trust-building and betrayal detection through communication (Iterated Prisoner's Dilemma), collaborative inference under constrained signaling (Codenames), and sustained deception under partial observability (Secret Mafia); see Figure~\ref{fig:game-summary} for a summary of key properties.

\paragraph{Colonel Blotto (2 players).}
A zero-sum resource allocation environment in which two players simultaneously allocate a fixed budget of troops across multiple battlefields \citep{roberson2006colonel}. Each battlefield is won by the player who commits more troops, and the overall winner is determined by the number of battlefields won. Allocations are hidden until both players have committed, requiring agents to infer opponent strategies and adapt over repeated rounds without a communication channel.

\paragraph{Iterated Prisoner's Dilemma (3 players).}
An extension of the classic Prisoner's Dilemma to three agents over multiple rounds \citep{matsushima1998evolution}. Each round begins with a bounded communication phase in which agents exchange free-form text messages, followed by a decision phase in which each agent privately chooses to cooperate or defect. Payoffs depend on the joint action profile: mutual cooperation yields high payoffs; unilateral defection yields a windfall for defectors and losses for cooperators; mutual defection yields low payoffs for all. For brevity, we abbreviate the Iterated Prisoner's Dilemma as ``IPD'' throughout this work.

\paragraph{Codenames (4 players).}
A cooperative word-association environment involving two teams of two players each \citep{stephenson2025codenames}. Each team has a spymaster and an operative. A public board contains target words, neutral words, and one assassin word. On each turn, the spymaster provides a one-word clue and a number indicating how many board words it relates to; the operative selects words accordingly. Spymasters must anticipate how operatives will interpret clues, while operatives must infer the intended associations under severe signaling constraints.

\paragraph{Secret Mafia (6 players).}
A social-deduction environment \citep{mafia}. Six players are secretly assigned to an informed minority (mafia) or uninformed majority (villagers), with possible special roles (doctor, detective). The game alternates between private night phases (mafia coordinate eliminations; special roles act) and public day phases (all players discuss and vote to eliminate a suspect). Players know only their own role and public events, creating a setting in which deception, deduction, and state tracking interact over many turns.

\subsection{Evaluation Structure and Rating System}
\label{subsec:tracks}

The benchmark defines two evaluation tracks and two model divisions. The \textbf{Generalization Track} evaluates agents on aggregate performance across Colonel Blotto, IPD, and Codenames, measuring cross-game adaptability. The \textbf{Social Deduction Track} evaluates agents solely on Secret Mafia, where success requires sustained deception and deduction that warrant separate assessment. Within each track, the \textbf{Efficient Agent} division restricts agents to open-weight models under 9B parameters, while the \textbf{Unlimited Agent} division imposes no parameter cap and permits closed-source models. Maintaining both divisions enables systematic comparison of what algorithmic design can achieve under resource constraints relative to frontier-scale systems.

The NeurIPS 2025 cycle instantiated this structure in two stages: an open online ladder (Stage~I) for iterative development, followed by a controlled final evaluation (Stage~II) with frozen submissions and reset ratings. This two-stage design separates exploration from evaluation and is reusable for future cycles. Stage~II enforced minimum-play requirements (50 games in Secret Mafia, 30 games for each Generalization environment) to ensure that ratings reflect sufficient interaction. Full protocol details are in Appendix~\ref{appendix:protocol}.


Agents are rated using TrueSkill \citep{herbrich2006trueskill}, a Bayesian skill rating system that maintains a posterior distribution $\mathcal{N}(\mu, \sigma^2)$ per agent, initialized at $(\mu=25, \sigma=25/3)$. After each match, the posterior is updated based on how surprising the observed outcome is under current beliefs. This makes the method well-suited to variable-size matches (2 to 6 players), team-based environments, and settings in which agents play different numbers of games. In particular, the uncertainty parameter $\sigma$ enables a more meaningful comparison than raw win rates alone under uneven matchmaking. Because any rating system is sensitive to schedule and outcome definition, we report cumulative reward alongside TrueSkill in later analyses, especially in environments where invalid actions and forfeits can distort leaderboard meaning.

\subsection{Formal Game Model}
\label{subsec:formal_game}

To precisely define the interactions in \mindgames, we formulate each multiplayer game $G$ as a tuple:
\begin{equation}
    G = \big(N, \mathcal{S}, s_1, \mathcal{A}, \mathcal{O}, \Omega, \mathcal{I}, \mathcal{T}, \mathcal{R}\big),
\end{equation}
where:
\begin{itemize}[leftmargin=*]
    \item \textbf{Players $N$}: A finite number of players, indexed by $i \in \{1, \dots, N\}$.
    \item \textbf{States $\mathcal{S}$}: A discrete set of internal, unobservable states maintained by the environment's simulator. 
    \item \textbf{Initial State $s_1 \in \mathcal{S}$}: The starting state that encodes initial conditions, symmetric or asymmetric role assignments, and determines the first active player(s) to act.
    \item \textbf{Actions $\mathcal{A}$}: The global action space, defined as all possible natural language token sequences an LLM can generate. Because game engines require specific syntactical structures to parse moves, we distinguish the \textit{admissible action subspace} $\bar{\mathcal{A}}_i(s_t) \subseteq \mathcal{A}$ for player $i$ at state $s_t$. Actions in $\mathcal{A} \setminus \bar{\mathcal{A}}_i(s_t)$ are structurally or legally invalid and handled via environment-specific error rules (see Appendix~\ref{app:invalid_action}).
    \item \textbf{Observations $\mathcal{O}$}: The space of textual observations. The observation function $\Omega_i: \mathcal{S} \to \mathcal{O}$ computes the player-specific observation $o_{i,t}$ for player $i$ from the current internal state $s_t$.
    \item \textbf{Turn Logic $\mathcal{I}$}: An active-player function $\mathcal{I}: \mathcal{S} \to \mathcal{P}(\{1, \dots, N\})$ indicating which subset of players must submit an action at state $s_t$.
    \item \textbf{Transitions $\mathcal{T}$}: A state transition function $s_{t+1} \sim \mathcal{T}(s_t, \boldsymbol{a}_t)$, capturing the environment's state update given the joint action $\boldsymbol{a}_t$ of all active players at turn $t$. 
    \item \textbf{Rewards $\mathcal{R}$}: A terminal reward function $\mathcal{R}_i: \mathcal{S}_{term} \to \mathbb{R}$ that evaluates the final state of the game $\mathcal{S}_{term}$ and assigns a scalar reward to player $i$. There are no intermediate rewards.
\end{itemize}

During gameplay, the environment records exactly one player-specific \textit{trajectory} per active player. The trajectory for player $i$ is denoted $\tau_i = \big((o_{i,1}, a_{i,1}), \dots, (o_{i,T_i}, a_{i,T_i})\big)$, logging their sequence of received observations and emitted actions, where $T_i$ is the number of turns player $i$ acted in. We denote the set of all played games across the benchmark as $\mathcal{G}$.

\paragraph{Specializing the Formalism to Benchmark Games.}
We now instantiate this general definition for each of the four environments (detailed examples of specific observation strings and action formats are available in Appendix~\ref{app:observation}):

\begin{enumerate}
\item \textbf{Colonel Blotto.}
\begin{itemize}[leftmargin=*, nosep]
    \item \textbf{Players \& Turns:} $N=2$. Both players are active simultaneously in every round, so $\mathcal{I}(s_t) = \{1, 2\}$ always. The game terminates after a fixed number of rounds.
    \item \textbf{State \& Observation:} The unobservable state $s_t$ stores the current score, the round number, and both players' secret troop allocations. $\Omega_i(s_t)$ reveals the round index and cumulative score, but hides the opponent's allocations until the round resolves.
    \item \textbf{Admissible Actions:} $\bar{\mathcal{A}}_i(s_t)$ rigidly accepts only valid troop allocations across three fields formatted as \texttt{[A$x$ B$y$ C$z$]}, subject to the explicit troop limit $x+y+z \le U$.
    \item \textbf{Reward:} $\mathcal{R}_i(s_{term})$ yields $+1$ for winning the majority of rounds, $-1$ for losing, and $0$ for a tie.
\end{itemize}

\item \textbf{Iterated Prisoner's Dilemma (IPD).}
\begin{itemize}[leftmargin=*, nosep]
    \item \textbf{Players \& Turns:} $N=3$. All players are simultaneously active, $\mathcal{I}(s_t) = \{1, 2, 3\}$.
    \item \textbf{State \& Observation:} The state $s_t$ alternates between communication phases and decision phases, while tracking accumulated pairwise scores. $\Omega_i(s_t)$ presents the public chat history and the current scoreboard.
    \item \textbf{Admissible Actions:} In communication phases, the admissible space is unrestricted natural language ($\bar{\mathcal{A}}_i = \mathcal{A}$). In decision phases, actions must structurally match the tags \texttt{[<opp-id> cooperate]} or \texttt{[<opp-id> defect]} for each opponent.
    \item \textbf{Reward:} $\mathcal{R}_i(s_{term})$ evaluates the aggregate pairwise payoffs after the fixed game length, assigning $+1$ to the player(s) with the highest total score and $-1$ to all others.
\end{itemize}

\item \textbf{Codenames.}
\begin{itemize}[leftmargin=*, nosep]
    \item \textbf{Players \& Turns:} $N=4$ partitioned into two teams (Red/Blue). Each team has a Spymaster and an Operative. The starting team's Spymaster acts first, e.g., $\mathcal{I}(s_1) = \{1\}$. Turns alternate sequentially between Spymaster and Operative.
    \item \textbf{State \& Observation:} The initial state $s_1$ generates a 25-word board and the hidden word alignments. $\Omega_i(s_t)$ strongly depends on role: Spymasters observe the full alignment mappings, whereas Operatives only observe the public board layout and previously revealed words.
    \item \textbf{Admissible Actions:} Spymaster actions in $\bar{\mathcal{A}}_i(s_t)$ are constrained to a single valid word and number \texttt{[clue $n$]}. Operative actions must exactly match an unrevealed board word \texttt{[word]} or \texttt{[pass]}.
    \item \textbf{Reward:} $\mathcal{R}_i(s_{term})$ maps to $+1$ for the team that correctly guesses all their words (or if the opponent guesses the assassin word), and $-1$ for the losing team.
\end{itemize}

\item \textbf{Secret Mafia.}
\begin{itemize}[leftmargin=*, nosep]
    \item \textbf{Players \& Turns:} $N=6$. The initial state $s_1$ secretly assigns roles. The turn function $\mathcal{I}(s_t)$ activates only specific roles during Night phases (e.g., Mafia) and activates all surviving players during Day phases.
    \item \textbf{State \& Observation:} The state $s_t$ tracks surviving players, role affiliations, and day/night phases. $\Omega_i(s_t)$ is highly asymmetric: Mafia members share a private observation channel with their teammates, while Villagers observe only public daytime eliminations and chat.
    \item \textbf{Admissible Actions:} During the Day phase, $\bar{\mathcal{A}}_i$ allows free text for social deduction and a structured voting token \texttt{[$j$]} targeting survivor $j$. During Night phases, Mafia use \texttt{[$j$]} to coordinate eliminations.
    \item \textbf{Reward:} $\mathcal{R}_i(s_{term})$ provides $+1$ to Mafia agents if their count equals or exceeds the Villagers; otherwise $+1$ to Villagers if they successfully eliminate all Mafia.
\end{itemize}
\end{enumerate}

\subsection{Dataset Release Format}
\label{sec:dataset}

The competition produces a highly rich, public dataset of multi-agent interaction trajectories, released under the CC-BY-4.0 license at \url{https://huggingface.co/datasets/mindgameschallenge/MGC2025}, accompanied by a validated Croissant 1.1 metadata file (Core + Responsible AI fields).
The dataset contains 29{,}571 played games across the four game environments from 944 distinct submitted models (belonging to 76 teams) and organizer-provided baselines, with full natural-language interaction traces and structured metadata.

As defined in Section~\ref{subsec:formal_game}, each game $G \in \mathcal{G}$ logs up to $N$ player-specific trajectories, $\{\tau_1, \dots, \tau_N\}$. Because the active player function $\mathcal{I}(s_t)$ and game termination mechanics control how many turns $T_i$ each player experiences, the lengths of the recorded trajectories ($\tau_i$) naturally vary within the same game. Furthermore, we only record trajectories for players who complete at least one turn ($T_i \ge 1$). Players who do not perform a single action in $G$ (for instance, a player in Secret Mafia who is eliminated in the very first night phase before acting) will not have a corresponding trajectory $\tau_i$ in the dataset.

Invalid actions (when an agent submits an action outside $\bar{\mathcal{A}}_i(s_t)$) and their consequences are environment-dependent. Some environments auto-correct invalid actions into default moves or allow skips, while others treat them as fatal, terminating the game. We record the literal string outputs of all invalid actions in the trajectory logs to facilitate post-hoc alignment and debugging. The exact environment-specific invalid action handling rules are documented in Appendix~\ref{app:invalid_action}.

In addition to the raw sequence of observations and actions, we distribute structured metadata for each game, including environment identifiers, model identifiers mapped to each player index $i$, and the terminal rewards $\mathcal{R}_i(s_{term})$. Table~\ref{tab:dataset-overview} summarizes the dataset scale by environment. The full tabular schema detailing these column structures is provided in Appendix~\ref{appendix:data-schema}.

All data are anonymized and released under a permissive license to support reproducibility and further research in multi-agent, partially observable language-based environments.

%% file: sections/4_submissions_summary.tex
\section{Submitted Methods and Key Findings}
\label{sec:subsummary}

The NeurIPS 2025 evaluation cycle received 944 distinct model submissions from 76 teams. Table~\ref{table:leaderboard-combined} lists the top three teams in each track and division with their final TrueSkill scores and method categories. In2AI ranks first in both Generalization divisions, indicating a consistent advantage across compute regimes within this cycle. In Social Deduction, the Efficient and Unlimited Agent divisions produced different rankings, partly because the two pools experienced different matchmaking dynamics: models in the \textit{Unlimited Agent} division interacted with a broader pool including stabilized baselines, while games in the \textit{Efficient Agent} division were concentrated among similarly rated agents. We therefore interpret rankings primarily within each track and division, and treat cross-division comparisons as qualitative indicators rather than strict like-for-like comparisons. For brevity, we use abbreviated model names throughout this paper. Full model names are given in Appendix~\ref{appendix:full_model_naming}. 

Unless stated otherwise, all subsequent analysis focuses on Stage II model responses, which provide a more stable basis for comparison.

\begin{table}[t]
\centering
\footnotesize
\setlength{\tabcolsep}{4pt}
\renewcommand{\arraystretch}{1.15}
\begin{adjustbox}{width=\textwidth}
\begin{tabular}{lll r c ccc c}
\toprule
& & & &
\textbf{Training} &
\multicolumn{3}{c}{\textbf{Inference-Time}} &
\makecell{\textbf{Output}} \\
\cmidrule(lr){5-5} \cmidrule(lr){6-8} \cmidrule(lr){9-9}
\textbf{Track} &
\textbf{Division} &
\textbf{Team} &
\makecell{\textbf{Final}\\\textbf{Score}} &
\makecell{\textbf{Fine-}\\\textbf{tuning}} &
\makecell{\textbf{Code}\\\textbf{Exec.}} &
\makecell{\textbf{State}\\\textbf{Tracking}} &
\makecell{\textbf{Multi-}\\\textbf{Agent}} &
\makecell{\textbf{Structured}\\\textbf{Output}} \\
\midrule

\multirow{6}{*}{Generalization}
& \multirow{3}{*}{Efficient}
& In2AI    & 34.2 & \cmark &        &        &        & \cmark \\
&         & STARS    & 26.8 &        & \cmark &        &        & \cmark \\
&         & RLGaming & 25.8 & \cmark &        &        &        &  \\
\cmidrule(lr){2-9}
& \multirow{3}{*}{Unlimited}
& In2AI    & 38.0 & \cmark &        &        &        & \cmark \\
&         & RLGaming & 37.1 &        &        &        &        &  \\
&         & Odyssean & 34.2 &        &        &        &        &  \\

\midrule

\multirow{6}{*}{\makecell[l]{Social\\Deduction}}
& \multirow{3}{*}{Efficient}
& RLGaming   & 27.2 & \cmark &        & \cmark &        & \cmark \\
&           & Tungsten  & 23.8 &        &        &        &        & \cmark \\
&           & Odyssean  & 18.4 & \cmark &        &        &        &  \\
\cmidrule(lr){2-9}
& \multirow{3}{*}{Unlimited}
& Phoebus    & 13.9 &        &        & \cmark & \cmark & \cmark \\
&           & CerebrAI  & 7.8  &        &        & \cmark &        & \cmark \\
&           & RLGaming  & 3.2  &        &        & \cmark &        & \cmark \\

\bottomrule
\end{tabular}
\end{adjustbox}

\vspace{6pt}
\caption{
Top three teams from \textbf{Stage II} across tracks and divisions, annotated by design dimensions (Section~\ref{subsec:methods}). \emph{Training}: fine-tuning via SFT or RL. \emph{Inference-time }: code execution, explicit state/memory tracking, multi-agent decomposition. \emph{Output control}: constrained decoding, schema enforcement, or private/public separation.
}
\label{table:leaderboard-combined}
\vspace{-1.5em}
\end{table}

\subsection{Design Patterns and Key Insights}
\label{subsec:methods}

Table~\ref{table:leaderboard-combined} annotates top teams along three design dimensions: whether agents fine-tune on the game trajectories $\mathcal{T}$, whether they employ inference-time scaffolding (code execution, explicit state tracking, multi-agent decomposition), and whether they enforce output validity through constrained decoding or information separation. Full method descriptions appear in Appendix~\ref{appendix:methods}; here we synthesize the cross-cutting patterns that emerged among the strongest submissions in this cycle.

\textbf{a. In the \textit{Efficient Agent} division, training is the common choice among top submissions; in the \textit{Unlimited Agent} division, inference-time structure is more common.}

The clearest pattern in Table~\ref{table:leaderboard-combined} follows the division boundary rather than the track boundary. In the Efficient Agent division, four of six top entries fine-tune on game trajectories $\mathcal{T}$ across the two tracks. In2AI applies reinforcement learning (Appendix~\ref{appendix:in2ai}), while three others use supervised fine-tuning (Appendices~\ref{appendix:rlgaming-gen},~\ref{appendix:rlgaming-sd-eff},~\ref{appendix:odyssean-sd}). Small open-weight models appear to benefit substantially from behavioral adaptation, though STARS (Appendix~\ref{appendix:stars}) demonstrates that code-augmented decision making can substitute, placing second in the Efficient Agent division Generalization track without weight updates. Qwen3-8B also emerged as the dominant Efficient backbone, selected by four of five top performers, with controlled comparisons in one submission reporting advantages over Llama-3.1-8B in five of six conditions (Appendix~\ref{appendix:tungsten}).

In the Unlimited Agent division, the pattern is different. Five of six top entries use no task-specific training, instead relying on stronger base models paired with prompting or inference-time scaffolding. The notable exception is In2AI, whose RL-trained Qwen3-8B, the same checkpoint submitted to Efficient, ranks first in the Unlimited Agent division Generalization track ahead of prompted GPT-5~\citep{openai2025gpt5}, and also received the competition's Most Novel Approach award. Its gains appear to come less from any single novel component than from disciplined system design: reward attribution for delayed credit, per-environment normalization, curriculum-based opponent selection, and asynchronous rollout infrastructure (Appendix~\ref{appendix:in2ai}). Taken together, the top submissions suggest that training-time adaptation is especially valuable under parameter constraints, while larger unrestricted systems can remain competitive through stronger base models and inference-time structure.

\textbf{b. Data curation appears more important than raw data volume.}

Among fine-tuned submissions, aggressive quality filtering repeatedly outperforms training on larger but noisier corpora. Strict score-5-only retention discards 80\% of LLM-judged trajectories yet produces stronger agents than training on the full corpus (Appendix~\ref{appendix:odyssean-gen}). Eligibility gating filters non-strategic steps before reward assignment, training only on decisions with observable consequences (Appendix~\ref{appendix:in2ai}). A third approach generates over 10{,}000 instances from carefully orchestrated scenarios against a strong teacher (Appendix~\ref{appendix:rlgaming-sd-eff}). Across these submissions, the recurring pattern is not simply ``more data helps,'' but rather that in multi-agent settings, where trajectory quality depends strongly on opponent behavior, curation often matters more than scale.

\textbf{c. Cognitive scaffolding without paired training often backfires.}

Perhaps surprisingly, adding memory, reflection, or structured reasoning modules to models not trained to use them actively can hurt performance rather than help it. Two independent teams provide convergent evidence. A cross-turn memory module via XML blocks causes an 8B model to underperform its simpler variant (Appendix~\ref{appendix:tungsten}), and adding a memory-and-deduction layer without fine-tuning drops win rate from 25\% to 16.7\%, while SFT on strong-model traces then raises the same architecture to 45\% (Appendix~\ref{appendix:rlgaming-sd-eff}). A third team explicitly keeps its LLM frozen as a feature extractor, avoiding the instabilities of end-to-end adaptation (Appendix~\ref{appendix:deceptionnet}). These reports suggest a common failure mode: models generate brittle summaries or unreliable intermediate states when they are not trained to exploit the added structure \citep{xie2026memo}. Cognitive architecture therefore seems most effective when paired with corresponding training, or else kept deliberately simple.

\textbf{d. Independent teams converge on modular perception--reasoning--action pipelines.}

Multiple teams, working independently, arrive at architectures that separate observation parsing, belief updating, and action generation into distinct components: a four-module pipeline with separate listening, belief tracking, state building, and policy heads (Appendix~\ref{appendix:deceptionnet}); a three-stage pipeline progressing from memory to review to tone-adapted action (Appendix~\ref{appendix:revac}); and a dual-agent split between situation assessment and action selection that yields a 4.7$\times$ improvement over monolithic baselines (Appendix~\ref{appendix:phoebus}). Competitive submissions also repeatedly introduce structured intermediate representations between raw game logs and final decisions, whether as probabilistic role distributions, social alignment graphs, belief--desire--intention (BDI) belief states, or graph-encoded game states. Despite different implementations, the shared intuition is that the gap between raw text observations and strategic action is too large to bridge reliably in a single unstructured forward pass.

\textbf{e. In social deduction, communication tone and deduction under uncertainty are decisive bottlenecks.}

Two findings specific to Secret Mafia are especially revealing. First, the communication tone itself becomes a strategic variable. One team selects among Aggressive, Withdrawing, Logically Anchoring, and Contrarian styles based on game state (Appendix~\ref{appendix:revac}), while another's iterative development across roughly 25 agent versions converges on influence-focused strategies after testing paradigms ranging from personality modeling to analytical reasoning (Appendix~\ref{appendix:cerebrai}). This has no close analog in traditional game-playing AI, where communication is usually absent or formalized. Second, architectural improvements disproportionately benefit the information-poor Villager role. One team reports a 9$\times$ improvement on the Village side versus 2.5$\times$ on the Mafia side (Appendix~\ref{appendix:phoebus}); another lifts Villager win rate from 16\% to 60\% while Mafia remains at 88--96\% (Appendix~\ref{appendix:rlgaming-sd-unl}). Across submissions, informed roles appear easier for baseline LLMs than deduction under uncertainty. Several teams also report that models remain reluctant to lie explicitly as Mafia, exposing a persistent tension between strategic deception and the truthfulness bias induced by RLHF.

\subsection{Failure Mode Analysis}
\begin{wrapfigure}{r}{0.5\textwidth}
    \centering
    \vspace{-1em}
    \includegraphics[width=\linewidth]{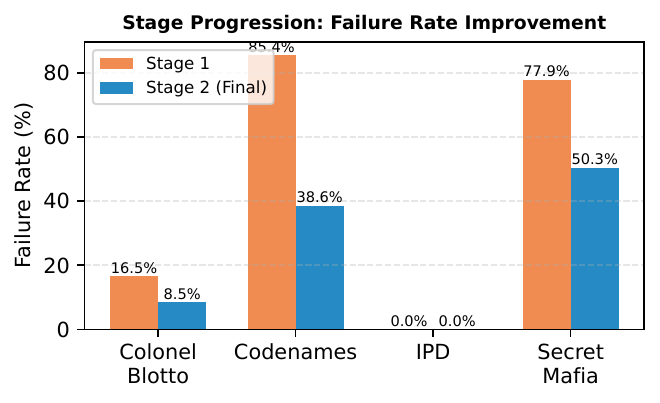}
    \caption{Error rates between Stage~I and Stage~II across all four environments. Codenames showed the greatest improvement (85.4\% to 38.6\%), followed by Secret Mafia (77.9\% to 50.3\%) and Colonel Blotto (16.5\% to 8.5\%). IPD maintained 0\% errors across both stages. Each bar reports the percentage of games containing at least one error (fatal or non-fatal). For Colonel Blotto and Codenames, all errors are fatal; for Secret Mafia, the rate includes non-fatal errors that are auto-corrected (see Appendix~\ref{app:invalid_action}).}
    \label{fig:error_stage_progression}
    \vspace{-2em}
\end{wrapfigure}
\label{subsec:failure_design}

\textbf{f. Non-LLM decision policies remain viable.}

A 6.8M-parameter graph attention network (Appendix~\ref{appendix:team-gb}) achieves 78.4\% win rate within Colonel Blotto using LLMs purely as teachers for preference generation and distillation. A hybrid architecture similarly treats its LLM as a frozen perception module while a trainable RL policy handles decisions (Appendix~\ref{appendix:deceptionnet}). These submissions suggest that LLMs are valuable as knowledge sources and feature extractors, but that effective game-playing policies can be orders of magnitude smaller than the models used to supervise them.



A key advantage of TextArena is that it enforces rule compliance and makes invalid actions observable. The exact consequences of invalid actions are environment-dependent: some are fatal and end the game or eliminate the player, while others are retried or silently corrected. This section analyzes the resulting error patterns across models and games; Appendix~\ref{app:invalid_action} gives the environment-specific handling rules in full. For brevity and clarity, we use abbreviated model names throughout this section. Full official model names are provided in Appendix~\ref{appendix:full_model_naming}.

\paragraph{Game-specific error patterns.}
Error prevalence varied dramatically across environments. Table~\ref{table:all_game_error_agg} aggregates all Stage~II games by environment and reports the percentage of games containing at least one error, whether fatal or non-fatal. IPD stands out as the only game with zero recorded errors, likely due to its simple binary action space and default fallback mechanism for invalid actions. In contrast, the other three environments exhibited substantial error rates.

Secret Mafia proved most problematic, with 50.3\% of games $\mathcal{G}$ (444 out of 882) containing at least one error. The dominant error type was invalid moves (399 cases), with a further 45 cases involving attempts to protect already-eliminated players, an error pattern analyzed in detail below. Codenames showed 38.6\% of games $\mathcal{G}$ (339 out of 879) containing an error. The sole cause was illegal clues (339 cases), where players directly revealed hidden words rather than providing indirect associations. Colonel Blotto exhibited an 8.5\% error rate split between invalid input formatting (114 cases) and illegal unit allocations (11 cases), reflecting LLM difficulties with numerical constraint satisfaction.

As shown in Figure~\ref{fig:error_stage_progression}, error prevalence reduced substantially between Stage~I and the final stage across all environments that exhibited failures. The steepest gains appeared in environments dominated by 
free-form language generation under constraints (Colonel Blotto and Codenames), while environments requiring longer-horizon state tracking (Secret Mafia) remained persistently difficult despite iterative refinement.

\begin{table}[t]
\centering
\small
\setlength{\tabcolsep}{4pt}
\renewcommand{\arraystretch}{0.95}
\caption{
Error aggregation across all four environments in Stage~II. \# Participants: number of unique models that participated in the environment; \# Games: total number of games played ($|\mathcal{G}|$); Error~(\%): percentage of games containing at least one error (fatal or non-fatal); Top Reason and 2nd Reason: most common error types with total occurrence counts in parentheses.
}
\begin{tabular}{lrrrrlll}
\toprule
Environment & \# Participants & \# Games & Clean (\%) & Error (\%) & Top Reason & 2nd Reason \\
\midrule
Colonel Blotto  & 47 & 1491 & 91.5   & 8.5  & Invalid input (114) & Illegal units (11) \\
Codenames       & 52 & 879  & 61.4   & 38.6 & Illegal clue (339) & None \\
IPD             & 41 & 1012 & 100.0  & 0.00 & None & None \\
Secret Mafia    & 55 & 882  & 49.7   & 50.3 & Invalid move (399) & Protection (45) \\
\bottomrule
\end{tabular}
\vspace{6pt}
\label{table:all_game_error_agg}
\end{table}

\paragraph{Constraint violations $\rightarrow$ structured output and code execution.}
LLMs struggle most with constraint satisfaction when the constraint space is combinatorial or semantically defined, as the error rates above make clear. Two design responses repeatedly appeared among stronger submissions. First, \emph{code-augmented decision making} through Program-Aided Language architectures allows the LLM to generate executable code to parse game state, compute candidate actions, and validate constraints before committing (Appendix~\ref{appendix:stars}). Second, \emph{structured output enforcement} uses action validators to check format and game-rule compliance during gameplay and to feed typed failure metadata into training, enabling models to learn from constraint violations rather than treating terminations as unstructured noise (Appendix~\ref{appendix:in2ai}). In Codenames, generated code filters candidate clues against assassin and opponent words; in Colonel Blotto, optimization routines compute legal allocations directly.

\paragraph{State tracking degradation $\rightarrow$ memory and deduction layers.}
Secret Mafia exhibited a 50.3\% error rate from two related causes: invalid moves (399 cases) and attempts to protect already-eliminated players (45 cases). Both reflect the same underlying pathology: models operating on stale representations of game state, either failing to produce contextually valid actions or acting on outdated information about who remains in play. This is consistent with documented long-context processing challenges \citep{li2024long, liu2024lost}. 

Top Social Deduction submissions shift state maintenance from implicit context-window reasoning to explicit structured representations, including role-specific memory with turn-by-turn summarization (Appendix~\ref{appendix:rlgaming-sd-eff}), BDI belief states with self-improvement loops (Appendix~\ref{appendix:rlgaming-sd-unl}), and dual-agent architectures separating situation assessment from action selection (Appendix~\ref{appendix:phoebus}). As discussed in Section~\ref{subsec:methods}, however, these memory structures appear to help only when paired with corresponding training; without that pairing, additional complexity can degrade performance.

\paragraph{Information leakage $\rightarrow$ private/public reasoning separation.}
In Secret Mafia, a further failure mode is inadvertent role exposure, where Mafia agents reveal knowledge of teammate identities or special-role players disclose their abilities through reasoning traces that leak into public actions. Because the game's core mechanic is information asymmetry, even a single leaked signal can collapse the strategic landscape.

Multiple teams addressed this through thinking-mode architectures that generate private reasoning blocks for role-aware strategic analysis while an external harness extracts only the public portion, architecturally preventing leakage without fine-tuning (Appendix~\ref{appendix:tungsten}, Appendix~\ref{appendix:cerebrai}). These systems suggest that inference-time separation between private reasoning and public communication can be sufficient for information control, even without weight updates.

\paragraph{Training data narrowness $\rightarrow$ multi-source trajectories and staged pipelines.}
Models trained on a narrow distribution of opponent strategies risk converging toward policies that exploit specific local weaknesses rather than developing robust play against diverse opponents. Several competition submissions reported exactly this brittleness: RLGaming's fine-tuned agents overfit to specific Colonel Blotto allocation patterns and collapsed into all-defect equilibria in IPD (Appendix~\ref{appendix:rlgaming-gen}), while DeceptionNet abandoned self-play PPO entirely because of policy degradation and non-stationarity (Appendix~\ref{appendix:deceptionnet}).

Teams responded by broadening the source distribution of training trajectories through mixtures of rule-based agents, RL-trained agents, and proprietary model traces (Appendix~\ref{appendix:rlgaming-gen}), and by using staged pipelines of SFT, CoT-SFT from stronger teachers, and GRPO with opponent diversity (Appendix~\ref{appendix:odyssean-gen}). As discussed in Section~\ref{subsec:methods}, aggressive curation at each stage appeared more important than sheer data volume. Even so, a persistent risk remains: many learned policies appear to converge toward exploiting the particular opponents encountered during training rather than developing robust opponent modeling. Closing this gap between localized strategy and more general social and strategic reasoning likely requires broader and more systematically varied opponent pools than any single team used in this cycle.

%% file: sections/5_evaluation_analysis.tex
\section{Auditing Live-Arena Measurement Validity}
\label{sec:evalanalysis}

Beyond the agent-level patterns of Section~\ref{sec:subsummary}, the corpus lets us audit what live-arena rankings actually measure. Three findings shape how the leaderboards from this cycle should be interpreted: an \emph{error-survival confound} that varies sharply by environment, a role-dependent advantage in social deduction, and a behavioral-similarity gradient that mirrors the rule structure of each game. We then introduce \textbf{MG-Ref}, an offline evaluation protocol that allows future agents to be scored against the 2025 cohort under the same measurement lens applied here.

\subsection{Evaluation Confounds}
\label{subsec:confounds}

\paragraph{The error-survival confound varies by game.}
Each game handles invalid actions differently (Appendix~\ref{app:invalid_action}): some errors are \emph{fatal} (causing game termination or player elimination), while others are \emph{non-fatal} (retried or auto-corrected). IPD auto-corrects all errors; Colonel Blotto and Secret Mafia allow one retry before a fatal consecutive error; Codenames treats illegal clues as immediately fatal. As a result, the meaning of a win or rating gain is environment-dependent. Full per-model error breakdowns are presented in Appendix~\ref{app:detailed_errors}.

IPD recorded zero forfeited games across all models; however, this is partly an artifact of the environment, which automatically replaces invalid actions with a default \texttt{[cooperate]} action (Appendix~\ref{app:invalid_action}). Consequently, comparative error rates across the benchmark reflect both the intrinsic difficulty of the reasoning demands and the differential leniency of the environments. At the other extreme, Colonel Blotto’s top three models caused zero game terminations across a combined 446 games, reflecting highly stable constraint satisfaction.

Table~\ref{table:top_model_specific_error} reports error statistics for the top-3 ranked Efficient Agent models in Stage~II. \textit{Clean} counts games where no player committed any error. \textit{Caused} counts games in which the focal model committed at least one error (fatal or non-fatal). \textit{Witnessed} counts games in which an opponent committed at least one error. \textit{Self-Forf.}\ and \textit{Opp-Forf.}\ count only fatal errors that terminated the game. These categories are not mutually exclusive: a game can appear in both \textit{Caused} and \textit{Witnessed} if both the focal model and an opponent erred.

The table reveals a clear confound gradient. In Codenames, caused-error rates range from 5.8\% (In2AI) to 24.6\% (STARS), while opponent error rates range from 34\% to 52\%; since all Codenames errors are fatal, \textit{Caused} and \textit{Self-Forf.}\ coincide. In Secret Mafia, the asymmetry is much stronger: RLGaming caused errors in only 4 of 130 games but witnessed opponent errors in 129, while Tungsten caused errors in 10 of 132 games but witnessed opponent errors in 131. Only one game per top model was fully clean. Crucially, most Mafia errors are non-fatal: RLGaming’s 4 caused errors all produced self-forfeits, but of 129 witnessed opponent-error games, only 61 led to opponent forfeits. This gap between \textit{Caused}/\textit{Witnessed} and the forfeit columns shows that error prevalence far exceeds game termination rates, and that top-ranked models can benefit substantially from simply surviving games in which other agents fail.

\begin{table}[t]
\centering
\small
\setlength{\tabcolsep}{4pt}
\renewcommand{\arraystretch}{0.95}
\caption{Error statistics for the top-3 ranked Efficient Agent models per game in Stage~II. \textit{Clean}: games with no errors from any player. \textit{Caused}: games in which this model committed at least one error (fatal or non-fatal). \textit{Witnessed}: games in which an opponent committed at least one error. \textit{Self-Forf.}: games terminated by this model's fatal error. \textit{Opp-Forf.}: games terminated by an opponent's fatal error. For Colonel Blotto and Codenames all errors are fatal, so Caused~$=$~Self-Forf.\ and Witnessed~$=$~Opp-Forf. For Secret Mafia, most errors are non-fatal, producing a large gap between Caused/Witnessed and the forfeit columns. Categories are not mutually exclusive. Colonel Blotto, Codenames, and IPD rankings use the Generalization Track ranking.}\label{table:top_model_specific_error}
\begin{tabular}{l r l r r r r r r}
\toprule
Environment & Rank & Model & Games & Clean & Caused & Witnessed & Self-Forf. & Opp-Forf. \\
\midrule
\multirow{3}{*}{Colonel Blotto}
 & 1 & In2AI        & 98 & 97 & 0 & 1 & 0 & 1 \\
 & 2 & STARS V7     & 105 & 103 & 0 & 2 & 0 & 2 \\
 & 3 & RLG-8B-V12   & 243 & 240 & 0 & 3 & 0 & 3 \\
\midrule
\multirow{3}{*}{Codenames}
 & 1 & In2AI        & 69 & 31 & 4 & 34 & 4 & 34 \\
 & 2 & STARS V7     & 69 & 28 & 17 & 24 & 17 & 24 \\
 & 3 & RLG-8B-V12   & 69 & 27 & 6 & 36 & 6 & 36 \\
\midrule
\multirow{3}{*}{IPD}
 & 1 & In2AI        & 195 & 195 & 0 & 0 & 0 & 0 \\
 & 2 & STARS V7     & 163 & 163 & 0 & 0 & 0 & 0 \\
 & 3 & RLG-8B-V12   & 257 & 257 & 0 & 0 & 0 & 0 \\
\midrule
\multirow{3}{*}{Secret Mafia}
 & 1 & RLG-TS-V7        & 130 & 1 & 4 & 129 & 4 & 61 \\
 & 2 & Tungsten-Soc-V2  & 132 & 1 & 10 & 131 & 5 & 63 \\
 & 3 & Odyssean-Soc     & 106 & 1 & 0 & 105 & 0 & 46 \\
\bottomrule
\end{tabular}
\end{table}

\paragraph{Codenames: an intermediate confound.}
Codenames clean-game rates are uniformly low among the top three, ranging from 39\% (RLG-8B) to 45\% (In2AI), meaning over half of all games involve at least one error. Caused-error rates vary substantially: STARS caused errors in 24.6\% of games, RLG-8B in 8.7\%, and In2AI in only 5.8\%. Opponent error rates range from 34.8\% to 52.2\%, indicating that a large fraction of games are affected by opponent errors. Since all Codenames errors are fatal (illegal clues cause immediate termination), non-forfeited and error-free are equivalent in this environment. These patterns suggest that Codenames rankings reflect a mixture of strategic capability and error avoidance.

\begin{figure}[t]
    \centering
    \includegraphics[width=\linewidth]{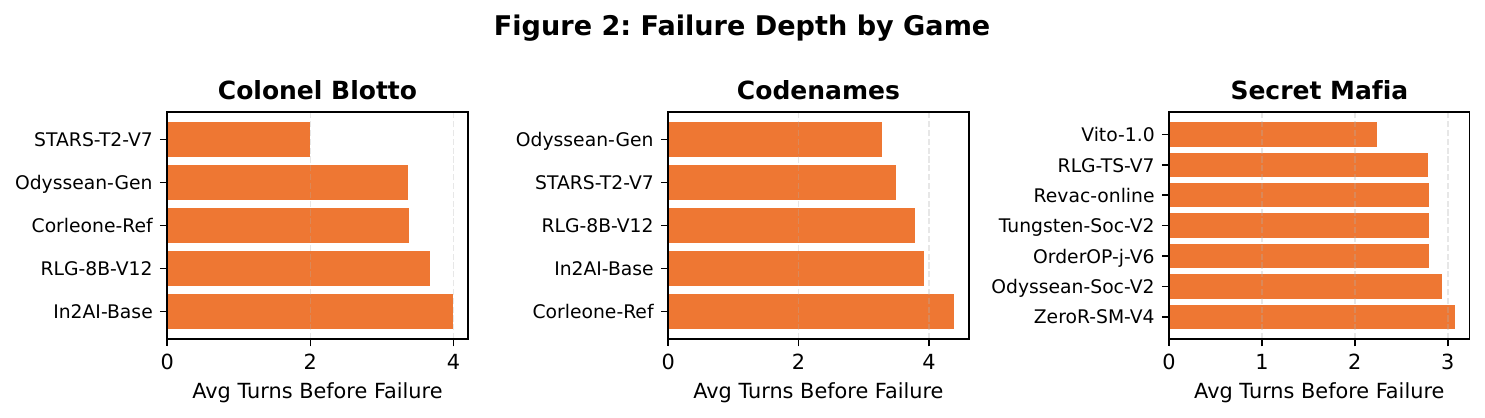}
    \caption{Average number of turns before failure for top models in each Stage~II game with premature terminations (IPD excluded due to zero failures). Secret Mafia shows particularly early failures (under 3 turns) relative to typical game length (8--12 turns), indicating that most terminations occurred before meaningful strategic play. Colonel Blotto and Codenames also show early failures (3--4 turns), but these represent a larger fraction of expected game length.}
    \label{fig:failure_depth}
\end{figure}

\paragraph{Secret Mafia: a deeper confound.}
In Secret Mafia, fewer than 1\% of games for any top model are fully error-free: virtually all games contain non-fatal errors (invalid Detective or Doctor actions that are silently corrected), even when the game completes without forfeit. Figure~\ref{fig:failure_depth} reinforces the severity: games that do terminate end after fewer than 3 turns on average, well below the expected 8--12 turn game length, meaning many terminations occur before meaningful strategic interaction. This \emph{error-survival confound}, where rankings reflect robustness to opponent errors rather than strategic capability, has direct implications for benchmark design, as discussed in Section~\ref{sec:discuss}.

\paragraph{A simple termination-depth diagnostic.}
Combined with the per-environment asymmetry above, the failure-depth pattern in Figure~\ref{fig:failure_depth} produces a clean validity hierarchy: IPD and Colonel Blotto yield interpretable strategic rankings; Codenames mixes strategic skill with rule-adherence robustness; Secret Mafia rankings in this cycle are dominated by error survival. We propose two simple environment-level diagnostics that reproduce this hierarchy and transfer to other live-arena evaluations: \emph{game-level error rate} and \emph{median termination depth as a fraction of expected length}. When error rate exceeds roughly $30\%$ and median termination depth falls below half of expected length, leaderboard position should be interpreted as \emph{robustness to opponent failure} rather than as strategic skill.

\begin{figure}[t]
    \centering
    \includegraphics[width=\linewidth]{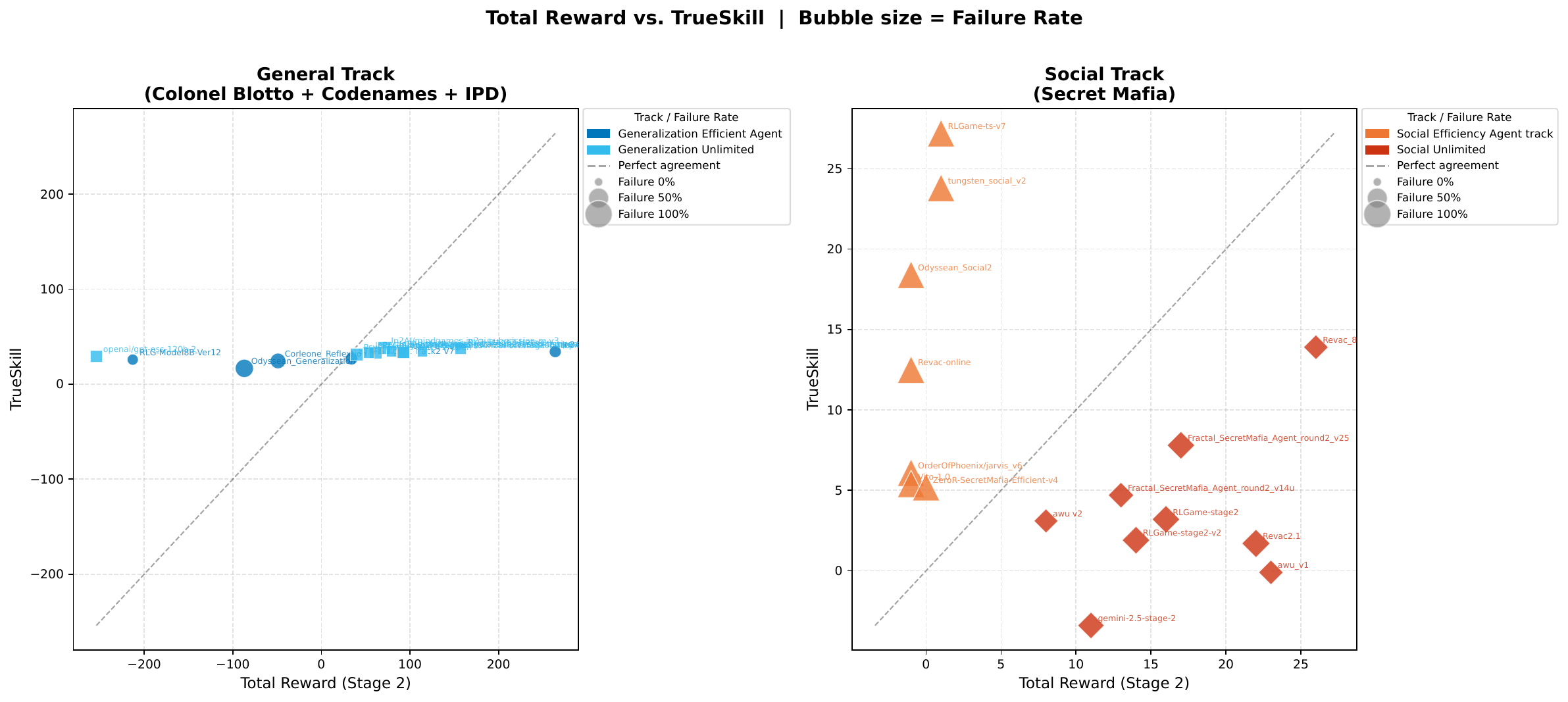}
    \caption{TrueSkill rating versus total reward for top models across the Generalization Track (Colonel Blotto, Codenames, and IPD; left) and Social Deduction Track (Secret Mafia; right) in Stage~II. The dashed diagonal indicates perfect agreement between metrics; points far from it indicate metric disagreement. Bubble size reflects the proportion of games in which the model itself caused a failure. In the Generalization Track, models span a wide reward range ($\approx -120$ to $+150$) yet compress into a narrow TrueSkill band ($\approx 0$ to $+50$), showing that TrueSkill is relatively insensitive to cumulative performance differences under the observed outcome structure. In the Social Deduction Track, models from the Efficient Agent division (triangles) accumulate positive TrueSkill despite near-zero reward, while models from the Unlimited Agent division (diamonds) show the inverse, illustrating how metric choice can reverse relative rankings depending on failure distribution.}
    \label{fig:figure_trueskill_vs_reward_panels}
\end{figure}

\paragraph{TrueSkill and reward capture different signals.}
Figure~\ref{fig:figure_trueskill_vs_reward_panels} compares TrueSkill ratings against cumulative reward. The two metrics are complementary, and their divergence is itself informative. In the Generalization Track (left), TrueSkill ratings compress into a narrow band ($\approx 0$ to $+50$) despite raw rewards spanning $-120$ to $+170$. This partly reflects the nature of pairwise rating systems under calibrated matchmaking: TrueSkill emphasizes relative outcome consistency against the opponent pool, while cumulative reward preserves margin and absolute outcome accumulation. As a result, TrueSkill alone can obscure substantial differences in total performance.

In the Social Deduction Track (right), the divergence is more diagnostically useful. Efficient Agent division models (triangles) accumulate positive TrueSkill despite near-zero total reward. Under calibrated matchmaking, near-zero reward is partly expected, since agents face similarly rated opponents and win roughly as often as they lose. But in combination with the error patterns in Table~\ref{table:top_model_specific_error}, this also reflects the error-survival confound: models can accumulate pairwise wins when opponents fail, inflating TrueSkill without corresponding gains from sustained strategic play. Unlimited Agent division models (diamonds) show the inverse pattern, accumulating higher rewards but lower TrueSkill, suggesting that stronger absolute outcomes were not always translated into consistent pairwise dominance. The main lesson is not that one metric is ``correct'' and the other is not, but that leaderboard interpretation depends on which notion of performance is being emphasized.

\subsection{Behavioral Diversity}
\label{subsec:diversity}

\begin{figure}[t]
    \centering
    \begin{subfigure}[t]{0.4\textwidth}
        \centering
        \includegraphics[width=\textwidth]{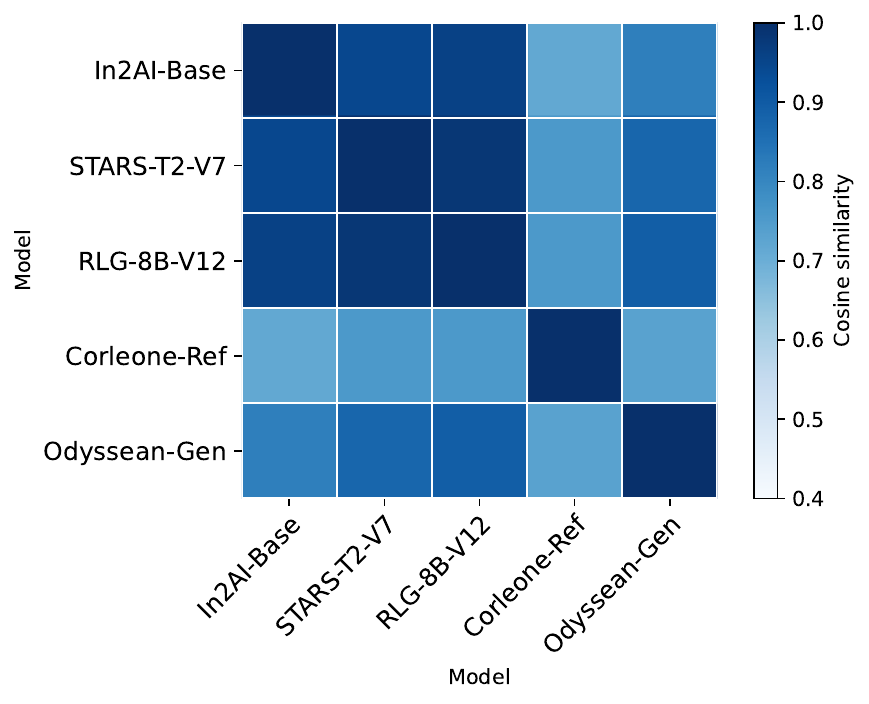}
        \caption{}
    \end{subfigure}
 \hspace{1cm}
    \begin{subfigure}[t]{0.4\textwidth}
    \centering
    \includegraphics[width=\textwidth]{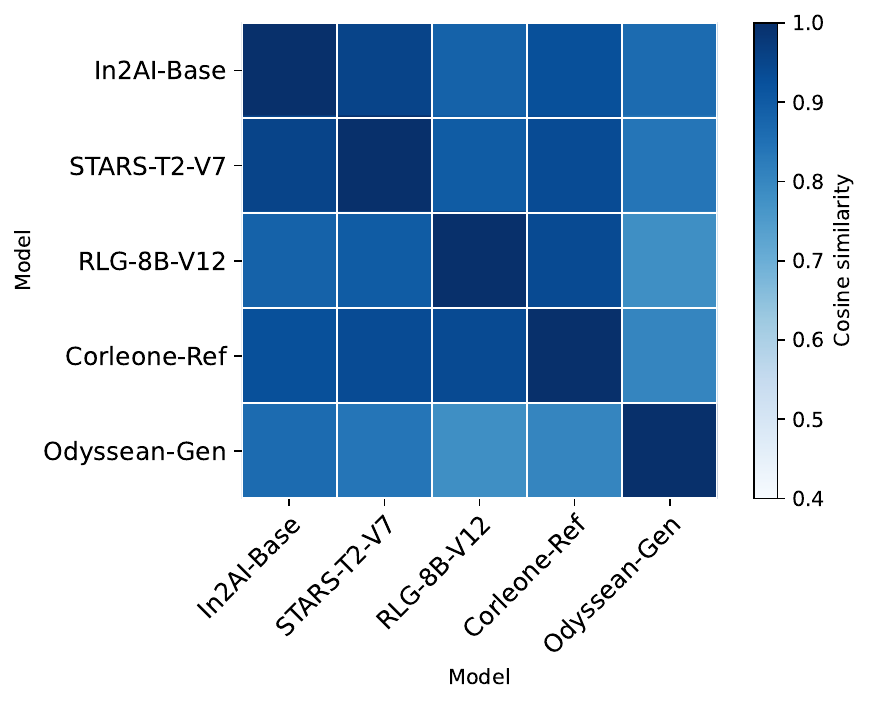}
    \caption{}
    \end{subfigure}
 \hfill
    \begin{subfigure}[t]{0.4\textwidth}
    \centering
    \includegraphics[width=\textwidth]{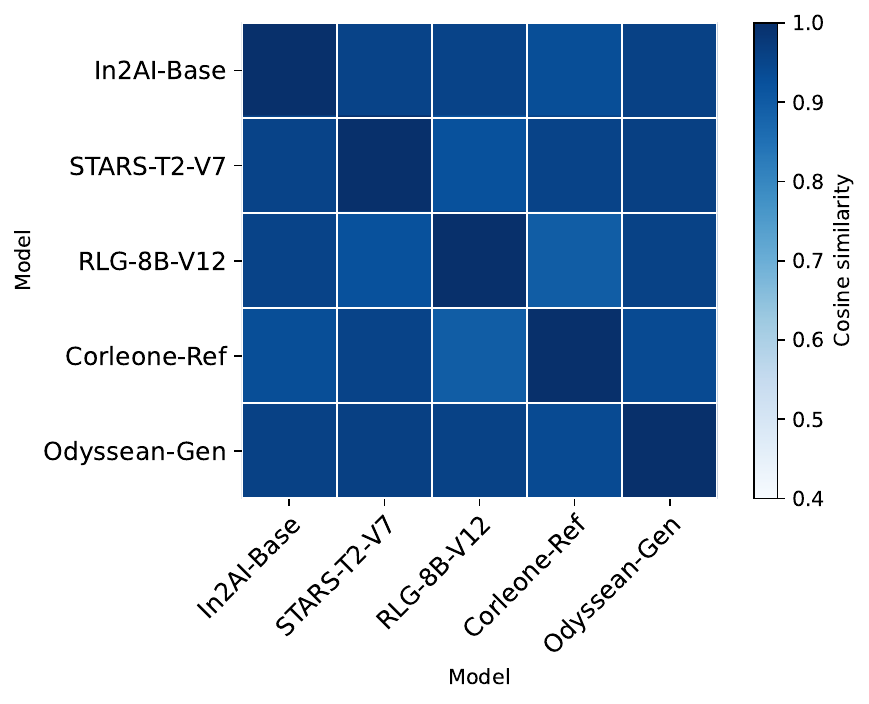}
    \caption{}
    \end{subfigure}
\hspace{1cm}
    \begin{subfigure}[t]{0.4\textwidth}
    \centering
    \includegraphics[width=\textwidth]{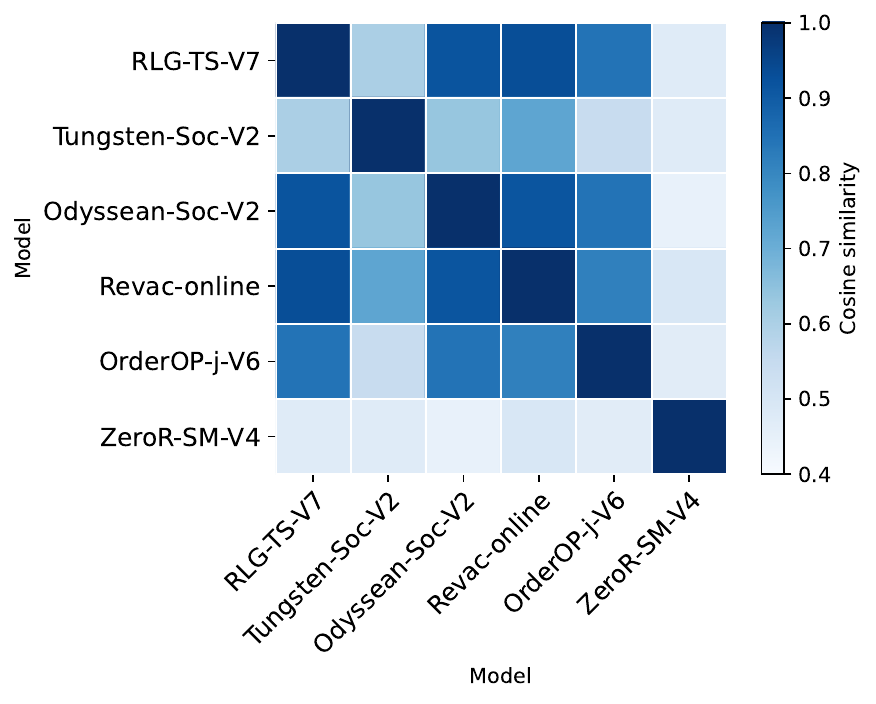}
    \caption{}
    \end{subfigure}
    \caption{Cosine similarities between the average responses $\bar{\bm{r}}$ of the Stage~II Efficient Agent models. Responses are grouped by environment: (a) Colonel Blotto, (b) IPD, (c) Codenames, and (d) Secret Mafia.}
    \label{fig: embeddings}
\end{figure}

We analyze strategic convergence and divergence among models using textual embeddings of their game responses. Focusing on Efficient Agent division Stage~II submissions, we embed each response into $\bm{r} \in \mathbb{R}^{1536}$ using \textit{OpenAI text-embedding-3-small}~\citep{OpenAITE3}. Next, we represent each model by the mean of its embedded responses,
\begin{equation}
    \bar{\bm{r}} = \frac{1}{n}\sum_{i = 1}^{n} \bm{r}_i,
\end{equation}
and compute pairwise cosine similarity,
\begin{equation}
    S_{i,j} = \frac{\bar{\bm{r}}_i \cdot \bar{\bm{r}}_j}{\|\bar{\bm{r}}_i\| \, \|\bar{\bm{r}}_j\|}
    \label{eq: cosine_sim}
\end{equation}
for all model pairs. Figure~\ref{fig: embeddings} shows the similarity matrices per game $G$. Because response format strongly influences surface similarity, Codenames and Colonel Blotto, which use highly structured outputs, naturally exhibit higher cross-model similarity than IPD and Secret Mafia.

The results reveal environment-dependent convergence patterns. In IPD (Figure~\ref{fig: embeddings}b), all response similarities exceed 0.8, consistent with the environment's simple action space. In Colonel Blotto (Figure~\ref{fig: embeddings}a), the top three models, In2AI, STARS, and RLG-8B with TrueSkill scores of 36.8, 31.8, and 29.8 respectively (Appendix~\ref{appendix: final rankings}), cluster tightly, while lower-ranked models such as Corleone (TrueSkill: 17.4) and Odyssean (5.5) diverge substantially, suggesting that winning strategies occupy a relatively narrow region of the behavioral space. Codenames (Figure~\ref{fig: embeddings}c) shows uniformly high similarity across all models, which is unsurprising given its highly constrained output format of a single word and number per turn. Secret Mafia (Figure~\ref{fig: embeddings}d) presents the most heterogeneous landscape, with the top four models splitting into two similarity clusters. Here, Tungsten's (23.8) responses diverge markedly from those of RLGaming (27.2), Odyssean (18.4), and Revac (12.5), despite achieving comparable TrueSkill. This suggests that multiple viable behavioral styles coexist in the social deduction setting. Because only final responses, rather than internal reasoning traces, are available for embedding, this analysis should be interpreted as capturing behavioral signatures rather than underlying reasoning processes.

\subsection{Role Advantages}

For Codenames and Secret Mafia, we analyze both the distribution of role assignments within the games played and their impact on win rates.

For each model \( m \) and role \( r \), we define the role advantage as
\begin{equation}
A_{m,r} = w_{m,r} - \bar{w}_m, \quad
\bar{w}_m = \frac{1}{|R_m|} \sum_{r' \in R_m} w_{m,r'}
\end{equation}
where \( w_{m,r} \) is the win rate of model \( m \) when assigned role \( r \), and \( \bar{w}_m \) is the model’s average win rate across roles. Positive values indicate that a model performs better than its own average when assigned a given role, while negative values indicate worse-than-average performance.

\begin{figure}[t]
    \centering
    \vspace{-1em}
    \begin{subfigure}[t]{0.45\textwidth}
        \centering
        \includegraphics[width=\textwidth]{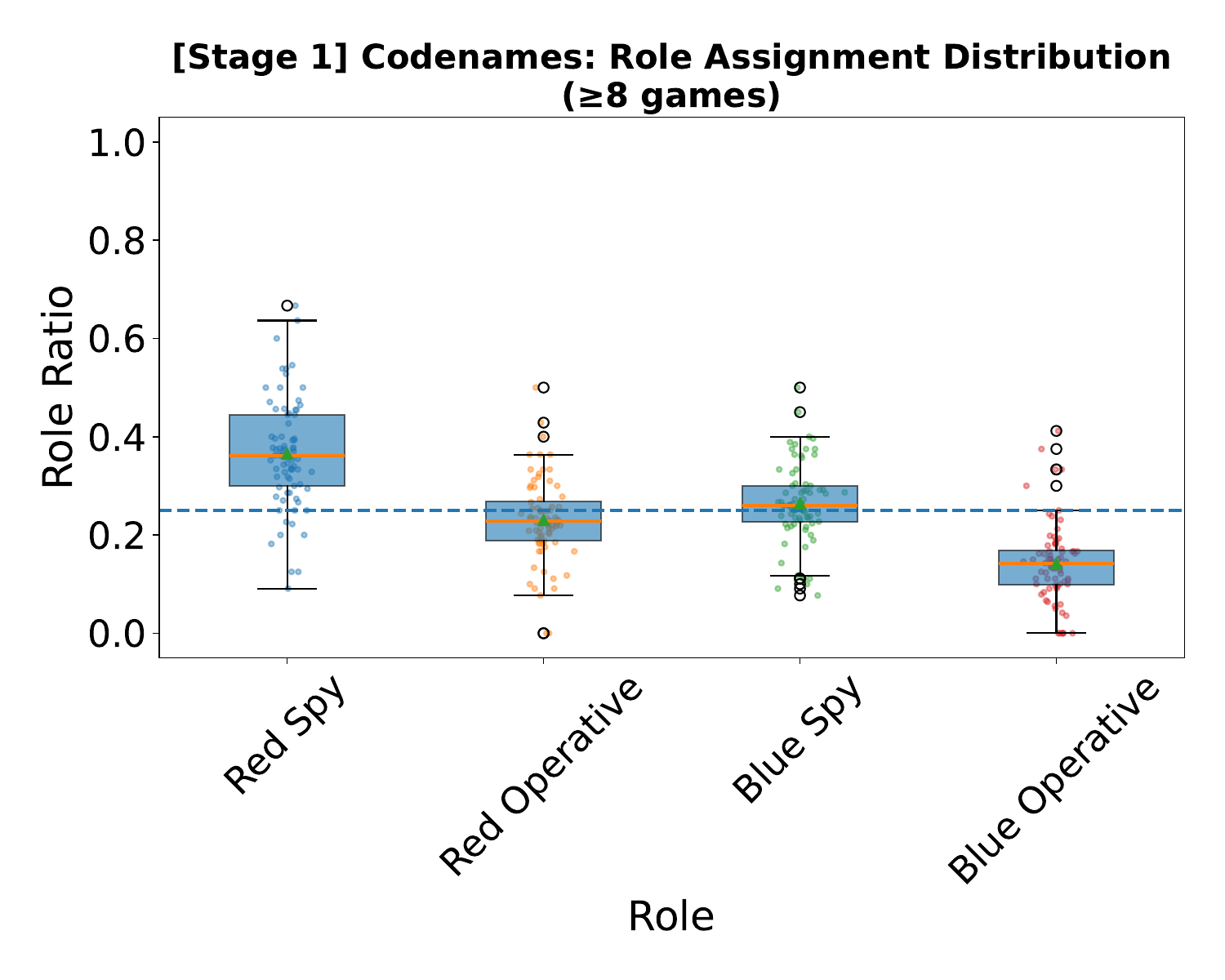}
        \caption{}
    \end{subfigure}
 \hspace{1cm}
    \begin{subfigure}[t]{0.45\textwidth}
    \centering
    \includegraphics[width=\textwidth]{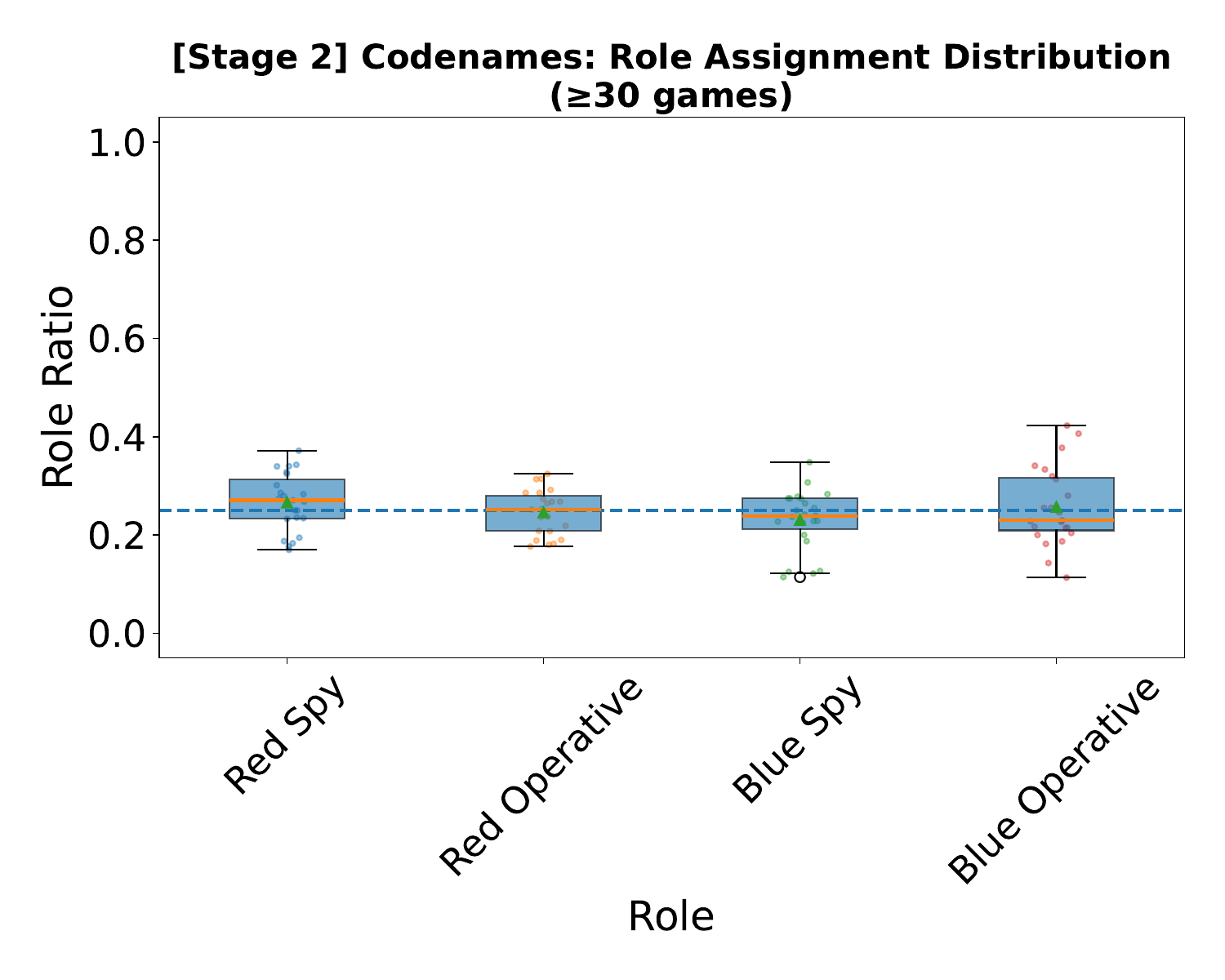}
    \caption{}
    \end{subfigure}
 \hfill
    \begin{subfigure}[t]{0.45\textwidth}
    \centering
    \includegraphics[width=\textwidth]{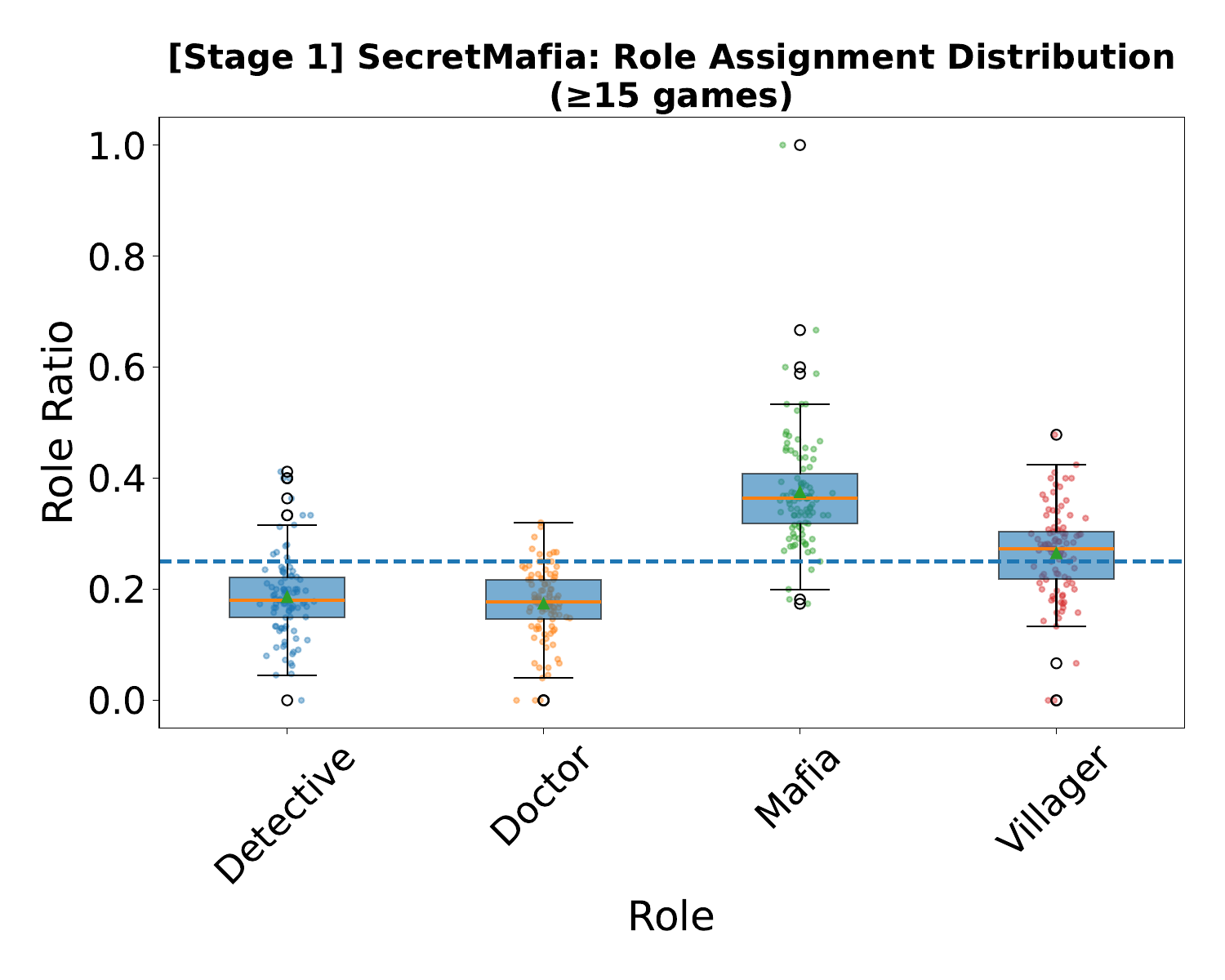}
    \caption{}
    \end{subfigure}
\hspace{1cm}
    \begin{subfigure}[t]{0.45\textwidth}
    \centering
    \includegraphics[width=\textwidth]{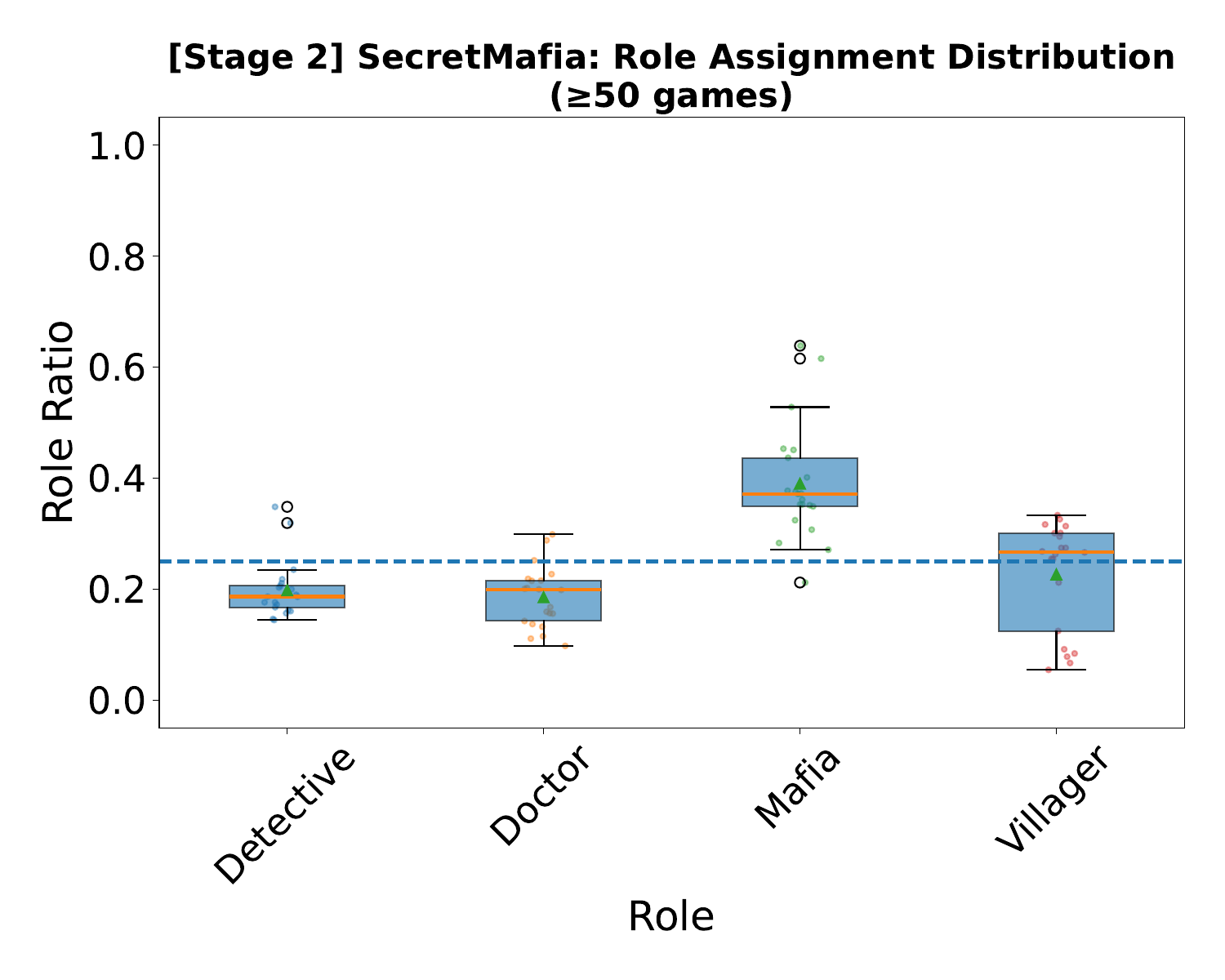}
    \caption{}
    \end{subfigure}
    \caption{Role distribution across Codenames and Secret Mafia over Stage~I (left) and Stage~II (right). Each boxplot summarizes the distribution of role assignment ratios across models. The central line indicates the median, while the box represents the interquartile range (IQR). The dashed horizontal line denotes the expected uniform assignment across roles. Individual points show per-model role frequencies, and circular markers highlight statistical outliers beyond 1.5$\times$IQR.}
    \label{fig:role-dist}
    \vspace{-1em}
\end{figure}

As shown in Figure~\ref{fig:role-dist}, Stage~I of Codenames exhibits a slight imbalance in role assignments when allowing models with as few as 8 games, with a higher frequency of the Red Spy role. Increasing the minimum participation threshold to 30 games in Stage~II results in a substantially more uniform distribution across roles.

\begin{figure}[t]
    \centering
    \begin{subfigure}[t]{0.45\textwidth}
        \centering
        \includegraphics[width=\textwidth]{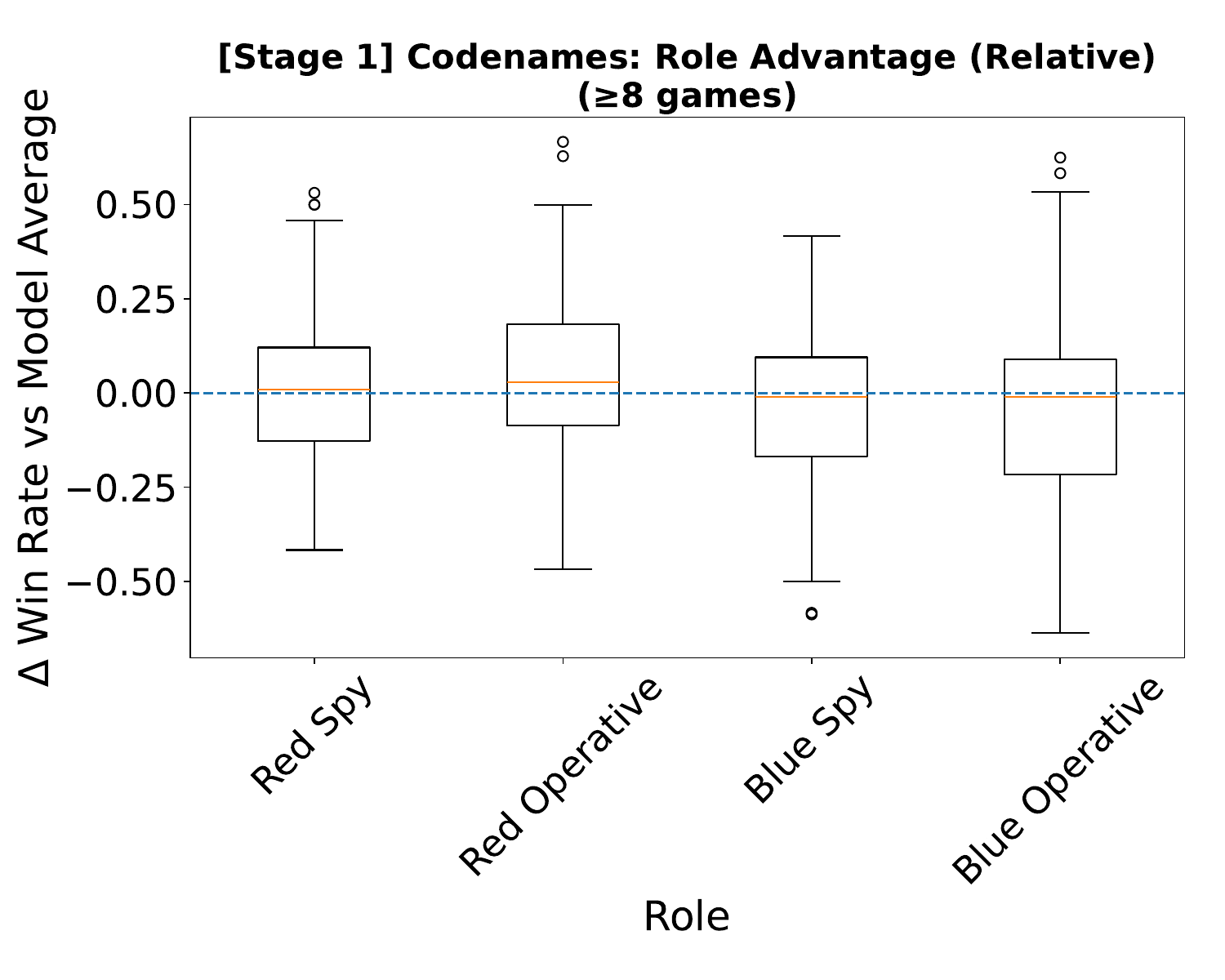}
        \caption{}
    \end{subfigure}
 \hspace{1cm}
    \begin{subfigure}[t]{0.45\textwidth}
    \centering
    \includegraphics[width=\textwidth]{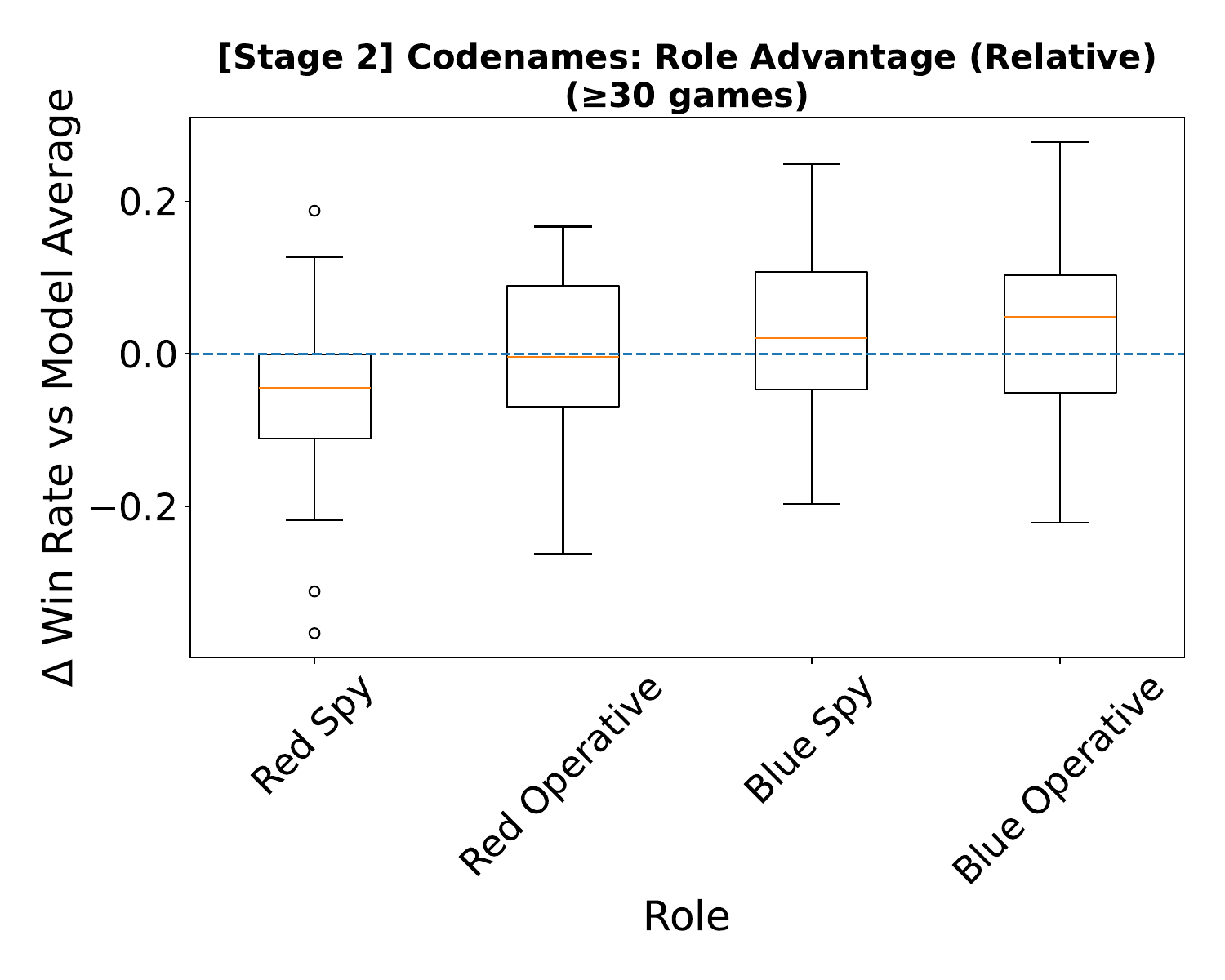}
    \caption{}
    \end{subfigure}
 \hfill
    \begin{subfigure}[t]{0.45\textwidth}
    \centering
    \includegraphics[width=\textwidth]{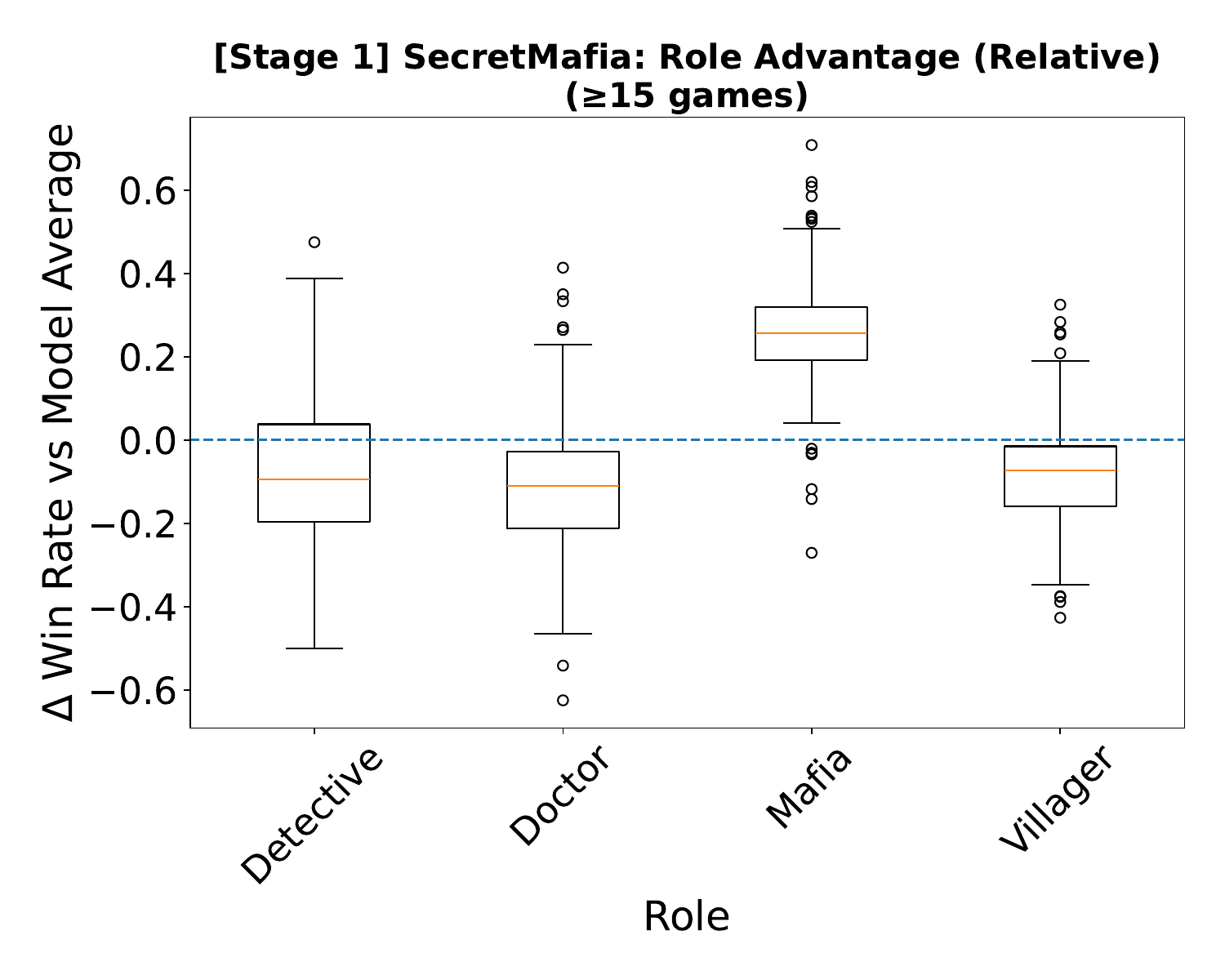}
    \caption{}
    \end{subfigure}
\hspace{1cm}
    \begin{subfigure}[t]{0.45\textwidth}
    \centering
    \includegraphics[width=\textwidth]{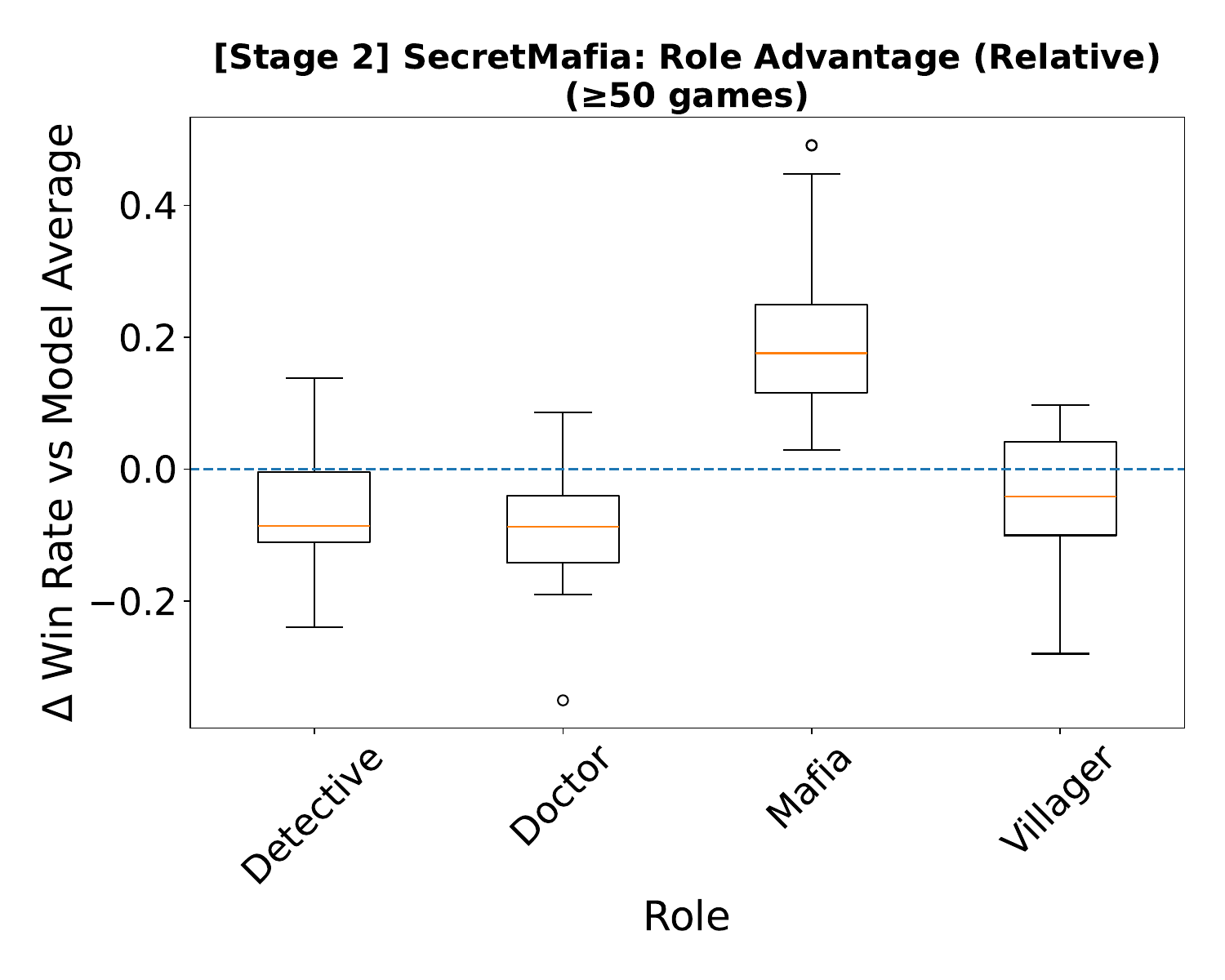}
    \caption{}
    \end{subfigure}
    \caption{Role Advantage across Codenames and Secret Mafia over the two stages of the competition. Boxplots show the distribution of win-rates across models for each role. The dashed horizontal line denotes the win rate of 50\%. In Codenames, win rates are quite identical, indicating minimal role bias, whereas in Secret Mafia, the "Mafia" role exhibits consistently higher win rates. Circular markers highlight statistical outliers beyond 1.5$\times$IQR.}
    \label{fig:role-advantage}
     \vspace{-1em}
\end{figure}

Despite this initial imbalance, Figure~\ref{fig:role-advantage} shows that role advantages in Codenames remain centered around zero for both stages, indicating that no role consistently provides a systematic advantage across models. This suggests that random role assignment is sufficient to ensure outcome fairness at scale, even when assignment frequencies are slightly uneven.

For Secret Mafia, the observed role distribution reflects the inherent structure of the environment. Each game $G$ contains two mafia roles and two villagers, leading to naturally higher frequencies for these roles. The slightly lower-than-expected frequency of villagers is due to dataset filtering, as players eliminated early in the game do not produce observable actions and are therefore absent from the released trajectories.

In contrast to Codenames, Secret Mafia exhibits a clear role-dependent advantage. As shown in Figure~\ref{fig:role-advantage}, the mafia role consistently yields positive relative win rates, while detective, doctor, and villager roles tend to underperform relative to model averages. This indicates that the mafia role carries an inherent strategic advantage rather than the observed differences arising purely from sampling noise.

Overall, these results highlight that while role assignment can be balanced through sufficient sampling, role-dependent advantages are environment-specific. For future evaluations, we therefore recommend reporting role-conditioned performance alongside aggregate metrics, and preserving role assignment in proportions consistent with the game mechanics.

Refer to Appendix~\ref{appendix:win-rate-adv} for the win rate distribution of each role.

\subsection{MG-Ref: An Offline Evaluation Protocol for Future Agents}
\label{subsec:reference}

To enable future agents to be evaluated against the 2025 cohort without depending on a live matchmaking server, we release \textbf{MG-Ref}, a deterministic offline tournament protocol that pairs a new agent $m$ against a frozen reference pool $\mathcal{M}_\mathrm{ref}$ of top-ranked Stage~II Efficient submissions. The protocol produces \mindgames{}-compatible TrueSkill, cumulative-reward, and error-attribution metrics. The selection criteria pair strong TrueSkill with low caused-error rates so that reference opponents are strategically informative and offline matchups remain reliable.

\paragraph{Selection.}
MG-Ref retains the top-ranked Stage~II Efficient qualifiers (Appendix~\ref{appendix: final rankings}) that pair strong TrueSkill with low caused-error rates. For the Generalization Track we include the three top-ranked teams. Across Stage~II, these three references caused zero fatal errors over their 1{,}061 combined Colonel Blotto and IPD games (Table~\ref{tab:error_by_model}); in the more error-prone Codenames environment, caused-error rates span 6\%--25\%, which tracks the game's intrinsic difficulty ceiling (Section~\ref{subsec:confounds}) rather than unusually weak reference play. For the Social Deduction Track we include the four top-ranked Efficient teams, whose self-forfeit rates range from 0\% to 9\% of Secret Mafia games, well below the 33\% rate observed for lower-ranked submissions such as Phoenix. Evaluating against this set controls for the error-survival regime identified in Section~\ref{subsec:confounds}: reference opponents are strategically informative rather than dominated by forfeit dynamics.

\begin{table}[h]
\centering
\small
\setlength{\tabcolsep}{5pt}
\caption{The MindGames Reference Set (MG-Ref). TrueSkill values are frozen Stage~II Efficient posteriors (Appendix~\ref{appendix: final rankings}); caused-error counts are drawn from Table~\ref{tab:error_by_model}. RLGaming's Generalization and Social Deduction entries correspond to two distinct team submissions and are used as separate references within their respective tracks.}
\label{tab:reference-set}
\begin{tabular}{llcl}
\toprule
Track & Team & Stage~II TrueSkill (CB / IPD / CN or SM) & Caused errors \\
\midrule
\multirow{3}{*}{Generalization}
& In2AI    & 36.8 / 39.3 / 26.6 & 0 / 98; 0 / 195; 4 / 69 \\
& STARS    & 31.8 / 25.1 / 23.5 & 0 / 105; 0 / 163; 17 / 69 \\
& RLGaming & 29.8 / 19.3 / 28.2 & 0 / 243; 0 / 257; 6 / 69 \\
\midrule
\multirow{4}{*}{Social Deduction}
& RLGaming & $27.2 \pm 2.8$ & self-forf.\ 4 / 130 (3.1\%) \\
& Tungsten & $23.8 \pm 2.8$ & self-forf.\ 5 / 132 (3.8\%) \\
& Odyssean & $18.4 \pm 2.9$ & self-forf.\ 0 / 106 (0.0\%) \\
& Phoebus  & $12.5 \pm 2.8$ & self-forf.\ 11 / 122 (9.0\%) \\
\bottomrule
\end{tabular}
\end{table}

\paragraph{Protocol.}
A participant model $m$ is scheduled against $\mathcal{M}_\mathrm{ref}$ under a balanced factorial design that meets or exceeds the Stage~II minimum-play floors: 30 Colonel Blotto games, 36 IPD games, 36 Codenames games balanced across all four team roles, and 96 Secret Mafia games balanced across duplication modes, seats, and the natural role distribution $(2,1,1,2)$ for (Mafia, Doctor, Detective, Villager). The per-environment game counts were chosen at levels where TrueSkill posteriors have substantially converged in the Stage~I and Stage~II time-series trajectories (Appendix~\ref{appendix:ts_timeseries}); further convergence analyses are shipped with the starter kit. References are instantiated with their frozen Stage~II TrueSkill posteriors and do not update during offline play, reproducing the Stage~II fix for rating farming (Appendix~\ref{appendix:stage2}); only $m$'s posterior changes. All five metrics from Section~\ref{subsec:confounds}, TrueSkill, win rate with 95\% Wilson intervals, cumulative reward $\sum_{G}\mathcal{R}_m(s_\mathrm{term})$, role-conditioned win rates, and the Caused / Witnessed / Self-Forf.\ / Opp-Forf.\ columns, are reported per environment and in aggregate. Full factorial design, seed allocation, reporting schema, and reference implementations live in Appendix~\ref{appendix:tournament} and the starter kit.

%% file: sections/6_discussion.tex
\section{Discussion}
\label{sec:discuss}

The NeurIPS 2025 competition validates \mindgames{} as a useful evaluation resource while also revealing important limits on what live multi-agent leaderboards currently measure. The central lesson is not that one aggregate score cleanly captures social or strategic reasoning, but that evaluation reliability depends strongly on the game environment and its failure dynamics.

\paragraph{Evaluation reliability differs sharply across environments.}
The environments included in the Generalization Track produce rankings of different interpretability. IPD recorded zero forfeited games across all participants, though this is partly enabled by the environment’s default fallback for invalid actions. Colonel Blotto is also relatively clean: the top three models caused zero game terminations, and the overall error rate remained low. Codenames presents an intermediate case. Non-forfeited rates among the top models range from 39\% to 45\%, and self-forfeit rates vary substantially (5.8--24.6\%; Table~\ref{table:top_model_specific_error}), indicating that rankings reflect a mixture of strategic skill and differential error avoidance. Secret Mafia presents a fundamentally different picture. As detailed in Section~\ref{subsec:confounds}, the 50.3\% game-level error rate, together with the fact that terminated games end after fewer than 3 turns on average (Figure~\ref{fig:failure_depth}), means that many high-ranked models benefited primarily from surviving games in which other agents failed. Table~\ref{table:top_model_specific_error} makes this especially clear: the top two Mafia models self-forfeited in only 4 and 5 games, respectively, but witnessed opponent errors in 129 and 131 games. In this environment, leaderboard position in the current cycle is therefore better interpreted as robustness to a failure-heavy interaction regime than as a direct measure of social deduction skill.

\paragraph{Signals of strategic reasoning remain visible, but only locally and unevenly.}
High error rates and metric confounds preclude strong claims about general social reasoning or generalized theory of mind. Even so, the subset of successful trajectories still contains meaningful evidence of nontrivial strategic behavior. Performance is clearly not driven by parameter count alone, as illustrated by the strong performance of In2AI’s 8B model in both the Efficient and Unlimited Agent divisions (Appendix~\ref{appendix:in2ai}). More broadly, the strongest submissions repeatedly relied on explicit engineering structure, including action-validity control, state maintenance, reward attribution, and modular reasoning pipelines, rather than raw model scale alone. This suggests that principled system design can yield strategic advantages that outweigh parameter count in structured interactive settings. 

More broadly, successful trajectories in Colonel Blotto and Codenames indicate that some agents can, at least locally, anticipate opponent allocations, exploit repeated interaction, and tailor semantic associations to particular allies. The strongest claim supported by the current cycle is therefore not that agentic LLMs have solved rich social reasoning, but that they exhibit partial and environment-dependent capabilities that become visible when the interaction mechanics are sufficiently reliable.

\paragraph{Implications for future evaluation.}
Taken together, the results suggest a hierarchy of benchmark reliability in the current instantiation. IPD and Colonel Blotto provide the most interpretable rankings. Codenames provides a meaningful but mixed signal, because strategic performance and error avoidance remain entangled. Secret Mafia rankings from this cycle should be treated as preliminary and robustness-dominated. This does not reduce the value of Secret Mafia as an environment. On the contrary, the difficulty of sustained deception, long-horizon state tracking, and deduction under uncertainty makes it an especially valuable stress test. But to serve as a strategically interpretable leaderboard, future cycles will need stronger action handling and more explicit reporting of evaluation confounds. At minimum, future benchmark releases should distinguish clean games from failure-affected games, report self-forfeit and opponent-forfeit rates alongside leaderboard scores, and treat metric choice as a substantive design decision rather than an implementation detail. 

To make these recommendations immediately actionable, we release MG-Ref and its offline tournament protocol (Section~\ref{subsec:reference}, Appendix~\ref{appendix:tournament}), which enforce the Stage~II minimum-play floors, balance opponent and seat exposure, and surface the full error-attribution breakdown alongside TrueSkill so that new agents can be compared to \mindgames{} Stage~II performers under the same measurement lens applied here. More broadly, the current cycle suggests that progress in multi-agent LLM evaluation will come not only from stronger models, but also from more reliable environments and sharper measurement protocols.

%% file: sections/9_conclusion.tex
\section{Conclusion}
\label{sec:conclusion}

We presented \mindgames{}, a live benchmark for evaluating agentic LLMs through repeated play in text-based multi-agent game environments, and reported the first large-scale evaluation cycle through the NeurIPS 2025 Challenge. Across 944 submitted agents and four environments, the results reveal recurring engineering patterns among top-performing systems: training-time adaptation is especially valuable under parameter constraints, inference-time structure remains highly competitive at frontier scale, data curation often matters more than raw volume, and added cognitive scaffolding can hurt when models are not trained to use it.

\vspace{-0.5em}
\paragraph{From competition to evaluation resource.}
Beyond the competition itself, the primary contribution of \mindgames{} is as an evaluation resource: a curated multi-game suite targeting complementary forms of social and strategic reasoning, a large-scale trajectory corpus for post-hoc analysis and training, the MindGames Reference Set (MG-Ref) of top Stage~II submissions paired with a deterministic offline tournament protocol, and a live evaluation interface that supports reproducible multi-agent benchmarking. The released dataset of $\mathcal{G} = 29{,}571$ games across four environments captures a wide range of strategies, failure modes, and interaction dynamics observed during the evaluation cycle, enabling future work on agent training, behavioral analysis, and benchmark design.

\vspace{-0.5em}
\paragraph{Measurement lessons from the first cycle.}
The first competition cycle also yields a more general lesson about evaluation itself. Live multi-agent leaderboards do not measure a single stable construct across environments. In cleaner settings such as IPD and, to a lesser extent, Colonel Blotto, rankings are relatively interpretable as signals of strategic performance. In Codenames, rankings mix strategic ability with rule adherence. In Secret Mafia, the current cycle exposed a pronounced error-survival confound, where robustness to opponent failures can dominate the leaderboard signal. This makes \mindgames{} valuable not only as a benchmark, but also as a platform for studying when benchmark scores reflect genuine strategic ability and when they are artifacts of the environment. Put differently, \mindgames{} is most useful not because it compresses multi-agent intelligence into a single score, but because it exposes when that score is meaningful and when it is not.

\vspace{-0.5em}
\paragraph{Practical evaluation protocol.}
The \mindgames{}
The benchmark is designed for minimal friction. Any system that accepts text observations and returns text actions can participate through a unified HTTP API, with no game-specific integration required. For stable Stage~II ratings, the online protocol requires 50 games in Secret Mafia for social deduction track and 30 games in each Generalization environment (Colonel Blotto, IPD, and Codenames). Participants can additionally run the offline MG-Ref tournament (Section~\ref{subsec:reference}, Appendix~\ref{appendix:tournament}) against the released reference set for 198 games per agent across the four environments, obtaining paper-compatible TrueSkill, cumulative-reward, and error-attribution metrics without depending on the live server. The TrueSkill trajectories from the NeurIPS cycle suggest that these budgets are sufficient for substantially reducing early uncertainty, while complete game trajectories $\mathcal{T}$ are logged automatically for downstream analysis of failure modes, strategic adaptation, and behavioral diversity.

\paragraph{Acknowledgments.}
We thank all competition participants for their creativity and perseverance, and the NeurIPS organizers for their support. We also thank our sponsors, Sentient Foundation, Mithril, and Modal for funding the computational costs and prizes of the competition.

%% file: sections/A_appendix_methods.tex
\clearpage
\section{Detailed Method Descriptions}
\label{appendix:methods}

This appendix provides extended technical details for each winning submission, supplementing the summaries in Section~\ref{sec:subsummary}.

\subsection{Generalization Track}

\subsubsection{Team In2AI (1st Efficient, 1st Unlimited)}
\label{appendix:in2ai}

Team In2AI's approach centers on addressing credit assignment challenges in multi-agent, multi-turn games through a principled episode lifecycle with delayed reward attribution and eligibility gating.

\paragraph{System architecture}
The training system consists of four main components:
\begin{enumerate}
    \item \textbf{Action Validator}: Enforces reasoning-template compliance, action-format constraints, and game-rule validity during gameplay. Episodes are terminated with typed failure metadata when violations occur, enabling downstream attribution of responsibility.
    \item \textbf{Players Builder}: Reconstructs outcomes after episode termination and computes granular episode-level rewards beyond binary win/loss (e.g., normalizing by fraction of rounds won in Colonel Blotto rather than match outcome alone), including responsibility attribution for premature termination.
    \item \textbf{Steps Filter}: Excludes training steps that lack observable outcomes --- for instance, a valid Codenames clue whose operative produced a parsing failure has no guesses to evaluate and is therefore gated. Crucially, error steps themselves remain eligible and receive penalty rewards to teach format compliance; only steps with no outcome to learn from are excluded.
    \item \textbf{Reward Assigner}: Performs environment-specific backward attribution so that logically coupled actions share credit or blame based on realized outcomes. Per-step rewards are additionally modulated by episode outcome: actions in winning games receive full credit regardless of intermediate results, while the same actions in losing games receive reduced credit, ensuring that locally suboptimal moves within a winning strategy are credited appropriately.
\end{enumerate}

\paragraph{Training infrastructure}
The team uses an asynchronous training stack with vLLM continuous batching, where each inference request completes independently rather than in synchronized batches, eliminating idle time from heterogeneous generation demands. Completed trajectories are buffered in an episode bank to decouple rollout collection from batch construction. Three samplers configure each new episode: an environment sampler (round-robin across games), a position sampler (uniform coverage of all roles and turn orders), and an opponent sampler (curriculum-based selection described below). Training batches are balanced across environments, episodes, and reward strata using stratified sampling.

\paragraph{Optimization details}
The team fine-tunes Qwen3-8B using RLOO-style policy gradient optimization adapted to multi-step settings via per-environment advantage normalization: since no two steps share the same prompt, advantages are computed relative to other steps from the same game type rather than the same prompt, preventing cross-game reward-scale interference. The KL divergence penalty is disabled (no reference model), allowing the policy to diverge freely from initialization. Notably, the same RL-trained 8B model was submitted to both the Efficient and Unlimited divisions, demonstrating that a well-engineered small model can match or exceed substantially larger proprietary systems. Final performance is improved through per-environment generation parameter tuning across temperature, top-p, top-k, and min-p, with different optimal configurations selected for each game.

\paragraph{Curriculum and opponent selection}
Opponent selection follows a two-phase curriculum using external models rather than self-play. The first phase trains exclusively against gpt-oss-120B configured with varied system prompts inducing different behavioral profiles (aggressive, cooperative, unpredictable, analytical), letting the model learn rules, formats, and basic strategies without facing opponents too strong for a novice agent. The second phase introduces frontier models (GPT-5, Gemini 2.5 Pro, Grok-4, Qwen3-235B) accessed via API with weighted sampling that reserves expensive models for periodic challenge. Phase 1 opponents remain in the pool throughout training to preserve generalization across skill levels.

\subsubsection{Team STARS (2nd Efficient)}
\label{appendix:stars}

Team STARS develops code-augmented agents using Qwen3-8B with Ollama for local inference, combining three techniques: Guided Generation, ReAct, and Program-Aided Language (PAL).

\paragraph{Three-technique architecture}
The system integrates three complementary approaches:
\begin{enumerate}
    \item \textbf{Guided Generation}: Uses Pydantic-based constrained decoding to enforce structured output with a dedicated reasoning field, ensuring outputs conform to game-specific schemas.
    \item \textbf{ReAct framework}: The model generates self-authored code blocks within its reasoning traces, executing them inline for computation and verification.
    \item \textbf{PAL (Program-Aided Language)}: Deterministic computation is offloaded to Python execution, avoiding the numerical errors inherent in token-level arithmetic.
\end{enumerate}

\paragraph{Zero-heuristic design principle}
A deliberate design choice: the system uses no hardcoded lookup tables, decision trees, or game-specific heuristics. All strategic decisions emerge from code generation and LLM reasoning. The agent follows a predefined questionnaire to analyze each game state before generating code-augmented actions.

\paragraph{Game-specific code generation}
For each game type, the generated code handles domain constraints:
\begin{itemize}
    \item \textbf{Colonel Blotto}: Optimization routines compute legal allocations, yielding a 22\% relative improvement over non-code baselines.
    \item \textbf{Three-player IPD}: Code tracks cooperation history and implements conditional strategies through explicit state machines.
    \item \textbf{Codenames}: Code-based constraint checking filters candidate clues against assassin and opponent words, reducing illegal clue rate.
\end{itemize}

\paragraph{Feedback loop}
A lesson-learned mechanism extracts post-game analysis from completed episodes and feeds it into future reasoning, enabling the agent to adapt across games without fine-tuning. Action evaluation via self-generated validation code verifies constraint compliance before committing actions. Overall, the system achieves a 15\% average win rate improvement across games.

\subsubsection{Team RLGaming (3rd Efficient, 2nd Unlimited)}
\label{appendix:rlgaming-gen}

Team RLGaming employs multi-source supervised fine-tuning on Llama-3.1-8B-Instruct with LoRA for the Efficient division, and GPT-5 with game environment-specific prompts for the Unlimited/Open division.

\paragraph{Three-source data collection}
Training data is collected from three complementary origins:
\begin{itemize}
    \item \textbf{Rule-based agents}: Nash equilibrium strategies for Colonel Blotto, defection strategies for IPD, providing coverage of game-theoretic baselines
    \item \textbf{RL-trained agents}: Population-based PPO agents for Blotto, introducing adaptive strategic variation
    \item \textbf{Proprietary model trajectories}: Games played by GPT-4o, GPT-4.1, and GPT-5 mini against varied opponents, providing high-quality strategic reasoning
\end{itemize}

\paragraph{Context optimization}
Observations are split by \texttt{[GAME]} phase blocks with duplicate information removed, reducing input length and improving context utilization. Prompting is used for formatting only; strategic behavior is trained via fine-tuning.

\paragraph{Open division}
The GPT-5 submission uses game environment-specific prompts, including deceptive strategies for IPD (e.g., building trust through early cooperation before defecting).

\paragraph{Observed limitations}
Fine-tuned agents showed vulnerability to overfitting: Blotto strategies converged to countering specific opponent allocation patterns, and IPD agents tended toward all-defect equilibria. Codenames proved least improvable via SFT, likely because the semantic reasoning required for clue generation resists pattern-level imitation.

\subsubsection{Team Odyssean (3rd Unlimited)}
\label{appendix:odyssean-gen}

Team Odyssean develops a three-stage training pipeline combining SFT with GRPO reinforcement learning, addressing three challenges: evaluation scalability, SFT data construction, and RL stability.

\paragraph{Stage 1: No-CoT SFT}
Initial fine-tuning on action-only supervision with rigorously curated data:
\begin{itemize}
    \item \textbf{Three-stage filtering}: (1)~TF-IDF deduplication at 0.95 cosine similarity, (2)~GPT-5 quality scoring on a 0--5 scale, (3)~strict retention of score-5 samples only
    \item Training data: 2{,}845 high-quality (observation, action) pairs across four game environments
    \item Objective: Establish game-rule compliance and basic action formatting
    \item LoRA rank 16, $\alpha=32$
\end{itemize}

\paragraph{Stage 2: CoT-SFT}
Chain-of-thought fine-tuning using teacher-generated reasoning:
\begin{itemize}
    \item Teacher model: Qwen3-235B-A22B generates structured reasoning traces
    \item Training data: 11{,}932 (observation, reasoning, action) triples
    \item Objective: Transfer strategic reasoning patterns from a 235B-parameter teacher to Qwen3-8B
\end{itemize}

\paragraph{Stage 3: GRPO refinement}
Group Relative Policy Optimization for online improvement:
\begin{itemize}
    \item Unified outcome reward $r \in \{-1, 0, 1\}$
    \item \textbf{Prompt pool}: Randomized system instructions to prevent format overfitting
    \item \textbf{Model pool}: Heterogeneous opponents from different training stages to avoid single-opponent overfitting
    \item Key challenge: single-opponent overfitting versus multi-opponent training instability
\end{itemize}

\paragraph{Deployment}
Efficient division uses qwen3-8B fine-tuned via the above pipeline; Open division uses GPT-5 directly. Secret Mafia was excluded from local evaluation due to high stochasticity in 6-player games.

\subsubsection{Team GB}
\label{appendix:team-gb}

Team GB develops a unified LLM-guided preference and distillation framework that trains a compact, non-LLM graph attention policy for multi-game strategic adaptation. Their final policy contains approximately 6.8 million parameters---orders of magnitude smaller than competing LLM-based agents.

\paragraph{Graph-based state encoding}
Game state is represented as a graph of 25--40 nodes. For Colonel Blotto, nodes correspond to battlefields with features including normalized troop allocations, round index, and cumulative score difference (feature dimension 32). For Codenames, nodes represent board words with features encoding remaining targets, valid historical clues, fraction of guessed words, and turn index (feature dimension 35). Three graph attention layers with 6 attention heads and hidden size 192 process state graphs into a fixed-dimensional embedding shared across games.

\paragraph{Policy architecture}
A single graph-based policy architecture operates across structurally different games. The policy consists of a common trunk, FiLM adaptation layers~\citep{perez2018film} conditioned on a 32-dimensional opponent embedding, and a portfolio action head over 6 latent strategies. Role-conditioned decoding is applied for spymaster and operative roles in Codenames.

\paragraph{Five-stage training curriculum}
Training proceeds through five coupled components:
\begin{enumerate}
    \item \textbf{Graph PPO training}: Clipped PPO ($\epsilon=0.2$, $\gamma=0.99$, $\lambda=0.95$) with auxiliary exploration and counterfactual updates.
    \item \textbf{Meta learning}: A bi-level update where a fast inner loop adapts FiLM parameters for 1--2 gradient steps to recent opponent behavior, while an outer loop optimizes for rapid adaptation.
    \item \textbf{Preference generation}: Two teacher LLMs (Qwen 2.5-Instruct and Llama 3-Instruct) propose candidate actions. Each candidate is evaluated via 4 stochastic rollouts, producing approximately 2{,}300 preference pairs.
    \item \textbf{Teacher alignment}: Supervised fine-tuning on chosen actions followed by direct preference optimization (DPO) on preference pairs to align a teacher model.
    \item \textbf{Policy distillation}: The aligned teacher generates state-to-action labels for 2{,}000 sampled states. The graph policy is trained by cross-entropy imitation, then continues PPO training to stabilize performance.
\end{enumerate}

\paragraph{Results}
On Colonel Blotto, the full curriculum attains a 78.40\% win rate (95\% CI: [77.36, 79.44]) over 1{,}000 games. PPO alone achieves 58.4\% with a 14.2\% collapse rate (allocating $>$60\% of units to a single field); distillation raises this to 67.9\% while reducing collapse to 8.8\%. On Codenames, distillation improves win rate from 44.8\% to 52.9\% while reducing assassin-triggered losses from 12.6\% to 6.9\%. Cross-agent evaluation against LLM opponents shows improvement from 46.2\% to 55.6\% after distillation. Preference learning primarily reduces failure probability rather than increasing raw win rate, acting as a risk regularizer under strategic uncertainty.

\subsection{Social Deduction Track}

\subsubsection{Team RLGaming (1st Efficient)}
\label{appendix:rlgaming-sd-eff}

Team RLGaming's Social Deduction Efficient submission uses a three-stage pipeline with Qwen3-8B.

\paragraph{Stage 1: Structured prompting with hard constraints}
A five-section prompt template:
\begin{enumerate}
    \item \textbf{Hard Constraints}: Prohibits identity leakage, repetition, and requires new reasoning each turn. Role-specific behavioral instructions guide strategy (e.g., Mafia agents are instructed to mislead without implicating teammates).
    \item \textbf{Game Message}: Current game state and available actions
    \item \textbf{Observation}: Filtered observation containing only system messages and player statements
    \item \textbf{Past Public Statements}: The agent's own previous public messages (for consistency)
    \item \textbf{Talk}: Space for generating the current response
\end{enumerate}

\paragraph{Stage 2: Memory and deduction layer}
\begin{itemize}
    \item \textbf{Observation preprocessing}: Regular expressions extract system-level information and player messages, removing formatting noise
    \item \textbf{Role-specific memory}: Dedicated memory per role stores utterances alongside running deductions about each player's likely identity
    \item \textbf{Two-task deduction prompt}: Task~1 compresses new statements into 1--3 sentences; Task~2 produces a one-line role hypothesis per player based on accumulated evidence
\end{itemize}

\paragraph{Stage 3: SFT on stronger-model traces}
\begin{itemize}
    \item Training data: 100+ games and 10{,}000+ training instances generated by gpt-oss-120B
    \item LoRA fine-tuning on Qwen3-8B
\end{itemize}

\paragraph{Ablation results}
A key finding: adding memory and deduction \emph{without} SFT actually hurts performance. Win rates by configuration: Base 21.7\% $\to$ +Prompt Refinement 25.0\% $\to$ +Memory/Deduction 16.7\% $\to$ +SFT 45.0\%. The memory layer introduces complexity that untrained models cannot exploit; SFT teaches the model to leverage structured memory effectively.

\subsubsection{Team Tungsten (2nd Efficient)}
\label{appendix:tungsten}

Team Tungsten uses inference-time optimization through structured prompting and private reasoning separation with Qwen3-8B, requiring no fine-tuning.

\paragraph{Three agent types}
The team developed three agent variants through an evolutionary design process:
\begin{enumerate}
    \item \textbf{Basic agent}: A single LLM call generates the response directly. Key failure: Mafia agents frequently leak role information into public responses.
    \item \textbf{Thinking agent}: The model generates a private reasoning block enclosed in \texttt{<outloud>} XML tags, containing role-aware strategic analysis. An external harness extracts only the public action portion, architecturally preventing information leakage.
    \item \textbf{Remembering agent}: Extends the Thinking agent with a \texttt{<remembering>} XML block for cross-turn knowledge persistence, where the LLM decides what to retain or discard.
\end{enumerate}

\paragraph{Key finding: memory without fine-tuning hurts}
Counterintuitively, the Remembering agent performed \emph{worse} than the Thinking agent. Without fine-tuning, the 8B-parameter model could not reliably exploit the structured memory, and the additional complexity degraded coherence. This parallels RLGaming's finding (Appendix~\ref{appendix:rlgaming-sd-eff}) that memory layers require SFT to be effective.

\paragraph{Prompt engineering}
Three prompt variants were tested: simplistic (8 tokens), handwritten (679 tokens with strategic guidance), and LLM-written (884 tokens, rewritten by Claude Sonnet 4). Qwen3-8B consistently outperformed Llama-3.1-8B across all configurations. Evaluation used Monte Carlo TrueSkill with 100 shuffles of 434 games.

\subsubsection{Team Odyssean (3rd Efficient)}
\label{appendix:odyssean-sd}

Team Odyssean applies their CoT-SFT methodology from the Generalization track to Social Deduction with Qwen3-8B.

\paragraph{CoT dataset construction}
\begin{itemize}
    \item Teacher model: Qwen3-235B-A22B generates structured reasoning traces
    \item Scenario diversity: Games against a heterogeneous opponent pool with varied capabilities
    \item Role coverage: Balanced sampling across Mafia, Doctor, and Villager roles
\end{itemize}

\paragraph{Fine-tuning details}
\begin{itemize}
    \item LoRA configuration: Rank 64, alpha 128
    \item Training: 3 epochs, learning rate 1e-5
    \item Batch size: 16 with gradient accumulation
\end{itemize}

\subsubsection{Team Phoebus (1st Unlimited)}
\label{appendix:phoebus}

Team Phoebus (team\_awu) develops a dual-agent architecture using GPT-5-mini that separates situation assessment from action selection.

\paragraph{Status Agent}
Processes the game history $H_t$, voting records $V_t$, and elimination events $E_t$ to produce:
\begin{itemize}
    \item Probabilistic role distributions $P(r_j \mid H_t)$ for each player $j$
    \item Confidence scores for each estimate
    \item Behavioral pattern detection (voting consistency, accusation targets, defense patterns)
    \item Game-state advantage assessment for the agent's faction
\end{itemize}

\paragraph{Action Agent}
Receives the Status Agent's full assessment along with the agent's assigned role, and reasons through a two-stage hierarchy:
\begin{enumerate}
    \item \textbf{Strategy formulation}: High-level objective selection (e.g., ``build coalition against Player~3'' or ``deflect suspicion from teammate'')
    \item \textbf{Tactic selection}: Specific action and dialogue that implements the chosen strategy
\end{enumerate}

\paragraph{Results across prompt configurations}
Three configurations reveal the impact of the multi-agent decomposition:
\begin{itemize}
    \item \textbf{Minimal prompts}: 15.0\% win rate (single-agent baseline)
    \item \textbf{Heuristic prompts}: 31.6\% win rate (hand-engineered strategy guidance)
    \item \textbf{Multi-agent architecture}: 70.0\% win rate (4.7$\times$ improvement over minimal)
\end{itemize}
Village-side performance improved 9$\times$ (8.3\% $\to$ 75.0\%); Mafia-side improved from 25\% to 62.5\%. The system produced zero invalid actions with GPT-5-mini. A temporal learning effect was also observed: the multi-agent system improved from 60\% to 80\% win rate over 20 consecutive games.

\subsubsection{Team CerebrAI (2nd Unlimited)}
\label{appendix:cerebrai}

Team CerebrAI implements a cognitive neuroscience-inspired architecture using GPT-5, drawing on constructs from social neuroscience research.

\paragraph{Exformation layer}
Before reasoning begins, an exformation step (inspired by Nørretranders' concept of selective omission) filters raw observations to extract only salient cues: voting patterns, accusation targets, defense statements, and behavioral inconsistencies. This reduces the cognitive load on downstream reasoning modules.

\paragraph{Three thinking modes}
The architecture organizes reasoning into three distinct modes:
\begin{enumerate}
    \item \textbf{Imaginative Thinking} (inspired by the Default Mode Network): Generates hypotheses about what players might do, feel, or plan. Handles perspective-taking, social simulation, and theory of mind.
    \item \textbf{Logical Thinking} (inspired by the Task-Positive Network): Tests imaginative hypotheses against behavioral evidence. Performs hypothesis verification, strategic planning, and vote optimization.
    \item \textbf{Deception Detection Thinking}: Identifies expectation-violations, where a player's statements or actions diverge from their predicted behavior pattern. Inspired by neuroscience research on humor and surprise processing, this mode flags inconsistencies as potential deception cues.
\end{enumerate}

\paragraph{Evolutionary development}
The team tested approximately 25 agent versions, evolving through distinct design philosophies: exformation (v14), persuasive reasoning (v14u), Dark Triad personality modeling (v16--17), cerebral analytical reasoning (v19--20), influence-based strategies (v21), and the final cerebral + deception detection combination (v25).

\paragraph{Results}
The final system achieved an 82\% win rate (41 wins, 9 losses) with TrueSkill $7.8 \pm 3.0$ and an average response time of 3.2 seconds.

\subsubsection{Team RLGaming (3rd Unlimited)}
\label{appendix:rlgaming-sd-unl}

Team RLGaming's Unlimited submission employs BDI reasoning with automated self-improvement, using gpt-5-nano.

\paragraph{Structured game observation}
Raw game messages are processed through two modules:
\begin{itemize}
    \item \textbf{Message Parser}: Temporal ordering and type classification of game messages (system announcements, player statements, vote results)
    \item \textbf{State Analyzer}: Extracts game phase, alive/dead player sets, and legal action targets
\end{itemize}

\paragraph{Hierarchical prompt architecture}
Three layers of context:
\begin{enumerate}
    \item \textbf{Global System Prompt}: Game rules, mechanics, and a JSON reply protocol that separates reasoning from public action
    \item \textbf{Role-Specific Strategy Guidance}: Goals and decision criteria tailored to the assigned role (not scripts, but principles)
    \item \textbf{Dynamic Game Context}: Compact state snapshot from the State Analyzer
\end{enumerate}

\paragraph{BDI reasoning scaffold}
\begin{itemize}
    \item \textbf{Beliefs}: Role probability estimates for each player, updated from behavioral evidence
    \item \textbf{Desires}: Current strategic goals derived from role and game state
    \item \textbf{Intentions}: Concrete action plans that implement the current strategy
\end{itemize}

\paragraph{Self-improvement via role heuristics}
An automated loop (Algorithm~1 in the original submission):
\begin{enumerate}
    \item Run self-play game batches
    \item Perform role-level post-hoc analysis of wins and losses
    \item Extract recurring heuristics from winning games (e.g., ``as Villager, vote with the majority in early rounds to build credibility'')
    \item Filter conflicting heuristics
    \item Integrate surviving heuristics into role-specific strategy guidance
\end{enumerate}

\paragraph{Results}
Progressive ablation shows cumulative gains: ReAct baseline 52\% $\to$ +Structured observation 62\% $\to$ +BDI reasoning 70\% $\to$ +Self-improvement 78\%. The largest gains accrue on the information-poor Villager side (16\% $\to$ 60\%), while Mafia performance remains consistently high (88--96\%) across all configurations, suggesting that BDI and self-improvement primarily help agents operating under uncertainty.

\subsubsection{Team DeceptionNet}
\label{appendix:deceptionnet}

Team DeceptionNet develops a hybrid imitation learning and reinforcement learning architecture with a strict separation of concerns: the LLM (Phi-3-mini-128k-instruct) is never fine-tuned but serves as a frozen feature extractor, while a separate trainable policy network handles all decision-making.

\paragraph{Hybrid Listener: dual-mode perception}
A two-tier perception system extracts signals from conversational game logs:
\begin{itemize}
    \item \textbf{Rule-based mode} ($\sim$2s latency): Three curated lexicons capture social dynamics---accusation cues (suspect, accuse, vote, eliminate), support cues (defend, clear, trust, ally), and sentiment markers. Structural analysis via regular expressions identifies player mentions, speaker-target relationships, voting-phase temporal patterns, and behavioral indicators (contradiction scores, bandwagon detection, mention frequency).
    \item \textbf{LLM-enhanced mode}: Phi-3-mini-4k-instruct (or Qwen-2.5-1.8B-Instruct), quantized to FP16, extracts structured JSON features (accusation relations, defense relations, role claims, speaker confidence, contradiction patterns). An LRU cache (size 16) avoids redundant computation. LLM features are \emph{blended} with rule-based outputs through additive augmentation, providing robustness when the LLM fails to parse.
\end{itemize}

\paragraph{BeliefNet: probabilistic hidden-state tracking}
Rather than feeding raw features directly to the policy, a learned Bayesian-filter-like module maintains structured beliefs:
\begin{itemize}
    \item \textbf{Per-player encoder}: 20 features per player (alive status, votes received/cast, social signals, behavioral patterns, role embedding) processed by a 2-layer MLP into 128-dimensional embeddings.
    \item \textbf{Global context encoder}: Game-level features (round number, alive ratio, estimated Mafia remaining, phase encoding) produce a 128-dimensional context vector.
    \item \textbf{Relational reasoning}: A 2-layer Transformer encoder (4 attention heads, 256-dim feedforward) captures cross-player patterns.
    \item \textbf{Temporal integration}: A GRU cell maintains memory across game rounds, tracking multi-round behavioral consistency.
    \item \textbf{Output heads}: Role Predictor (logits over 4 roles), Suspicion Scorer ($[0,1]$), Trust Scorer ($[0,1]$).
\end{itemize}

\paragraph{StateBuilder: graph-enhanced fusion}
Belief outputs are combined with raw features through four operations: feature fusion of player and conversation streams, cross-attention between conversation context and player representations, graph neural network message passing (2 hops) over a social graph constructed from voting patterns and mention co-occurrence, and GRU-based episodic memory.

\paragraph{Multi-head policy}
Specialized MLP heads handle different action types: Night actions (player embeddings + memory), Vote (player embeddings + conversation), Talk Intent (8 discrete intents: accuse, defend self, defend other, claim role, agree, question, filler, silent), Talk Target (alive players only), and a Value head (critic). Masked softmax ensures only valid actions receive a nonzero probability.

\paragraph{Presenter: action-to-language generation}
Two modes: (1)~Template-based (sub-second, used in competition), using pre-defined templates with dynamic slot-filling; (2)~LLM-based (Mistral-7B-Instruct, 30--50s latency), generating multi-sentence responses grounded in belief state summaries.

\paragraph{Training}
\begin{itemize}
    \item \textbf{Stage 1: Imitation learning.} 300 synthetic demonstrations generated from rule-based heuristics, trained via standard cross-entropy over all four action heads. Converges in 20--30 minutes on a single consumer GPU.
    \item \textbf{Stage 2: PPO fine-tuning (attempted, abandoned).} Self-play PPO proved too brittle: action drift (invalid talk or vote targets), reward ambiguity (most decisions lack short-term feedback), presenter--listener feedback loops (LLM leaks role information, creating artificial training signals), and self-play non-stationarity. Performance consistently degraded from the IL baseline. The competition submission relies entirely on the IL model.
\end{itemize}

\paragraph{Results}
Competitive performance with $\sim$5-second average response time versus 40+ seconds for LLM-heavy competitors. A configuration oversight caused the small LLM (Phi-3-mini-128k-instruct) to compete in the unconstrained category against significantly larger models, yet it remained competitive, demonstrating the robustness of explicit belief modeling. The authors conclude: ``less coupling can yield more robustness.''

\subsubsection{Team Revac (1st Open)}
\label{appendix:revac}

Team Revac develops a multi-module inference architecture that evolved through three major versions (Revac $\to$ Revac2\_1 $\to$ Revac\_8), achieving first place in the Open Division. No fine-tuning is used; the approach is purely architectural.

\paragraph{Core two-stage reasoning}
The foundational architecture separates analysis from action:
\begin{enumerate}
    \item \textbf{Reviewer Agent}: Takes the current game observation state (chat log, player status, game phase) and generates a detailed chain-of-thought review containing logical deductions, contradiction detection, and a probability assessment of each player's role. This internal review is not shown to other players.
    \item \textbf{Action Agent}: Takes the original observation state and the Reviewer's detailed analysis to formulate the final natural-language action (a statement, an accusation, or a vote). This separation ensures the final output is grounded in deep, structured analysis.
\end{enumerate}

\paragraph{Memory Module: Social Alignment Graph}
Introduced in Revac2\_1, the Memory Module overcomes short-term memory limitations through two components:
\begin{itemize}
    \item \textbf{Player Profiles}: Textual summaries of each player's behavioral history---claims made, votes cast, perceived consistency. Updated after each observation.
    \item \textbf{Social Alignment Graph (SAG)}: A dynamic, directed, weighted graph where players are nodes and edges represent social interactions. Edge types include: Accusation ($A \to B$, negative weight), Defense/Support ($A \to B$, positive weight), and Voting Alignment ($A \to B$, strong negative weight). The SAG enables relational inferences impossible with sequential text: collusion detection (mutual defense links persisting across rounds), group pressure dynamics (high in-degree of negative edges), and contradiction grounding (structured anchor against LLM hallucination).
\end{itemize}

\paragraph{Dynamic Tone Selector}
Added in Revac\_8, this third-stage module selects an optimal communication style based on the Reviewer's analysis and the current game state:
\begin{itemize}
    \item \textbf{Aggressive/Demanding}: Pressure suspects at critical moments (e.g., forcing a role claim at LyLo)
    \item \textbf{Withdrawing/Passive}: Deflect suspicion, used by Mafia after a failed lie or by Villagers to avoid becoming a target
    \item \textbf{Logically Anchoring}: Establish authority by summarizing confirmed evidence and proposing clear, rational votes
    \item \textbf{Contrarian/Skeptical}: Break potentially manufactured consensus when the SAG shows strong but suspicious group pressure
\end{itemize}

\paragraph{Evaluation}
The team developed a custom benchmark of 13 curated scenarios scored on two metrics: Role Identification Accuracy (proportion of correctly predicted roles, objective) and Reasoning Quality (LLM-judged logical soundness, 0--5 scale). Combined score: $0.5 \times \text{Metric A} + 0.5 \times \text{Metric B}_{\text{normalized}}$.

\paragraph{Results across versions and models}
Revac\_8 with gpt-5-mini achieves the best benchmark score of 0.80. Competition TrueSkill: Revac\_8 = 13.9 (1st), compared to CerebrAI = 7.8 (2nd). The version evolution shows cumulative architectural gains: Revac (two-stage reasoning, short-sighted decisions) $\to$ Revac2\_1 (+Memory Module and SAG, long-term strategic reasoning) $\to$ Revac\_8 (+Dynamic Tone Selector, persuasive communication and social camouflage). The team concludes that in social deduction, communication tone is itself a strategic action: ``A perfectly logical deduction, if delivered in a dry, robotic, or socially inappropriate manner, will fail to persuade.''

%% file: sections/B_appendix_observations.tex
\clearpage
\section{Observations and Action Spaces}
\label{app:observation}

\subsection{Colonel Blotto}
\begin{figure}[!h]
\centering
\begin{logbox}
\textbf{Turn 1}\hfill \texttt{2025-08-01 11:10:51Z}

\vspace{0.4em}
\textbf{Observation}
\begin{lstlisting}[style=mindgameslog]
[-1] You are Commander Alpha in a game of ColonelBlotto. Each round, you can allocate up to 20 units across fields: A, B, C
Format: '[A4 B2 C2]'
Win the majority of fields to win the round!
[-1] === COLONEL BLOTTO - Round 1/9 ===
Rounds Won - Commander Alpha: 0, Commander Beta: 0
Available fields: A, B, C
Units to allocate: 20
Format: '[A4 B2 C2]'.
\end{lstlisting}

\textbf{Action:} \texttt{[A7 B7 C6]}

\end{logbox}
\caption{Starting observation for Commander Alpha in Colonel Blotto.}
\label{fig:blotto-trace-full}
\end{figure}

\paragraph{Action space.}
Agents allocate a fixed number of units across battlefields A, B and C, using a structured action of the form:
\begin{center}
\texttt{[A$x$ B$y$ C$z$]}
\end{center}
where $x,y,z \in \mathbb{Z}_{\geq 0}$ and $x+y+z \leq U$, with $U$ the total available units. The agent wins the round by securing a majority of battlefields.

\vspace{1em}

\subsection{Three-Player Iterated Prisoner's Dilemma}
\begin{figure}[!h]
\centering
\begin{logbox}
\textbf{Turn 1}\hfill \texttt{2025-08-06 10:20:58Z}

\vspace{0.4em}
\textbf{Observation}
\begin{lstlisting}[style=mindgameslog]
[-1] You are Player 2 in a 3-player Iterated Prisoner's Dilemma. The match lasts 5 rounds.
Round structure:
  - 1 free-chat turns
  - 1 decision turn - submit one token per opponent: '[<opp-id> cooperate]' or '[<opp-id> defect]' (i.e. '[1 defect] [2 cooperate]'; the default is 'cooperate').
Pair-wise payoff matrix (applied to each unordered pair):
  - Both cooperate  ->  3
  - Both defect     ->  1
  - You defect, they cooperate -> 5
  - You cooperate, they defect -> 0
The player(s) with the highest score at the end of all rounds wins.

[-1] --- Starting Round 1 ---	You can converse freely for the next 1 rounds.
[0] [Trust and cooperation will benefit us all in the long run.]
[1] [Okay, I agree with Player 0. Let's try for cooperation early and see how the others react. My goal is to maximize my score while also trying to learn about the other players.]
\end{lstlisting}

\textbf{Action:} \texttt{[Player 2] [I agree with the plan to cooperate early...] }
\end{logbox}
\caption{Starting observation for Player 2 in Three-Player IPD.}
\label{fig:ipd-trace-full}
\end{figure}

\paragraph{Action space.}
The environment supports both communication and decision actions. Communication actions consist of unrestricted natural language messages. Decision actions require structured cooperation or defection choices for each opponent:
\begin{center}
\texttt{[$i$ cooperate]} \quad or \quad \texttt{[$i$ defect]}
\end{center}
for each opponent $i$.

\vspace{1em}

\subsection{Codenames}

\begin{figure}[!h]
\centering

\begin{logbox}
\textbf{Turn 1}\hfill \texttt{2025-08-04 15:23:25Z}

\vspace{0.4em}
\textbf{Observation}
\begin{lstlisting}[style=mindgameslog]
[-1] You are playing Codenames, a 2v2 word deduction game. Each team (Red and Blue) has a Spymaster and an Operative.
Rules:
1. The Spymaster gives a one-word clue + number (e.g., '[wind 2]') based on the team's secret words (the clue may not contain any of the words on the board).
2. The Operative guesses up to N+1 words (e.g., '[breeze]') based on the clue. They can also '[pass]'.
3. Avoid guessing opponent words, neutral words (N), or the Assassin (A), which causes instant loss.
4. First team to guess all their words wins.

You are Player 2, the Spymaster for Blue team. Give a one-word clue and number.
[-1] Spymaster of Red team, Player 0, submitted [metal 3].
[-1] Operator of Red team, Player 1, correctly guessed [brass].
[-1] Operator of Red team, Player 1, correctly guessed [plant].
[-1] Operator of Red team, Player 1, did not provide a valid guess. The teams turn will be skipped.
[-1] Codenames Words:
history  B
cruel    N
judge    N
soup     B
join     B
society  B
lift     R
plant    R revealed
brass    R revealed
son      N
desire   A
respect  N
example  R
part     R
price    B
man      N
knife    R
attack   R
ray      N
town     N
acid     B
look     R
shoe     B
cart     R
circle   B
\end{lstlisting}

\textbf{Action:} \texttt{[common 3]}
\end{logbox}
\caption{Starting observation for Player 2 in Codenames.}
\label{fig:codenames-trace-full}
\end{figure}

\paragraph{Action space.}
The action space depends on the agent's role. Spymasters provide a one-word clue and number where the number represents the number of codename words are related to:
\begin{center}
\texttt{[clue $n$]}
\end{center}
Operatives guess words or pass using:
\begin{center}
\texttt{[word]} \quad or \quad \texttt{[pass]}
\end{center}
These structured actions enable coordinated reasoning between teammates under partial information.

\vspace{1em}

\subsection{Secret Mafia}

\begin{figure}[!h]
\centering
\begin{logbox}
\textbf{Turn 1}\hfill \texttt{2025-10-24 11:15:12Z}

\vspace{0.4em}
\textbf{Observation}
\begin{lstlisting}[style=mindgameslog]
[-1] Welcome to Secret Mafia! You are Player 5.
Your role: Mafia
Team: Mafia
Description: A Mafia member. Eliminate villagers and gain majority.

Players: Player 0, Player 1, Player 2, Player 3, Player 4, Player 5

Your teammates are: Player 2, Player 5.

During DAY phase: Speak freely and vote.
During NIGHT phase: '[Player X]' to vote and eliminate a villager.
Win by eliminating villagers until Mafia equal or outnumber them.

[-1] Night has fallen. Mafia, agree on a victim.
Valid targets: [0], [1], [3], [4]
\end{lstlisting}

\textbf{Action:} \texttt{[Player 0]}
\end{logbox}
\caption{Starting observation for Player 5 in Secret Mafia.}
\label{fig:mafia-trace-full}
\end{figure}

\paragraph{Action space.}
Agents produce both communication and structured elimination actions depending on phase and role. Discussion phases allow unrestricted natural language communication. Voting and elimination actions follow the structured format:
\begin{center}
\texttt{[$i$]}
\end{center}
where $i$ is the selected target player. Mafia agents coordinate to eliminate villagers, while villagers vote to identify suspected Mafia members.

\clearpage
\subsection{Data Schema}
\label{appendix:data-schema}

Each environment in the released dataset (Section~\ref{sec:dataset}) is stored in a tabular format where each row corresponds to a player-trajectory for a single game trajectory. Table~\ref{tab:schema-codenames} provides the schema for all game environments.

\begin{table}[h]
\centering
\caption{Schema for all game environments. Each row corresponds to a player-specific record for a single game trajectory.}
\begin{tabular}{llp{0.62\linewidth}}
\toprule
Column & Type & Description \\
\midrule
\texttt{player\_game\_id} & int & Unique identifier for the player-episode record (typically derived from \texttt{game\_id} and \texttt{player\_id}). \\
\texttt{game\_id} & int & Unique identifier for the game episode (shared across all players in the match). \\
\texttt{env\_name} & str & Environment identifier (e.g., \texttt{Codenames-v0}). \\
\texttt{model\_name} & str & Model identifier of the agent assigned to this player for this game. \\
\texttt{player\_id} & int & Player index within the episode (e.g., 0--5). \\
\texttt{opponent\_names} & json & Mapping from other player indices to their assigned model identifiers. \\
\texttt{rewards} & json & Terminal reward mapping from each player index to its final reward. \\
\texttt{observations} & json & Turn-indexed logs of this player's observation text and emitted action. \\
\texttt{num\_turns} & int & Number of turns taken by this player in the episode. \\
\texttt{status} & str & Terminal episode status (e.g., \texttt{finished}). \\
\texttt{reason} & str & Explanation for the terminal status (e.g., normal termination or rule violation). \\
\bottomrule
\end{tabular}
\vspace{1em}
\label{tab:schema-codenames}
\end{table}

\section{Invalid Action Handling}
\label{app:invalid_action}

Table~\ref{tab:invalid_actions} summarizes how each environment handles invalid actions. Invalid actions are defined as model responses that fail to comply with game instructions (e.g., missing actions, incorrect output formats, or rule violations). Timeouts are not treated as invalid actions; instead, such game trajectories are discarded and not logged. The distinction between \emph{fatal} and \emph{non-fatal} errors is central to interpreting the error analysis in Section~\ref{subsec:confounds}: a fatal error terminates the game (or eliminates the player), while a non-fatal error is retried or auto-corrected without ending play. Importantly, the environment-level handling rules below are broader than the subset of errors that appear in the released confound tables. In particular, for Codenames, all \emph{recorded} errors in the released dataset are fatal illegal clues, which is why our confound analysis treats recorded Codenames errors as fatal.

For clarity, all main-text analyses that compare Codenames errors with forfeits refer to \emph{recorded} errors in the released dataset; in this cycle, those recorded errors are all fatal illegal clues.

Table~\ref{tab:failure_mode} breaks down the composition of failures within each game environment across both competition stages. Three mechanistically distinct failure modes are observed: \textit{format/structure errors}, where the model's output could not be parsed by the game engine regardless of its strategic content; \textit{rule violations}, where the output was syntactically valid but violated an explicit game constraint; and \textit{turn limit reached}, where neither team achieved a decisive outcome within the allotted move budget. Notably, IPD produces no failures in either stage, reflecting its auto-correction mechanism described in Table~\ref{tab:invalid_actions}. The remaining three environments each exhibit a dominant failure mode --- formatting errors in Colonel Blotto and Secret Mafia, and rule violations in Codenames --- though their relative prevalence shifts between stages. In particular, minority failure modes such as \textit{turn limit} disappear entirely in Stage II, and the share of format errors in Colonel Blotto grows from 62.2\% to 89.6\%, suggesting that while Stage II models became more rule-compliant, output formatting remained a persistent challenge in structured action spaces.

\begin{table}[H]
\centering
\footnotesize
\raggedright
\renewcommand{\arraystretch}{1.3}
\setlength{\tabcolsep}{4pt}
\caption{Invalid action handling across environments. Fatal errors cause game termination or player elimination; non-fatal errors are retried or auto-corrected.}
\begin{adjustbox}{max width=\textwidth}
\begin{tabular}{l p{4.5cm} p{3cm} p{3.5cm}}
\toprule
\textbf{Environment} & \textbf{Error Types} & \textbf{Retry Policy} & \textbf{Consequence} \\
\midrule

Colonel Blotto
& All invalid actions (format errors, illegal allocations) are fatal if consecutive.
& One retry allowed per action.
& Two consecutive invalid actions $\rightarrow$ loss (self-forfeit). Non-consecutive errors do not accumulate. \\

Three-Player IPD
& All invalid actions are non-fatal.
& No retry; auto-corrected.
& Invalid action replaced with default \texttt{[cooperate]}. No termination possible. \\

Codenames
& \textbf{Fatal}: illegal clues (e.g., revealing a secret word) are fatal. \textbf{Non-fatal}: other malformed actions may be skipped without terminating the game. In the released dataset for this competition cycle, all recorded Codenames errors are fatal illegal clues.
& No retry.
& Illegal clue $\rightarrow$ immediate game termination (self-forfeit). Non-fatal malformed actions result in a skipped turn. \\ 

Secret Mafia
& \textbf{Fatal}: invalid actions during the voting phase. \textbf{Non-fatal}: invalid Detective or Doctor actions.
& One retry per fatal action.
& Two consecutive fatal errors $\rightarrow$ player eliminated; game continues with remaining players. Non-fatal errors are silently corrected. \\

\bottomrule
\end{tabular}
\end{adjustbox}
\vspace{0.5em}
\label{tab:invalid_actions}
\end{table}

\begin{table}[!ht]
\centering
\caption{Failure mode composition across game environments and competition stages. 
Values are percentages of bad episodes within each environment and stage. 
\textbf{Bold} indicates the dominant failure mode per row.
\textit{Format/Structure}: output unparseable by the engine; 
\textit{Rule Violation}: parseable but illegal action; 
\textit{Turn Limit}: no winner within move budget.}
\label{tab:failure_mode}
\small
\begin{adjustbox}{max width=\textwidth}
\begin{tabular}{ll cccc cccc}
\toprule
& & \multicolumn{2}{c}{\textbf{Col. Blotto}} 
  & \multicolumn{2}{c}{\textbf{Codenames}} 
  & \multicolumn{2}{c}{\textbf{IPD}} 
  & \multicolumn{2}{c}{\textbf{Secret Mafia}} \\
\cmidrule(lr){3-4}\cmidrule(lr){5-6}\cmidrule(lr){7-8}\cmidrule(lr){9-10}
\textbf{Failure Category} & \textbf{Nature} 
  & I(\%) & II(\%) & I(\%) & II(\%) & I(\%) & II(\%) & I(\%) & II(\%) \\
\midrule
Format / Structure Error & Output unparseable    & \textbf{62.2} & \textbf{89.6} & ---  & ---    & --- & --- & \textbf{82.6} & \textbf{79.7} \\
Rule Violation (Content) & Parseable but illegal & 37.8          & 10.4          & \textbf{94.0} & \textbf{100.0} & --- & --- & 17.3 & 20.3 \\
Turn Limit Reached       & No winner in budget   & ---           & ---           & 6.0  & ---    & --- & --- & ---  & ---  \\
Other / Unknown          & Unclassified          & ---           & ---           & ---  & ---    & --- & --- & 0.1  & ---  \\
\bottomrule
\end{tabular}
\end{adjustbox}
\end{table}

\section{Detailed Error Statistics}
\label{app:detailed_errors}

Tables~\ref{tab:error_by_stage} and~\ref{tab:error_by_model} provide the data underlying the confound analysis in Section~\ref{subsec:confounds}. Table~\ref{tab:error_by_stage} reports aggregate statistics by environment and stage. Table~\ref{tab:error_by_model} extends Table~\ref{table:top_model_specific_error} to the top-5 ranked models, using the same column definitions: \textit{Clean} (no errors from any player), \textit{Caused} (games in which this model committed at least one error, fatal or non-fatal), and \textit{Witnessed} (games in which an opponent committed at least one error). For IPD, Colonel Blotto, and Codenames all recorded errors are fatal, so error-free and non-forfeited coincide; for Secret Mafia they diverge substantially due to non-fatal errors (see Table~\ref{tab:invalid_actions}).

\begin{table}[H]
\centering
\small
\setlength{\tabcolsep}{4pt}
\renewcommand{\arraystretch}{1.05}
\caption{Aggregate error statistics by environment and stage. Clean~(\%): games with zero errors. Error~(\%): percentage of games containing at least one error.}
\label{tab:error_by_stage}
\begin{adjustbox}{max width=\textwidth}
\begin{tabular}{llrrrrll}
\toprule
Environment & Stage & Models & Games & Clean (\%) & Error (\%) & Top Failure & 2nd Failure \\
\midrule
Colonel Blotto & I  & 232 & 7{,}828 & 83.5 & 16.5 & Invalid input (815) & Illegal units (479) \\
Colonel Blotto & II & 47  & 1{,}491 & 91.5 & 8.5  & Invalid input (114) & Illegal units (11) \\
\midrule
Codenames & I  & 177 & 5{,}603 & 14.6 & 85.4 & Illegal clue (4{,}583) & Turn limit (200) \\
Codenames & II & 52  & 879     & 61.4 & 38.6 & Illegal clue (339) & --- \\
\midrule
IPD & I  & 171 & 6{,}544 & 100.0 & 0.0 & --- & --- \\
IPD & II & 41  & 1{,}012 & 100.0 & 0.0 & --- & --- \\
\midrule
Secret Mafia & I  & 399 & 5{,}332 & 22.1 & 77.9 & Invalid move (3{,}863) & Protection (486) \\
Secret Mafia & II & 55  & 882     & 49.7 & 50.3 & Invalid move (399) & Protection (45)  \\
\bottomrule
\end{tabular}
\end{adjustbox}
\end{table}

Error rates improved substantially from Stage~I to Stage~II across all environments with errors: Codenames dropped from 85.4\% to 38.6\%, Colonel Blotto from 16.5\% to 8.5\%, and Secret Mafia from 77.9\% to 50.3\%. IPD maintained 0\% throughout. In Secret Mafia, the error-free rate rose from 22.1\% to 49.7\%.

\begin{table}[H]
\centering
\footnotesize
\setlength{\tabcolsep}{3pt}
\renewcommand{\arraystretch}{1.05}
\caption{Per-model error breakdown for top-5 ranked Stage~II Efficient Agent models. \textit{Clean}: games with no errors from any player. \textit{Caused}: games with at least one error by this model (fatal or non-fatal). \textit{Witnessed}: games with at least one opponent error. \textit{Self-Forf.}: games terminated by this model's fatal error. \textit{Opp-Forf.}: games terminated by an opponent's fatal error. For Colonel Blotto and Codenames all errors are fatal, so Caused~$=$~Self-Forf. For Secret Mafia, most errors are non-fatal. Categories are not mutually exclusive. Generalization Track models are ranked by overall standing; Secret Mafia by Social Deduction Track standing.}
\label{tab:error_by_model}
\begin{tabular}{llrlrrrrl}
\toprule
Game & Rank & Model & Games & Clean & Caused & Witnessed & Self-Forf. & Opp-Forf. \\
\midrule
\multirow{5}{*}{Blotto}
 & 1 & In2AI     & 98  & 97  & 0  & 1   & 0  & 1  \\
 & 2 & STARS     & 105 & 103 & 0  & 2   & 0  & 2  \\
 & 3 & RLG-8B    & 243 & 240 & 0  & 3   & 0  & 3  \\
 & 4 & Corleone  & 183 & 112 & 0  & 71  & 0  & 71 \\
 & 5 & Odyssean  & 79  & 2   & 77 & 0   & 77 & 0  \\
\midrule
\multirow{5}{*}{Codenames}
 & 1 & In2AI     & 69  & 31 & 4  & 34 & 4  & 34 \\
 & 2 & STARS     & 69  & 28 & 17 & 24 & 17 & 24 \\
 & 3 & RLG-8B    & 69  & 27 & 6  & 36 & 6  & 36 \\
 & 4 & Corleone  & 45  & 22 & 4  & 19 & 4  & 19 \\
 & 5 & Odyssean  & 44  & 16 & 12 & 16 & 12 & 16 \\
\midrule
\multirow{5}{*}{IPD}
 & 1 & In2AI     & 195 & 195 & 0 & 0 & 0 & 0 \\
 & 2 & STARS     & 163 & 163 & 0 & 0 & 0 & 0 \\
 & 3 & RLG-8B    & 257 & 257 & 0 & 0 & 0 & 0 \\
 & 4 & Corleone  & 139 & 139 & 0 & 0 & 0 & 0 \\
 & 5 & Odyssean  & 152 & 152 & 0 & 0 & 0 & 0 \\
\midrule
\multirow{5}{*}{\makecell[l]{Secret\\Mafia}}
 & 1 & RLGame    & 130 & 1 & 4   & 129 & 4  & 61 \\
 & 2 & Tungsten  & 132 & 1 & 10  & 131 & 5  & 63 \\
 & 3 & Odyssean  & 106 & 1 & 0   & 105 & 0  & 46 \\
 & 4 & Revac     & 122 & 1 & 11  & 121 & 11 & 58 \\
 & 5 & Phoenix   & 132 & 1 & 111 & 92  & 43 & 53 \\
\bottomrule
\end{tabular}
\end{table}

%% file: sections/C_appendix_rankings.tex
\clearpage
\section{Final Rankings}\label{appendix: final rankings}
\vspace{0.5em}
\renewcommand{\arraystretch}{1.05}
We present the models' final rankings in Stage II, reporting their TrueSkill (TS), number of games (G), win rate (W\%), and qualification status (Q) in Tables~\ref{tab:blotto}, \ref{tab:ipd}, \ref{tab:codenames}, and \ref{tab:mafia}. Refer to Appendix~\ref{appendix:protocol} for more information on qualification requirements.

\textbf{Disclaimer.} All statistics reported in this paper are derived from the publicly released dataset. Due to operational factors in the competition infrastructure, a small number of games played during the competition may not appear in the public dataset. As a result, game counts and derived metrics may differ slightly from those observed on the live leaderboard during the competition. The released dataset represents the authoritative record for reproducibility.
\newcommand{\xmark}{\textcolor{gray}{\ding{55}}}
\newcommand{\qualified}{\cmark}
\newcommand{\notqualified}{\xmark}

\newcommand{\modelname}[1]{{\scriptsize\ttfamily\seqsplit{#1}}}


\begin{table}[H]
\caption{Colonel Blotto rankings  (qualification requires $\geq30$ games).}
\centering
\footnotesize
\begin{tabular}{l c >{\raggedright\arraybackslash}p{4.8cm} l r c c c}
\toprule
\textbf{Div} & \textbf{R} & \textbf{Model} & \textbf{Team}
& \textbf{TS} & \textbf{G} & \textbf{W\%} & \textbf{Q} \\
\midrule

\multirow{5}{*}{Eff}
&1&\modelname{In2AI\_model}&In2AI&$36.8\pm1.1$&98&86.7&\qualified\\
&2&\modelname{STARS~Agent~Track2~V7}&STARS&$31.8\pm0.9$&105&61.9&\qualified\\
&3&\modelname{RLG-Model8B-Ver12}&RLGaming&$29.8\pm0.9$&243&48.1&\qualified\\
&4&\modelname{Corleone\_Reflextion}&Corleone&$17.4\pm1.2$&183&39.9&\qualified\\
&5&\modelname{Odyssean\_Generalization}&Odyssean&$5.5\pm1.5$&79&1.3&\qualified\\

\midrule
\multirow{10}{*}{Unlim}
&1&\modelname{RLG-Stage2-Heavy}&RLGaming&$46.0\pm0.9$&162&77.8&\qualified\\
&2&\modelname{In2AI/mindgames-in2ai-submission-m-v3}&In2AI&$44.5\pm2.2$&31&96.8&\qualified\\
&3&\modelname{In2AI/mindgames-in2ai-submission-temperature-1.0}&In2AI&$42.0\pm1.5$&64&82.8&\qualified\\
&4&\modelname{In2AI/mindgames-in2ai-submission-stage-7-step-30-temperature-1.0}&In2AI&$41.6\pm1.3$&95&86.3&\qualified\\
&5&\modelname{In2AI/mindgames-in2ai-submission-stage-7-step-30-temperature-0.6}&In2AI&$41.3\pm1.6$&57&87.7&\qualified\\
&6&\modelname{In2AI/mindgames-in2ai-submission-temperature-0.6}&In2AI&$40.9\pm1.5$&64&89.1&\qualified\\
&7&\modelname{In2AI/mindgames-in2ai-submission-stage-7-step-30-mix-v1}&In2AI&$40.0\pm1.6$&43&74.4&\qualified\\
&8&\modelname{BanMaHeavy}&Odyssean&$40.0\pm1.8$&30&83.3&\qualified\\
&9&\modelname{BanMaKimi}&Odyssean&$39.9\pm1.7$&18&44.4&\notqualified\\
&10&\modelname{In2AI/mindgames-in2ai-submission-stage-7-step-45-temperature-1.0}&In2AI&$39.8\pm1.2$&82&67.1&\qualified\\

\bottomrule
\end{tabular}
\vspace{1em}
\label{tab:blotto}
\end{table}

\begin{table}[H]
\centering
\footnotesize
\caption{Three-Player IPD rankings  (qualification requires $\geq30$ games).}
\begin{tabular}{l c >{\raggedright\arraybackslash}p{4.8cm} l r c c c}
\toprule
\textbf{Div} & \textbf{R} & \textbf{Model} & \textbf{Team}
& \textbf{TS} & \textbf{G} & \textbf{W\%} & \textbf{Q} \\
\midrule

\multirow{5}{*}{Eff}
&1&\modelname{In2AI\_model}&In2AI&$39.3\pm1.2$&195&97.9&\qualified\\
&2&\modelname{Corleone\_Reflextion}&Corleone&$29.0\pm0.9$&139&47.5&\qualified\\
&3&\modelname{STARS~Agent~Track2~V7}&STARS&$25.1\pm0.8$&163&17.8&\qualified\\
&4&\modelname{RLG-Model8B-Ver12}&RLGaming&$19.3\pm0.8$&257&1.9&\qualified\\
&5&\modelname{Odyssean\_Generalization}&Odyssean&$18.9\pm0.7$&152&7.2&\qualified\\

\midrule
\multirow{10}{*}{Unlim}
&1&\modelname{In2AI/mindgames-in2ai-submission-m-v3}&In2AI&$39.5\pm1.6$&32&93.8&\qualified\\
&2&\modelname{In2AI/mindgames-in2ai-submission-temperature-0.8}&In2AI&$36.7\pm1.7$&19&89.5&\notqualified\\
&3&\modelname{RLG-Stage2-Heavy-v2}&RLGaming&$35.9\pm1.3$&42&85.7&\qualified\\
&4&\modelname{In2AI/mindgames-in2ai-submission-stage-7-step-30-mix-v1}&In2AI&$35.6\pm2.6$&7&85.7&\notqualified\\
&5&\modelname{In2AI/mindgames-in2ai-submission-stage-7-step-45-mix-v1}&In2AI&$35.4\pm2.1$&13&84.6&\notqualified\\
&6&\modelname{In2AI/mindgames-in2ai-submission-mix-v1}&In2AI&$35.3\pm1.2$&38&81.6&\qualified\\
&7&\modelname{RLG-Stage2-Heavy}&RLGaming&$34.9\pm1.0$&54&68.5&\qualified\\
&8&\modelname{BanMaHeavy}&Odyssean&$34.8\pm0.9$&70&85.7&\qualified\\
&9&\modelname{In2AI/mindgames-in2ai-submission-stage-7-step-30-temperature-0.8}&In2AI&$34.7\pm1.5$&26&80.8&\notqualified\\
&10&\modelname{GameSwitchingAgent~-~Track2~\#1}&Odyssean&$34.0\pm5.7$&1&100.0&\notqualified\\

\bottomrule
\end{tabular}
\vspace{1em}
\label{tab:ipd}
\end{table}

\begin{table}[H]
\caption{Codenames rankings  (qualification requires $\geq30$ games).}
\centering
\footnotesize
\begin{tabular}{l c >{\raggedright\arraybackslash}p{4.8cm} l r c c c}
\toprule
\textbf{Div} & \textbf{R} & \textbf{Model} & \textbf{Team}
& \textbf{TS} & \textbf{G} & \textbf{W\%} & \textbf{Q} \\
\midrule

\multirow{5}{*}{Eff}
&1&\modelname{RLG-Model8B-Ver12}&RLGaming&$28.2\pm1.6$&69&58.0&\qualified\\
&2&\modelname{Corleone\_Reflextion}&Corleone&$26.7\pm2.0$&45&51.1&\qualified\\
&3&\modelname{In2AI\_model}&In2AI&$26.6\pm1.5$&69&56.5&\qualified\\
&4&\modelname{Odyssean\_Generalization}&Odyssean&$25.5\pm1.8$&44&40.9&\qualified\\
&5&\modelname{STARS~Agent~Track2~V7}&STARS&$23.5\pm1.5$&69&40.6&\qualified\\

\midrule
\multirow{10}{*}{Unlim}
&1&\modelname{GameSwitchingAgent~-~Track2~\#1}&Odyssean&$31.4\pm6.7$&1&100.0&\notqualified\\
&2&\modelname{General~simple}&Odyssean&$31.2\pm4.6$&5&80.0&\notqualified\\
&3&\modelname{General~v4}&Odyssean&$31.2\pm4.7$&4&75.0&\notqualified\\
&4&\modelname{In2AI/mindgames-in2ai-submission-stage-7-step-45-temperature-0.8}&In2AI&$30.8\pm6.8$&1&100.0&\notqualified\\
&5&\modelname{General~v7}&Odyssean&$30.7\pm4.5$&5&80.0&\notqualified\\
&6&\modelname{General~simple~v2}&Odyssean&$30.7\pm6.0$&3&66.7&\notqualified\\
&7&\modelname{RLG-Stage2-Heavy}&RLGaming&$30.4\pm1.4$&75&68.0&\qualified\\
&8&\modelname{In2AI/mindgames-in2ai-submission-m-v3}&In2AI&$29.9\pm1.5$&53&64.2&\qualified\\
&9&\modelname{In2AI/mindgames-in2ai-submission-stage-7-step-45-mix-v1}&In2AI&$29.7\pm2.8$&14&64.3&\notqualified\\
&10&\modelname{In2AI/mindgames-in2ai-submission-stage-7-step-30-temperature-0.8}&In2AI&$29.6\pm2.7$&19&68.4&\notqualified\\

\bottomrule
\end{tabular}
\vspace{1em}

\label{tab:codenames}
\end{table}

\begin{table}[H]
\centering
\caption{Secret Mafia rankings (qualification requires $\geq50$ games).}
\footnotesize

\begin{tabular}{l c >{\raggedright\arraybackslash}p{4.2cm} >{\raggedright\arraybackslash}p{2.2cm} r c c c}
\toprule
\textbf{Div} & \textbf{R} & \textbf{Model} & \textbf{Team}
& \textbf{TS} & \textbf{G} & \textbf{W\%} & \textbf{Q} \\
\midrule

\multirow{7}{*}{Eff}
&1&\modelname{RLGame-ts-v7}&RLGaming&$27.2\pm2.8$&130&73.1&\qualified\\
&2&\modelname{tungsten\_social\_v2}&tungsten&$23.8\pm2.8$&132&64.4&\qualified\\
&3&\modelname{Odyssean\_Social2}&Odyssean&$18.4\pm2.9$&106&46.2&\qualified\\
&4&\modelname{Revac-online}&Phoebus&$12.5\pm2.8$&122&48.4&\qualified\\
&5&\modelname{OrderOfPhoenix/jarvis\_v6}&Order of The Phoenix&$6.1\pm2.8$&132&13.6&\qualified\\
&6&\modelname{Vito-1.0}&Corleone&$5.4\pm3.8$&79&32.9&\qualified\\
&7&\modelname{ZeroR-SecretMafia-Efficient-v4}&ZeroR&$5.2\pm3.0$&97&20.6&\qualified\\

\midrule
\multirow{10}{*}{Unlim}
&1 & \modelname{Revac\_8} & Phoebus & 13.9 $\pm$3.3 & 52 & 82.7\% &\qualified\\
&2 & \modelname{Fractal\_SecretMafia\_Agent\_round2\_v25} & CerebrAI & 7.8 $\pm$3.0 & 50 & 82.0\%&\qualified\\
&3 & \modelname{Fractal\_SecretMafia\_Agent\_round2\_v14u} & CerebrAI & 4.7 $\pm$2.8 & 50 & 66.0\% &\qualified\\
&4 & \modelname{RLGame-stage2} & RLGaming & 3.2 $\pm$2.6 & 52 & 75.0\% &\qualified\\
&5 & \modelname{awu} v2 & Awu & 3.1 $\pm$2.4 & 59 & 67.8\% &\qualified \\
&6 & \modelname{RLGame-stage2-v2} & RLGaming & 1.9 $\pm$3.0 & 52 & 69.2\% & \qualified\\
&7 & \modelname{Revac2.1} & Phoebus & 1.7 $\pm$2.2 & 85 & 62.4\% &\qualified\\
&8 & \modelname{awu\_v1} & Awu & -0.1 $\pm$2.8 & 53 & 77.4\%&\qualified\\ 
&9 & \modelname{gemini-2.5-stage-2} & reflection & -3.4 $\pm$2.1 & 74 & 55.4\%&\qualified\\
&10 & \modelname{openai/gpt-oss-120b-4} & Individual & -3.7 $\pm$1.3 & 538 & 50.6\%&\qualified\\


\bottomrule
\end{tabular}
\vspace{1em}
\label{tab:mafia}
\end{table}

%% file: sections/D_appendix_naming.tex
\clearpage
\section{Model Naming}\label{appendix:full_model_naming}
Due to space constraints in tables, model identifiers are abbreviated in the main text. This appendix provides the full identifiers and defines the abbreviation scheme used throughout the paper. All abbreviated names uniquely map to a single model instance.

Abbreviations follow the format: 
\[
\texttt{<Family>[-S<stage>][-St<step>][-T<temperature>][-Variant]}
\]
Where:
\begin{enumerate}
    \item Family denotes the model team (e.g., In2AI, RLG).
    \item Sx denotes training stage x.
    \item Stx denotes training step x.
    \item Tx.x denotes decoding temperature.
    \item Variant denotes additional configuration tags (e.g., Mix, v2).
\end{enumerate}

\begin{table}[H]
\centering
\small
\setlength{\tabcolsep}{4pt}
\renewcommand{\arraystretch}{0.95}
\caption{Mapping between abbreviated model names used in tables and full model identifiers.}
\label{tab:model_name_mapping}

\begin{tabular}{ll}
\toprule
Abbrev. & Full Model Identifier \\
\midrule

RLG-S2 & RLG-Stage2-Heavy \\
RLG-S2-V2 & RLG-Stage2-Heavy-v2 \\
RLG-8B-V12 & RLG-Model8B-Ver12 \\
RLG-TS-V7 & RLGame-ts-v7 \\
Corleone-Ref & Corleone\_Reflextion \\
STARS-T2-V7 & STARS Agent Track2 V7 \\
Odyssean-Gen & Odyssean\_Generalization	\\
Odyssean-Soc-V2 & Odyssean\_Social2 \\
Tungsten-Soc-V2 & tungsten\_social\_v2 \\
BanMaHeavy & BanMaHeavy \\
Revac-online & Revac-online \\
Vito-1.0 & Vito-1.0 \\
ZeroR-SM-V4 & ZeroR-SecretMafia-Efficient-v4 \\
OrderOP-j-V6 & OrderOfPhoenix/jarvis\_v6 \\
In2AI-Base & In2AI\_model \\
In2AI-S7-St30-T1.0 & In2AI/mindgames-in2ai-submission-stage-7-step-30-temperature-1.0 \\
In2AI-S7-St45-T1.0 & In2AI/mindgames-in2ai-submission-stage-7-step-45-temperature-1.0 \\
In2AI-S7-St45-T0.6 & In2AI/mindgames-in2ai-submission-stage-7-step-45-temperature-0.6 \\
In2AI-T1.0 & In2AI/mindgames-in2ai-submission-temperature-1.0 \\
In2AI-T0.8 & In2AI/mindgames-in2ai-submission-temperature-0.8 \\
In2AI-T0.6 & In2AI/mindgames-in2ai-submission-temperature-0.6 \\
In2AI-S7-St30-T0.8 & In2AI/mindgames-in2ai-submission-stage-7-step-30-temperature-0.8 \\
In2AI-Mix-V1 & In2AI/mindgames-in2ai-submission-mix-v1 \\
In2AI-Mv3 & In2AI/mindgames-in2ai-submission-m-v3 \\

\bottomrule
\end{tabular}
\end{table}

%% file: sections/E_appendix_trueskill_timeseries.tex
\clearpage
\section{TrueSkill Time Series}\label{appendix:ts_timeseries}

\subsection{Stage 1}
    
    
    
    
    
    
    
    
    

\begin{figure}[!ht]
    \centering
    
    \begin{subfigure}[t]{0.48\linewidth}
        \centering
        \includegraphics[width=\linewidth]{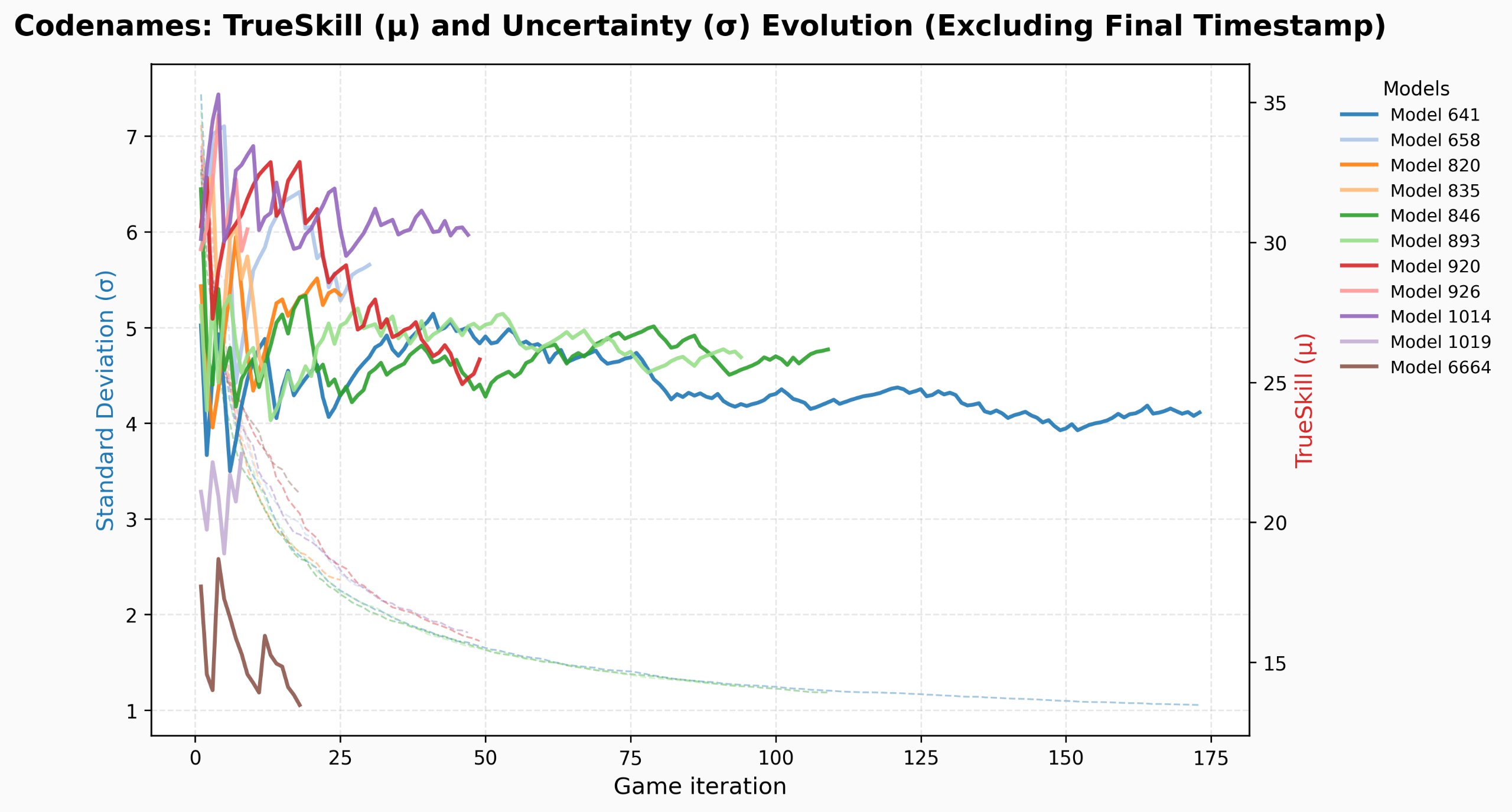}
        \caption{Codenames}
        \label{fig:codenames-stage1}
    \end{subfigure}
    \hfill
    \begin{subfigure}[t]{0.48\linewidth}
        \centering
        \includegraphics[width=\linewidth]{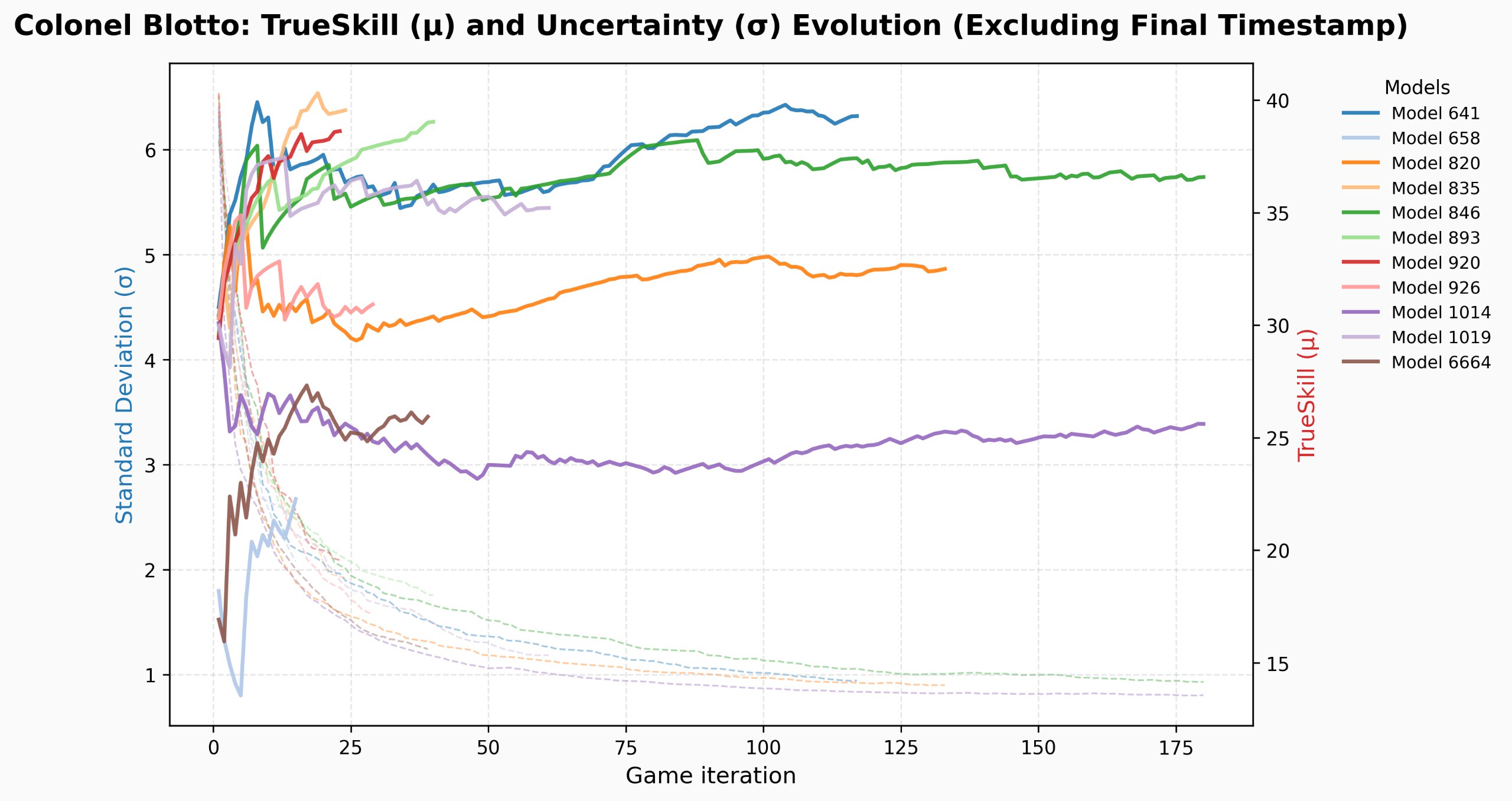}
        \caption{Colonel Blotto}
        \label{fig:blotto-stage1}
    \end{subfigure}

    \vspace{0.5em}

    \begin{subfigure}[t]{0.48\linewidth}
        \centering
        \includegraphics[width=\linewidth]{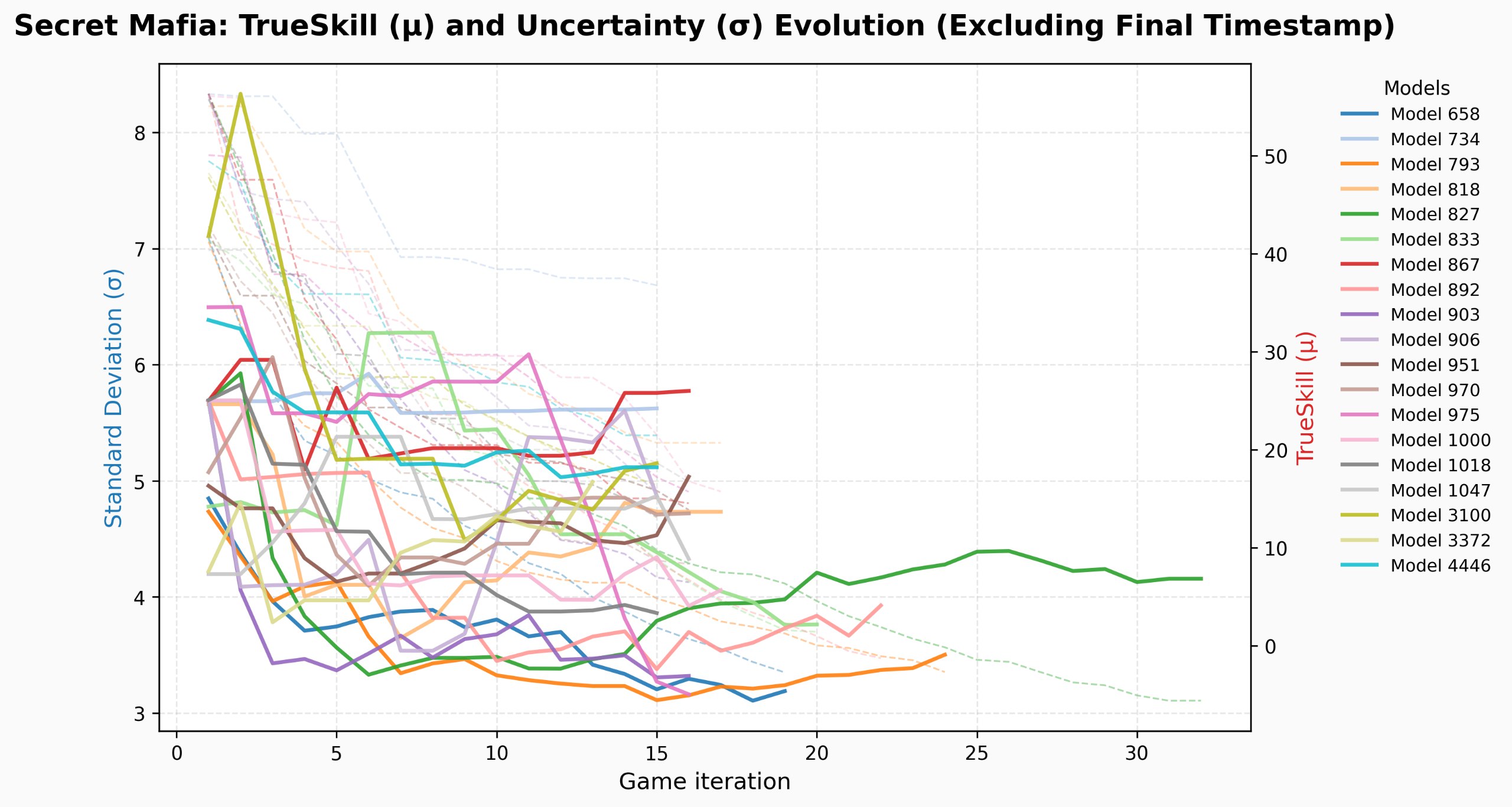}
        \caption{Secret Mafia}
        \label{fig:mafia-stage1}
    \end{subfigure}
    \hfill
    \begin{subfigure}[t]{0.48\linewidth}
        \centering
        \includegraphics[width=\linewidth]{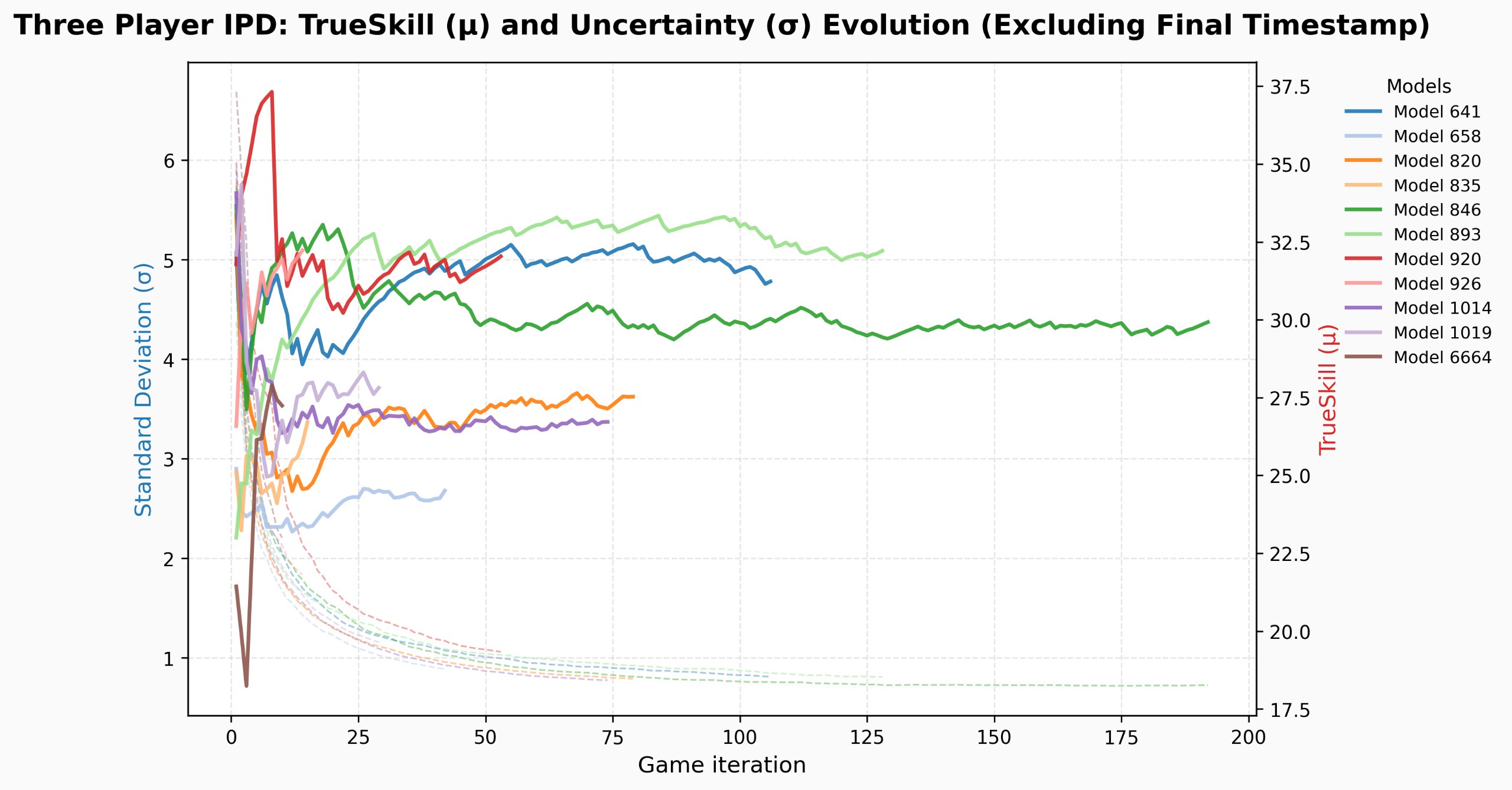}
        \caption{Iterated Prisoner's Dilemma}
        \label{fig:ipd-stage1}
    \end{subfigure}

    \caption{TrueSkill trajectories across game environments. }
    \label{fig:trueskill-all-stage1}
\end{figure}

\subsection{Stage 2}

\begin{figure}[!ht]
    \centering
    
    \begin{subfigure}[t]{0.48\linewidth}
        \centering
        \includegraphics[width=\linewidth]{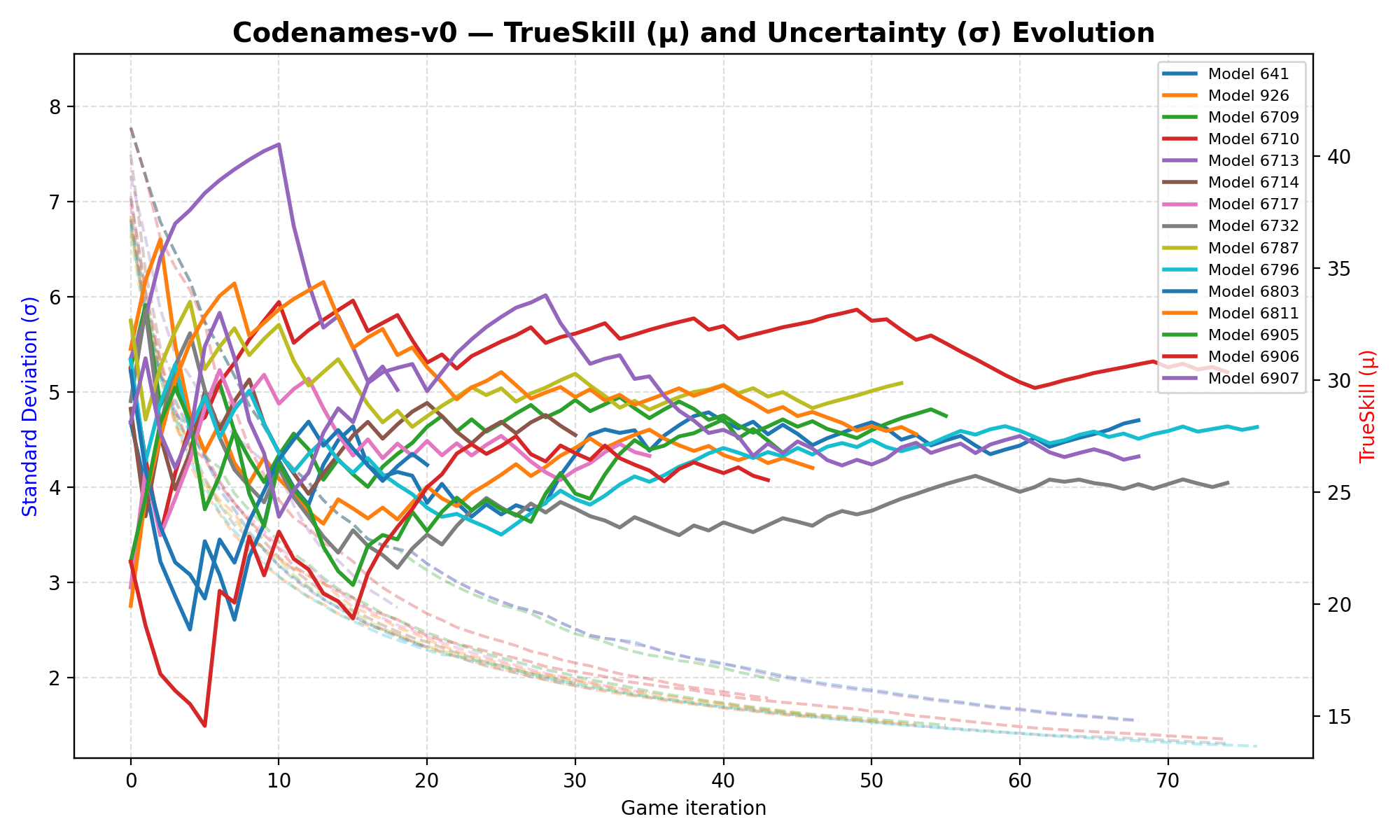}
        \caption{Codenames}
        \label{fig:codenames-stage2}
    \end{subfigure}
    \hfill
    \begin{subfigure}[t]{0.48\linewidth}
        \centering
        \includegraphics[width=\linewidth]{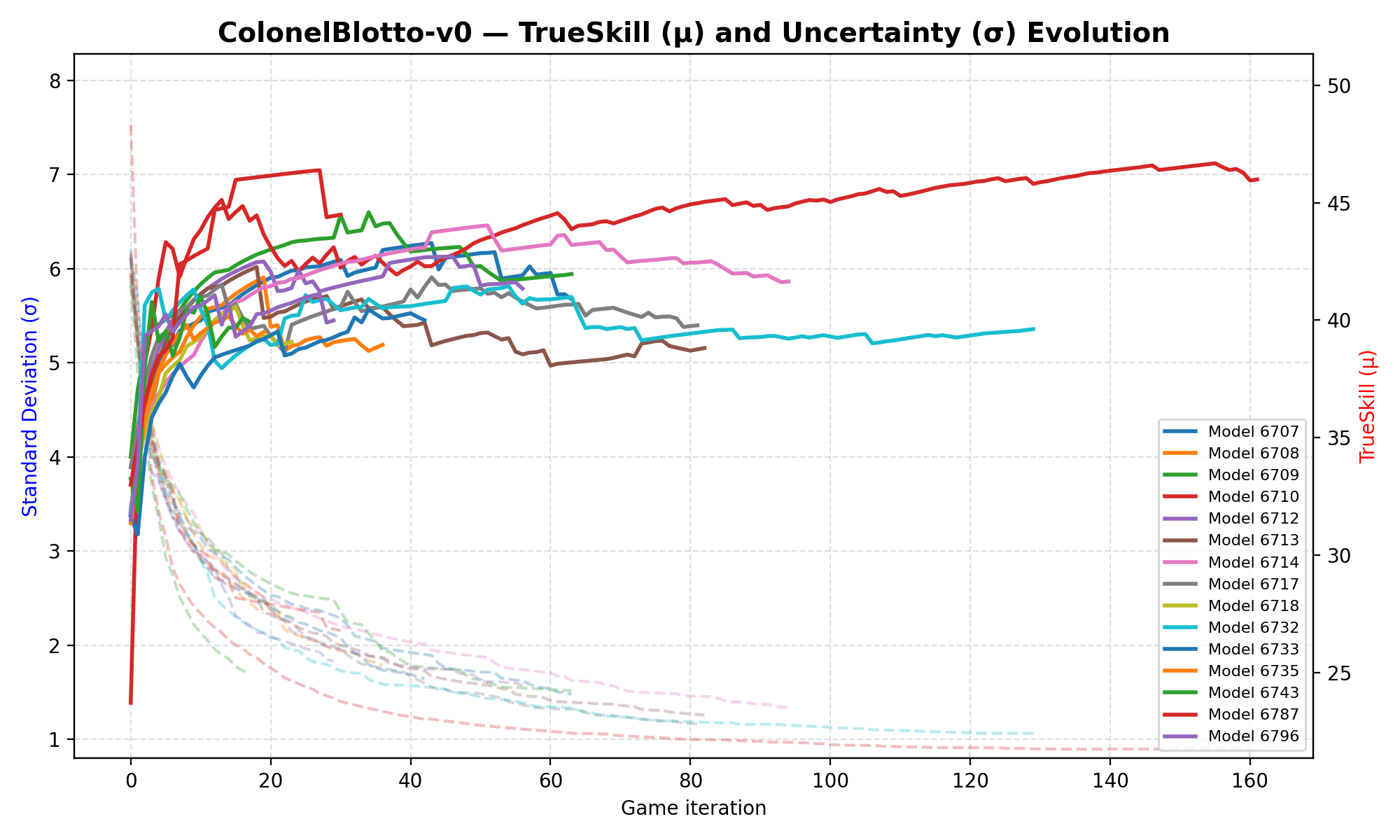}
        \caption{Colonel Blotto}
        \label{fig:blotto-stage2}
    \end{subfigure}

    \vspace{0.5em}

    \begin{subfigure}[t]{0.48\linewidth}
        \centering
        \includegraphics[width=\linewidth]{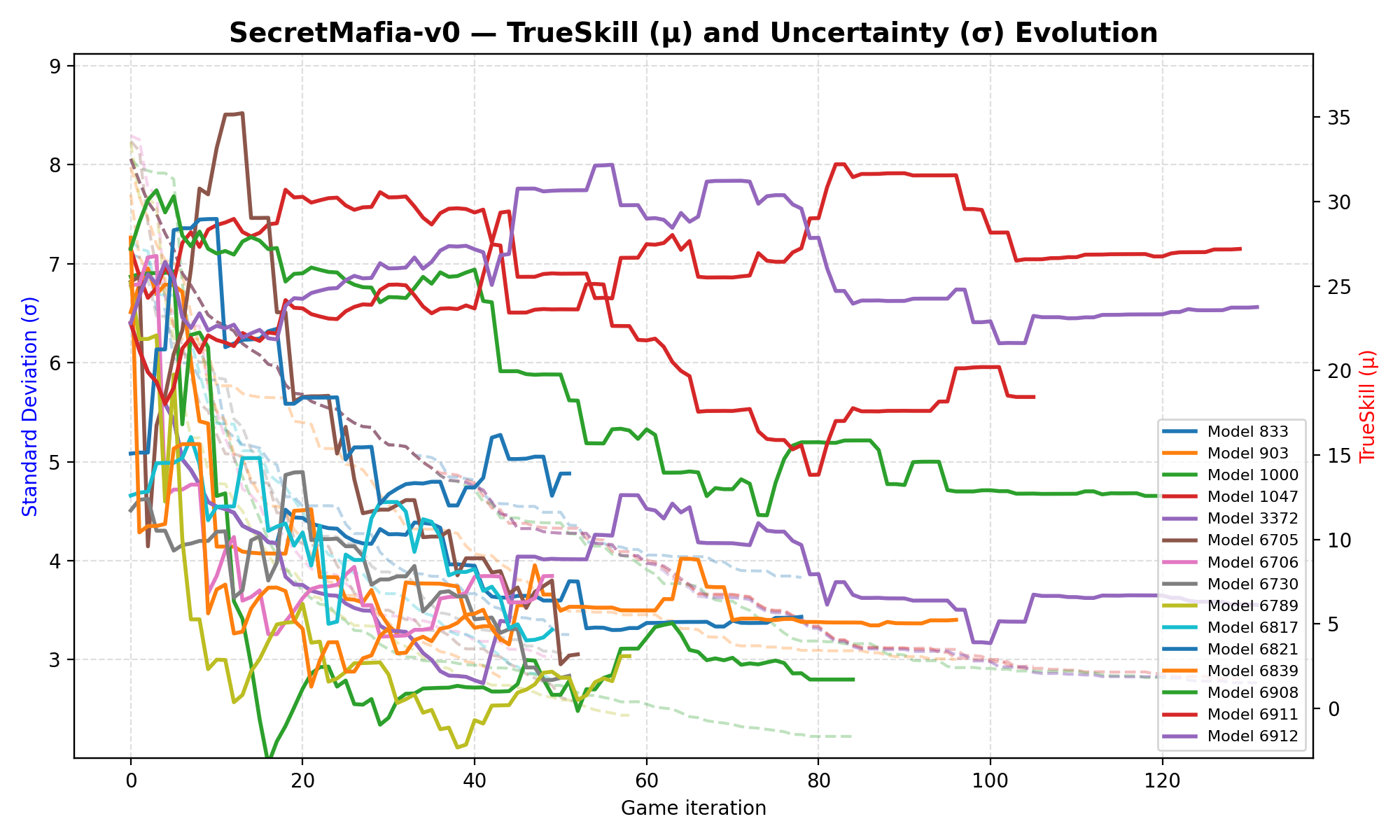}
        \caption{Secret Mafia}
        \label{fig:mafia-stage2}
    \end{subfigure}
    \hfill
    \begin{subfigure}[t]{0.48\linewidth}
        \centering
        \includegraphics[width=\linewidth]{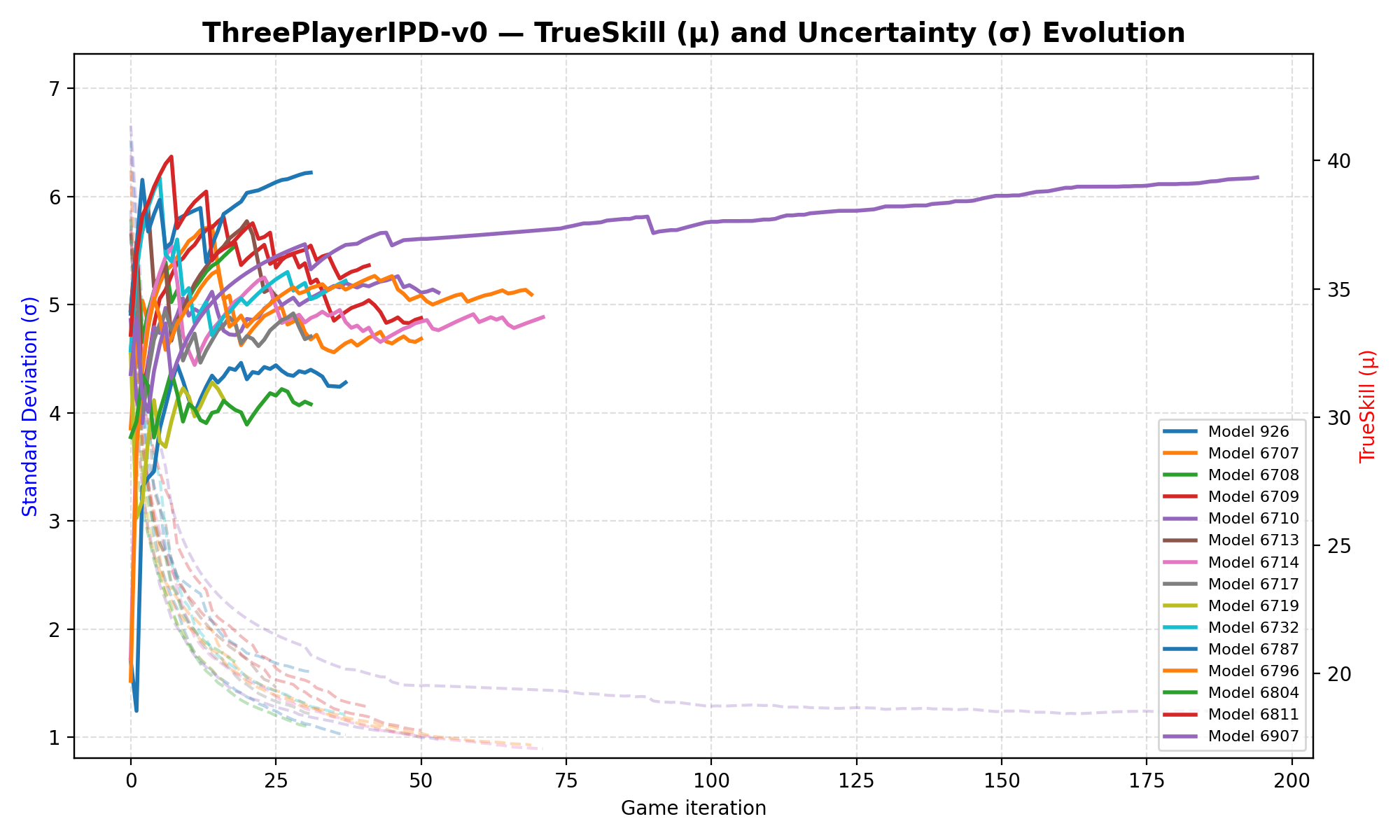}
        \caption{Iterated Prisoner's Dilemma}
        \label{fig:ipd-stage2}
    \end{subfigure}

    \caption{TrueSkill trajectories across game environments.}
    \label{fig:trueskill-all-stage2}
\end{figure}

\section{Win Rate Distributions by Role}\label{appendix:win-rate-adv}

\begin{figure}[!ht]
    \centering
    \begin{subfigure}[t]{0.45\textwidth}
        \centering
        \includegraphics[width=\textwidth]{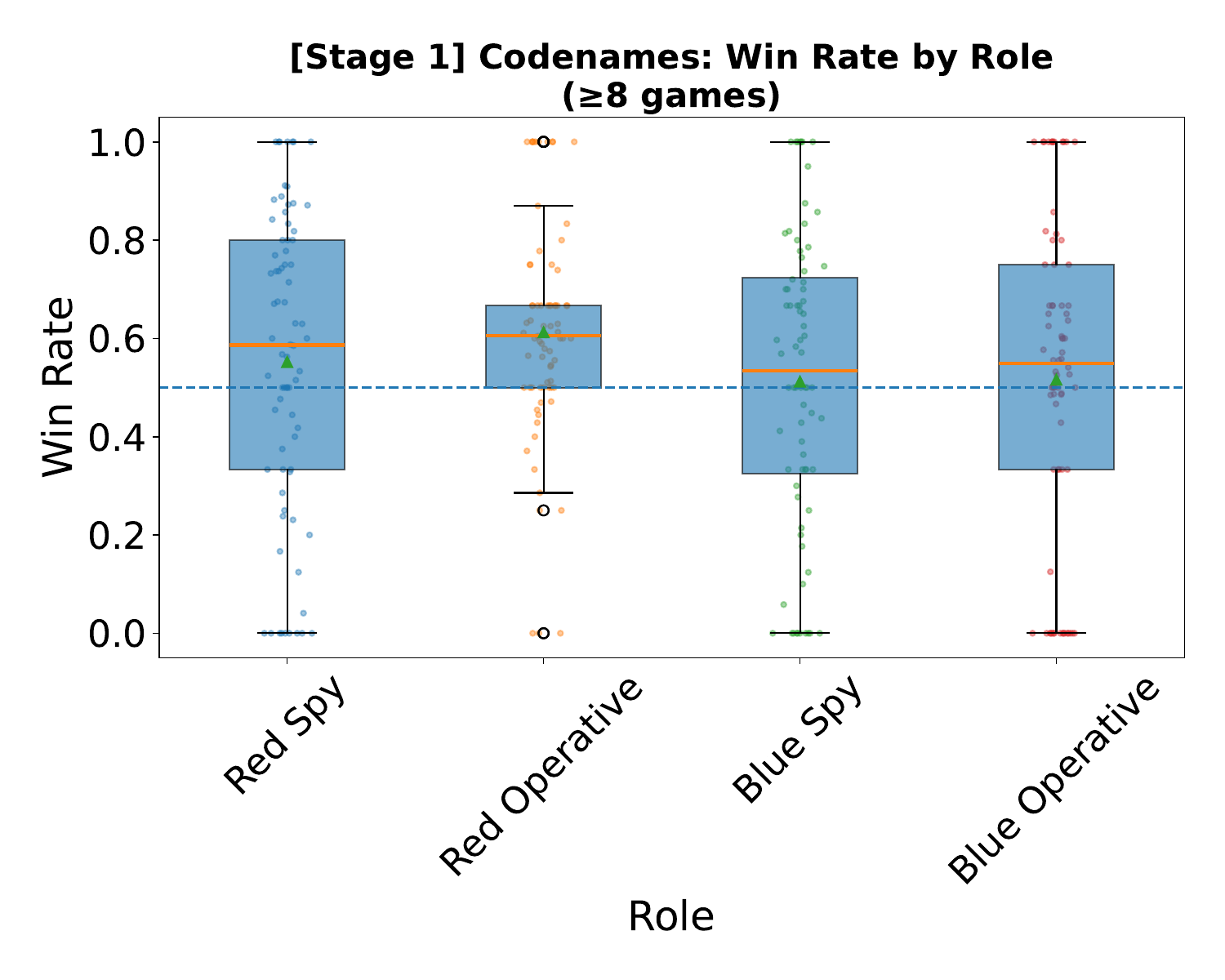}
        \caption{}
    \end{subfigure}
 \hspace{1cm}
    \begin{subfigure}[t]{0.45\textwidth}
    \centering
    \includegraphics[width=\textwidth]{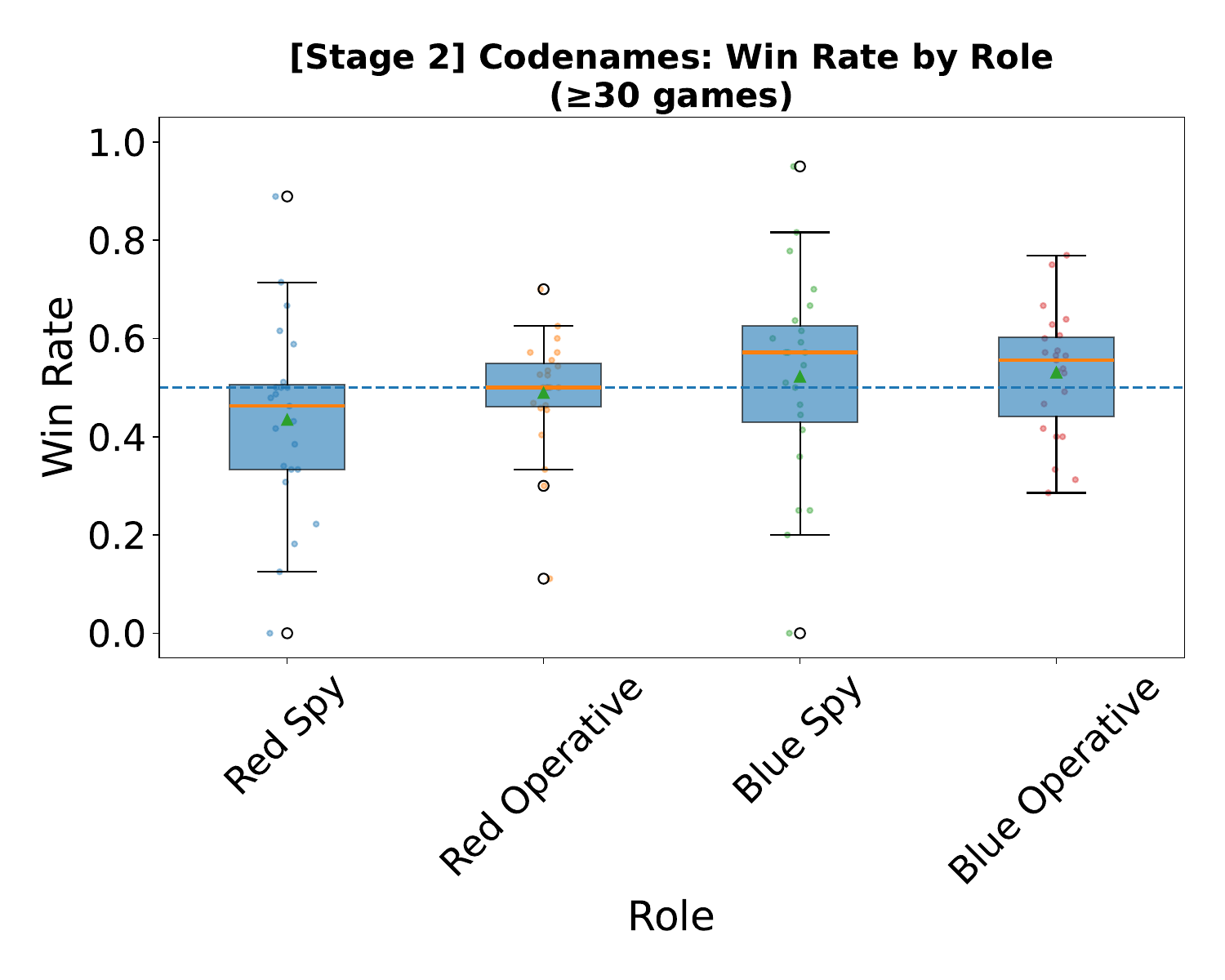}
    \caption{}
    \end{subfigure}
 \hfill
    \begin{subfigure}[t]{0.45\textwidth}
    \centering
    \includegraphics[width=\textwidth]{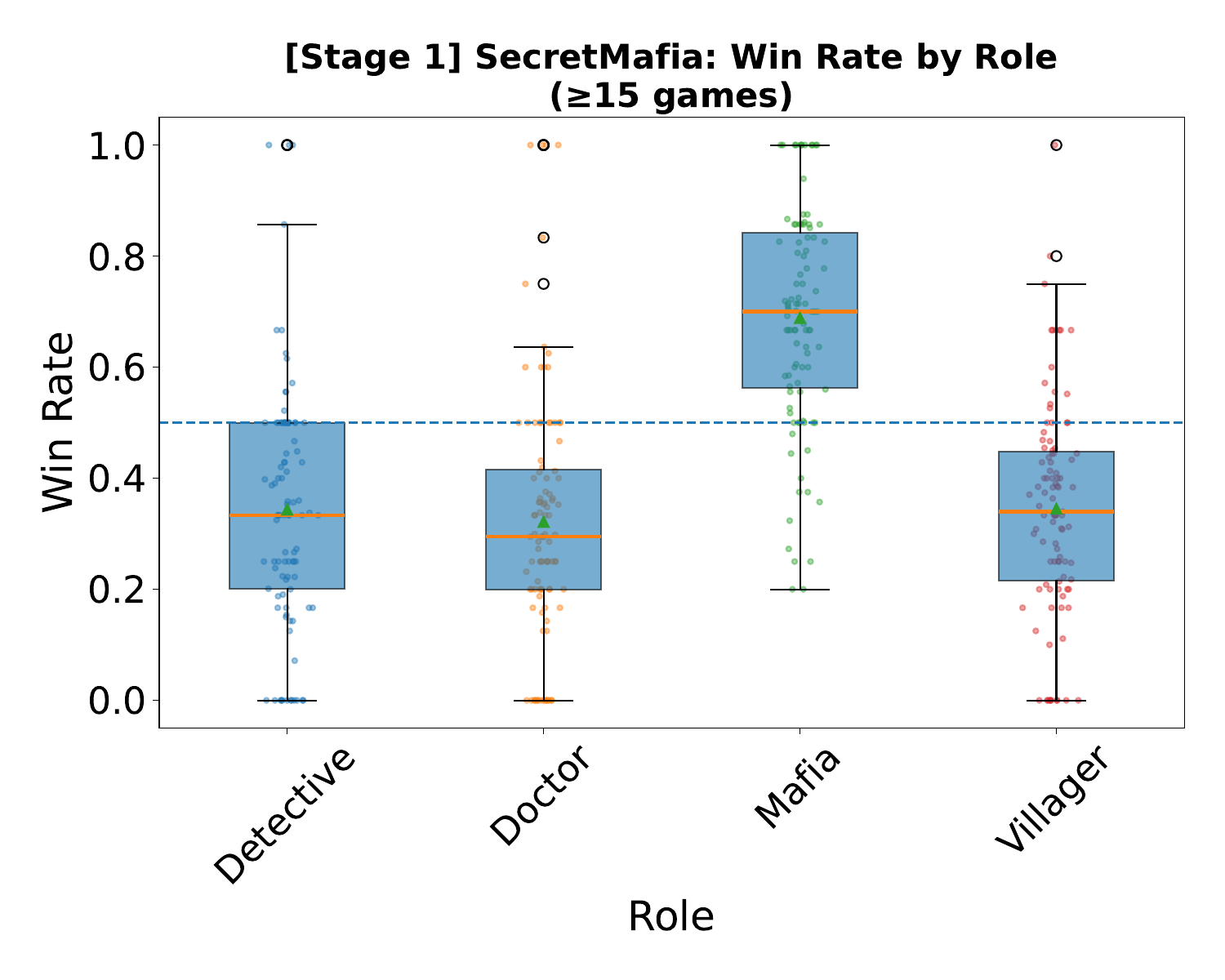}
    \caption{}
    \end{subfigure}
\hspace{1cm}
    \begin{subfigure}[t]{0.45\textwidth}
    \centering
    \includegraphics[width=\textwidth]{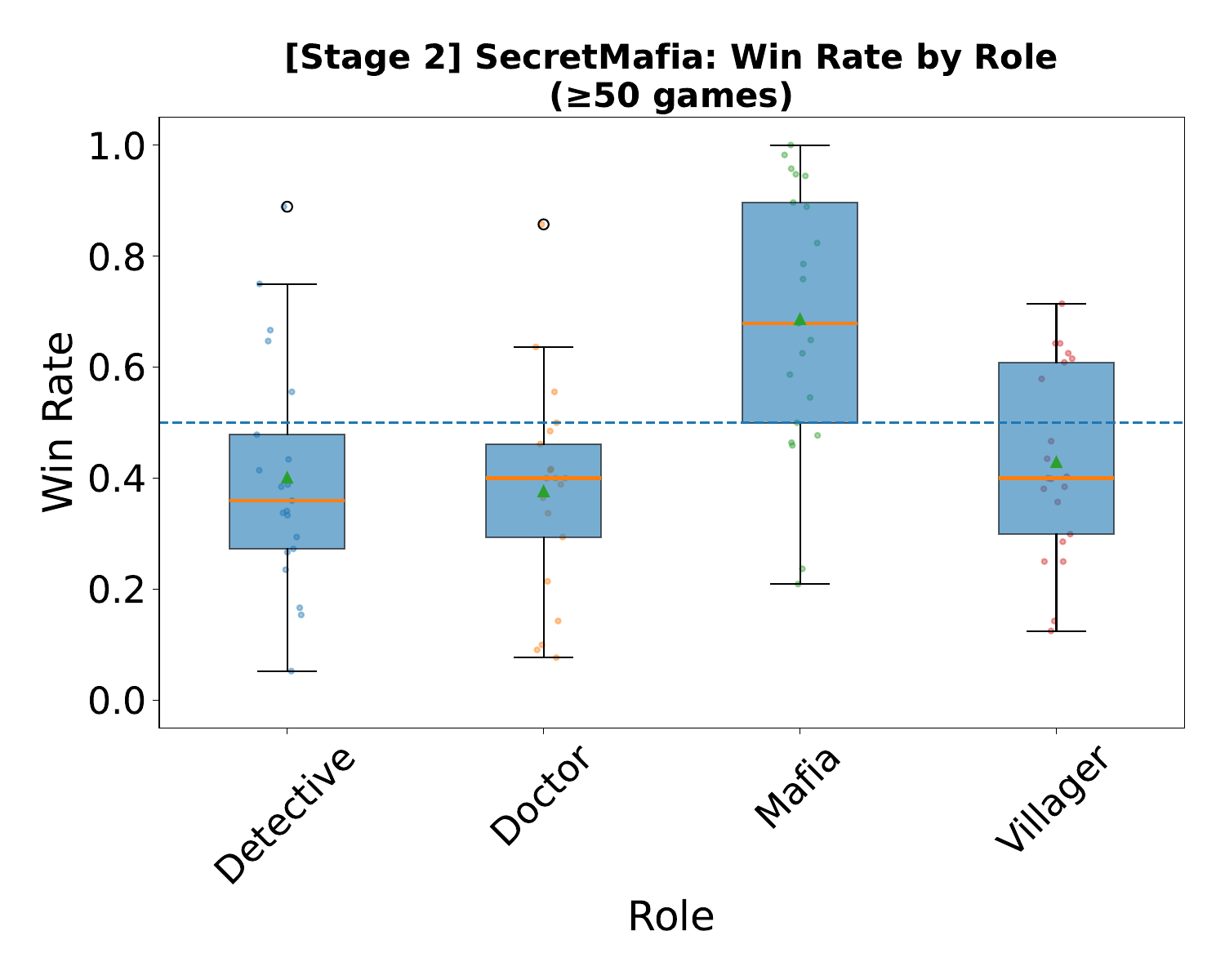}
    \caption{}
    \end{subfigure}
    \caption{Win Rate across Codenames and Secret Mafia over the two stages of the competition. Boxplots show the distribution of win-rates across models for each role. The dashed horizontal line denotes the win rate of 50\%. In Codenames, win rates are quite identical, indicating minimal role bias, whereas in Secret Mafia, the "Mafia" role exhibits consistently higher win rates. Circular markers highlight statistical outliers beyond 1.5$\times$IQR.}
    \label{fig:win-rate-advantage}
     \vspace{-1em}
\end{figure}

%% file: sections/F_appendix_evaluation_protocol.tex
\clearpage
\section{NeurIPS 2025 Competition Evaluation Protocol Details}
\label{appendix:protocol}

This appendix documents the operational details of the NeurIPS 2025 competition. While the benchmark specification in Section~\ref{sec:setup} describes the reusable components, this appendix records the specific schedule, qualification criteria, and matchmaking procedures used in this evaluation cycle.

\subsection{Stage I: Open Online Ladder (July 7 -- October 7, 2025)}
\label{appendix:stage1}

Stage~I was a persistent online ladder open to the broader community throughout the summer. Teams could register at any time and deploy agents into either the Unlimited or Efficient division, or both simultaneously. Agents were matched asynchronously using a queue-based scheduler and a shared TrueSkill-based rating system. Organizer-provided baseline agents of varying strength were injected into the matchmaking pool to maintain queue activity when participant traffic was low.

Stage~I served two objectives:
\begin{enumerate}[leftmargin=1.5em]
    \item \textbf{Experimentation.} Teams could iteratively debug, tune, and redeploy agents under realistic multi-game load, observing how strategies scaled across different game settings.
    \item \textbf{Qualification.} A ``NeurIPS subset'' window (September 23 -- October 7, 2025) defined eligibility for Stage~II, requiring minimum activity thresholds per environment (at least 15 games in Secret Mafia and 10 games in each of the Generalization environments). Agents below these thresholds were excluded from the final Stage~I ranking.
\end{enumerate}

\subsection{Stage II: Final Evaluation (October 24 -- November 10, 2025)}
\label{appendix:stage2}

Stage~II was an independent final evaluation designed to stress-test robustness under a denser, more controlled schedule. Submissions from the top-qualified teams in each division were frozen, and ratings were reset so that only Stage~II games contributed to the final scores.

\paragraph{Efficient Division.} Organizers collected participants' agent implementations and evaluated them in a single shared matchmaking pool over a three-day window.

\paragraph{Unlimited Division.} Teams submitted models to the live matchmaking system over two weekends, with matchmaking criteria that prioritized team-submitted models playing against one another. Baseline agents with fixed TrueSkill priors from Stage~I remained in the queue as stable reference opponents.

\paragraph{Minimum-play requirements.} Each agent was required to complete at least 50 games in Secret Mafia and at least 30 games for each Generalization environment (Colonel Blotto, IPD, and Codenames) during Stage~II.

\paragraph{Matchmaking lessons.} Stage~I revealed that static, always-online baselines were exploitable, enabling agents to inflate ratings by farming wins against weak opponents. In Stage~II, we concentrated matchmaking among participant-submitted agents, fixed baselines at stable ratings, and increased minimum game counts. These changes reduced farming incentives and produced more informative TrueSkill rankings. However, the qualified agent pool was relatively small, leading to sparse matchmaking; larger pools ($>$15 concurrently active models per track) would likely yield more stable ratings in future cycles. 

%% file: sections/G_appendix_tournament_protocol.tex
\clearpage
\section{MG-Ref: Offline Tournament Protocol}
\label{appendix:tournament}

This appendix specifies the offline tournament protocol that pairs a participant model $m$ against the MindGames Reference Set (MG-Ref, Section~\ref{subsec:reference}). Only the components needed to reproduce the design are given here; full implementation, YAML manifests, convergence analyses, and per-cell confidence-interval derivations are in the starter kit at \url{https://github.com/mind-games-challenge/mindgames-starter-kit}.

\subsection{Schedule per Environment}
\label{appendix:tournament-schedule}

For each environment, the schedule is the smallest clean factorial design that (i)~meets or exceeds the Stage~II minimum-play floor (Appendix~\ref{appendix:protocol}), (ii)~balances $m$'s exposure across seats and roles, and (iii)~rotates reference identities so that every member of $\mathcal{M}_\mathrm{ref}$ appears equally often in every non-$m$ seat. Table~\ref{tab:tournament-schedule} summarizes the design.

\begin{table}[h]
\centering
\small
\setlength{\tabcolsep}{6pt}
\caption{MG-Ref offline tournament schedule. $|\mathcal{M}_\mathrm{ref}|$ is the number of references in the track; $|\mathcal{G}|$ is the number of scheduled games per participant. Confidence intervals are Wald 95\% at $p=0.5$; role-conditioned subsets have wider intervals scaling as $1/\sqrt{n}$.}
\label{tab:tournament-schedule}
\begin{tabular}{llrrlc}
\toprule
Track & Environment & $|\mathcal{M}_\mathrm{ref}|$ & $|\mathcal{G}|$ & Factorial design & 95\% CI \\
\midrule
Generalization   & Colonel Blotto & 3 & 30 & 3 opp $\times$ 2 seat $\times$ 5 rep                        & $\pm 18\%$ \\
Generalization   & IPD            & 3 & 36 & 3 multiset $\times$ 3 seat $\times$ 2 order $\times$ 2 rep  & $\pm 16\%$ \\
Generalization   & Codenames      & 3 & 36 & 4 role $\times$ 3 teammate $\times$ 3 rep                   & $\pm 16\%$ \\
Social Deduction & Secret Mafia   & 4 & 96 & 4 dup $\times$ 6 seat $\times$ 4 rep                        & $\pm 10\%$ \\
\midrule
\multicolumn{3}{l}{Total per participant} & \textbf{198} & & \\
\bottomrule
\end{tabular}
\end{table}

\paragraph{Generalization environments.}
Colonel Blotto (two-player) uses all three references as opponents across both seats. Three-player IPD uses the three two-reference multisets (the three two-element subsets of $\mathcal{M}_\mathrm{ref}$), times three $m$-seats and two orderings of the remaining seats; mirror matchups among references are excluded. Codenames assigns $m$ to each of the four team roles (Red Spymaster, Red Operative, Blue Spymaster, Blue Operative) and rotates the teammate and opposing-team references across replicates, directly yielding the role-conditioned estimates recommended in Section~\ref{subsec:diversity}.

\paragraph{Secret Mafia.}
Six players are required and $\mathcal{M}_\mathrm{ref}$ provides four references plus $m$, so one reference is duplicated per game. The schedule indexes over four duplication modes $D_X$ (one per reference $X$), six $m$-seats, and four replicates. Roles are assigned by the environment's natural shuffle, matching the $(2,1,1,2)$ proportions for (Mafia, Doctor, Detective, Villager) at six players; role-conditioned counts are reported alongside aggregates.

\paragraph{Choice of game counts.}
The per-environment counts in Table~\ref{tab:tournament-schedule} were chosen at levels where TrueSkill posteriors have visibly converged in the Stage~I and Stage~II time-series trajectories (Appendix~\ref{appendix:ts_timeseries}). The starter kit additionally ships convergence curves for $\sigma_m$ as a function of $|\mathcal{G}|$ under this factorial design, which confirm that the recommended counts stabilize the posterior within the confidence bands reported above. Participants seeking tighter per-cell intervals can scale via the \texttt{replicate\_multiplier} knob documented in the repository.

\subsection{Rating, Determinism, and Reporting}
\label{appendix:tournament-rating}

\paragraph{Rating.}
Each reference is instantiated with its frozen Stage~II TrueSkill posterior $(\mu_r, \sigma_r)$ from Appendix~\ref{appendix: final rankings}; $m$ starts from the competition prior $\mathcal{N}(\mu=25,\sigma=25/3)$ (Section~\ref{subsec:tracks}). Updates are restricted to $m$'s posterior: reference ratings are held constant. This reproduces the Stage~II design choice that removed the rating-farming incentive present in Stage~I (Appendix~\ref{appendix:stage2}) and is what makes offline scores comparable across independent participants.

\paragraph{Determinism.}
Each match is indexed by $(\mathrm{env}, k)$ with $k \in \{0,\dots,|\mathcal{G}|-1\}$, and the environment seed is a fixed offset plus $k$ (base offsets 10{,}000 / 20{,}000 / 30{,}000 / 40{,}000 for Colonel Blotto / IPD / Codenames / Secret Mafia). Role shuffles in Secret Mafia use Python's global RNG, which the runner reseeds immediately before each \texttt{env.reset()}. Two executions of the same manifest with the same reference checkpoints produce identical seating, role assignment, and match ordering.

\paragraph{Reporting.}
Every run emits, per environment and in aggregate: TrueSkill posterior $(\mu_m, \sigma_m)$, win rate with 95\% Wilson intervals, cumulative reward $\sum_{G}\mathcal{R}_m(s_\mathrm{term})$, role-conditioned win rates with realized per-role counts, and the error-attribution columns \textit{Clean}, \textit{Caused}, \textit{Witnessed}, \textit{Self-Forf.}, \textit{Opp-Forf.}\ as defined in Table~\ref{table:top_model_specific_error}. No single summary score is promoted above the others: in environments where TrueSkill and error-attribution disagree, that divergence is itself the measurement Section~\ref{subsec:confounds} recommends surfacing.

\subsection{Caveats}
\label{appendix:tournament-caveats}

Two caveats accompany the protocol. First, Secret Mafia runs with only five active identities (four references plus $m$), well below the 15-agent pool Appendix~\ref{appendix:stage2} identifies as producing stable Stage~II ratings; offline Secret Mafia TrueSkill should therefore be read as a local estimate against $\mathcal{M}_\mathrm{ref}$, not a reproduction of the Social Deduction leaderboard, and the clean / caused / witnessed / self-forfeit breakdown should be reported before any TrueSkill point estimate. Second, MG-Ref is anchored to Stage~II posteriors and will drift from any future live leaderboard as new agents enter the pool; the starter kit documents how to extend the factorial designs in Table~\ref{tab:tournament-schedule} when the pool is refreshed in subsequent cycles.